\documentclass{article} 
\usepackage{ChEF,times}


\maketitle

\begin{abstract}

Multimodal Large Language Models (MLLMs) have shown impressive abilities in interacting with visual content with myriad potential downstream tasks.
However, even though a list of benchmarks has been proposed, the capabilities and limitations of MLLMs are still not comprehensively understood, due to a lack of a standardized and holistic evaluation framework.
To this end, we present the first \textit{Comprehensive Evaluation Framework} (ChEF) that can holistically profile each MLLM and fairly compare different MLLMs. 
%
%
First, we structure ChEF as four modular components, \emph{i.e.}, \textit{Scenario} as scalable multimodal datasets, \textit{Instruction} as flexible instruction retrieving formulae, \textit{Inferencer} as reliable question-answering strategies, and \textit{Metric} as indicative task-specific score functions. 
Based on them, ChEF facilitates versatile evaluations in a standardized framework, and new evaluations can be built by designing new \textit{Recipes} (systematic selection of these four components).
Notably, current MLLM benchmarks can be readily summarized as recipes of ChEF.
Second, we introduce 6 new recipes to quantify competent MLLMs' desired capabilities (or called desiderata, \textit{i.e.}, calibration, in-context learning, instruction following, language performance, hallucination, and robustness) as reliable agents that can perform real-world multimodal interactions.
Third, we conduct a large-scale evaluation of 9 prominent MLLMs on 9 scenarios and 6 desiderata. Our evaluation summarized over 20 valuable observations concerning the generalizability of MLLMs across various scenarios and the composite capability of MLLMs required for multimodal interactions. 
%
%
Codes and data are now available at \href{https://github.com/OpenLAMM/LAMM}{\texttt{\textcolor[HTML]{EC0089}{https://openlamm.github.io}}}

\end{abstract}
\section{Introduction}
\vspace{-0.2cm}
\begin{figure}[t]
    \centering
    \includegraphics[width=0.9\textwidth]{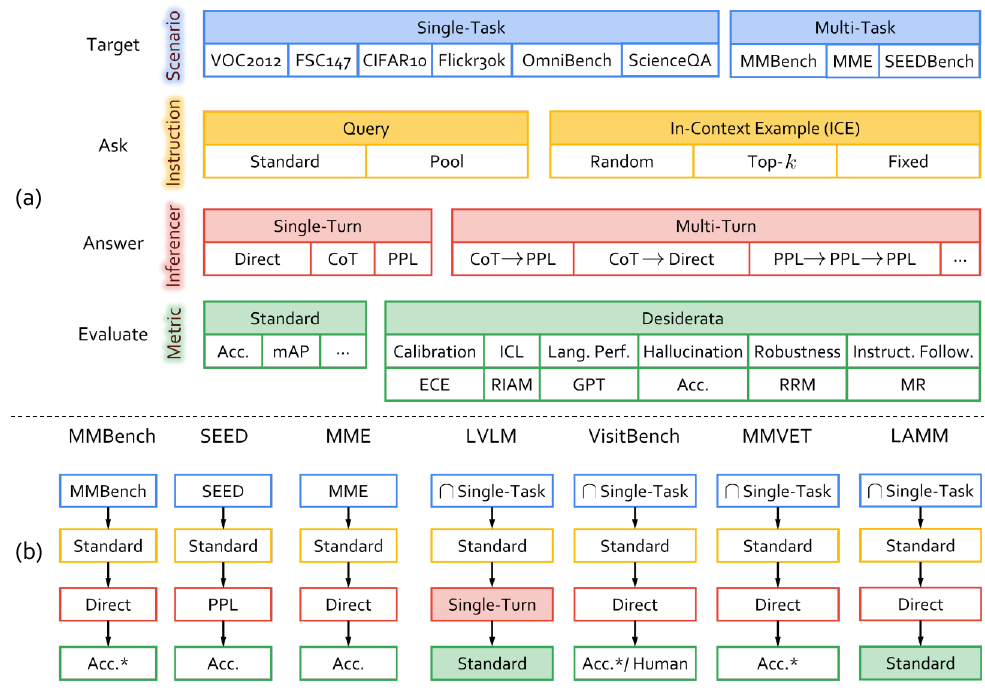}
    \caption{(a) ChEF Overview. (b) Current MLLM benchmarks can be readily absorbed into ChEF. \textit{Acc.} is the accuracy. \textit{Acc.*} is the accuracy from GPT-based metric. \textit{$\cap$} means overlap with ChEF.  \textit{ICL}, \textit{Lang. Perf.}, \textit{Instruct. Follow.} are shorts for in-context learning, language performance, and instruction following, respectively.}
    \vspace{-0.6cm}
  \label{fig:overview}
\end{figure}

By applying the powerful Large Language Models (LLMs)~\citep{openai2023gpt4,chiang2023vicuna,touvron2023llama} as a universal task interface, recent works on Multimodal Large Language Models (MLLMs)~\citep{llava,minigpt4,dai2023instructblip} have shown impressive abilities to interact with visual contents through question-answering dialogues and are expected to address more complex multimodal tasks that can harness LLMs' generalization ability to myriad downstream scenarios.
Yet the capabilities and limitations of MLLMs are still not well understood, and we observe a lack of a standardized framework that can comprehensively evaluate different MLLMs.
Recent benchmarks often focus on building a multimodal evaluation dataset for MLLMs~\citep{li2023seedbench,liu2023mmbench,fu2023mme} or only evaluate one or a few factors of MLLMs~\citep{shao2023tinylvlm, POPE,yu2023mmvet,bitton2023visitbench}, or attempt to establish a framework but lack scalability and have limits in their comprehensiveness~\citep{yin2023lamm,xu2023lvlmehub}~\footnote{More related works are provided in Supplementary Materials (Section A).}. 
This makes a thorough assessment of each model and reliable comparisons among various models challenging.

To address these issues, we believe that a comprehensive evaluation framework, which is specially designed for MLLMs, should encompass scalable datasets about multimodal tasks that can be handled by MLLMs. For each model, we should evaluate the performance in a broad set of perspectives (\emph{i.e.} capabilities more than multimodal perception and reasoning, such as robustness, in-context learning, and \emph{etc.}) that are vital to profile the intrinsic properties of MLLMs, especially as the agents that can perform real-world multimodal interaction. 
Moreover, meaningful comparisons among MLLMs require standardization in the evaluation process so that each model can be conveniently adapted.
To this end, as shown in Figure~\ref{fig:overview}(a), we present ChEF, a Comprehensive Evaluation Framework for reliable and indicative assessment of MLLMs, which is highly scalable and can be flexibly modified to adapt to the evaluation of any new model or task. It is modularly designed with four components, \emph{i.e.}, \textit{Scenario}, \textit{Instruction}, \textit{Inferencer}, and \textit{Metric}.
\vspace{-0.1cm}

\textbf{(1)} \textbf{Scenarios} are a set of datasets concerning representative multimodal tasks that are suitable for MLLMs. \textit{Scenarios} are scalable by design, allowing the inclusion of any related dataset if necessary. We have included several prominent single-task datasets, such as CIFAR-10~\citep{cifar10} for image classification, VOC2012~\citep{pascal-voc-2012} for object detection, ScienceQA~\citep{scienceqa} for multimodal question-answering. Recent multi-task benchmark datasets proposed for evaluating MLLMs, such as MMBench~\citep{fu2023mme} and SEEDBench~\citep{li2023seedbench}, are also accessible as \textit{Scenarios}.
\vspace{-0.1cm}

\textbf{(2)} \textbf{Instruction} focuses on how to pose questions and set instruction examples to the MLLMs. We integrate various standard queries and query pools adaptive to each MLLM, and multimodal in-context example (\texttt{ICE}) retrieving strategies for in-context learning (ICL)~\citep{wu2023openicl,brown2020language}. Both are tailored to specific \textit{Scenarios}. To the best of our knowledge, we are the first to incorporate ICL into the evaluation framework. The design of \textit{Instruction} makes it flexible to evaluate diverse \textit{Scenarios} within the same framework.
\vspace{-0.1cm}

\textbf{(3)} \textbf{Inferencer} pertains to how an MLLM answers questions. In a single-turn question-answering (QA), in addition to the standard textual outputs (\texttt{Direct}) that may be hard to compare with the ground-truth answers, we can employ the Perplexity (\texttt{PPL})~\citep{klein-etal-2017-opennmt} to select the most probable candidate answers, or Chain-of-Thought (\texttt{CoT})~\citep{cotref} prompting to increase the reliability of the prediction.
The \textit{Inferencer} also allows \texttt{Multi-Turn}, in which \texttt{PPL}, \texttt{CoT}, and \texttt{Direct} outputs can be applied in turns, and makes the evaluation result reliable.
\vspace{-0.1cm}

\textbf{(4)} \textbf{Metrics} are a set of score functions designed to evaluate the performance of each MLLM. 
For example, we include task-specific metrics such as accuracy for classification or multi-choice QA, mAP for detection, BLEU for captioning, and \emph{etc.} 
More metrics can be included when evaluating the MLLMs from new perspectives, such as Expected Calibration Error (ECE)~\citep{Pakdaman_Naeini_Cooper_Hauskrecht_2015} if we would like to know how the model is aware of its uncertainty in prediction, GPT-based metric~\citep{ChiangL23} if we would like the outputs to be readable as natural language. The inclusion of appropriate and newly defined metrics ensures that the evaluation results are more indicative.
\vspace{-0.1cm}

With a systematic selection of \textit{Scenarios}, \textit{Instructions}, \textit{Inferencers}, and \textit{Metrics}, ChEF facilitates versatile evaluations in a standardized framework. Users can easily build new evaluations according to new \textit{Recipes} (\emph{i.e.} specific choices of the four components). For example, current MLLM benchmarks~\citep{fu2023mme,li2023seedbench,liu2023mmbench,bitton2023visitbench,yu2023mmvet,xu2023lvlmehub,yin2023lamm} can be summarized as different \textit{Recipes}, as shown in Figure~\ref{fig:overview}(b), and thus can be readily absorbed into ChEF. We will extensively discuss the design principles in Section~\ref{sec:design_principles}.
Moreover, we view ChEF as a growing framework, where each component can be evolved according to the emerging techniques or applications. We will continuously update the ChEF framework with a wider range of accessible models and evaluation tasks.
%


%
Based on ChEF, it becomes rather convenient to set up new evaluations to quantify the desired capabilities (or called \textbf{desiderata}) that a competent MLLM model should possess, as a reliable agent that can perform real-world multimodal interactions.
These desiderata include:
\vspace{-0.1cm}
\begin{itemize}
    \item \textbf{Calibration}: Does MLLM express accurate uncertainty and confidence?
    \vspace{-0.1cm}
    \item \textbf{In-context Learning}: Does MLLM learn from instruction examples?
    \vspace{-0.1cm}
    \item \textbf{Instruction Following}: Does MLLM adhere to instructions?
    \vspace{-0.1cm}
    \item \textbf{Language Performance}: Does MLLM describe visual content in readable language?
    \vspace{-0.1cm}
    \item \textbf{Hallucination}: Does MLLM avoid mentioning objects that do not exist in the images?
    \vspace{-0.1cm}
    \item \textbf{Robustness}: Is MLLM robust to corruptions in the multimodal inputs?
\end{itemize}
\vspace{-0.1cm}
Each desideratum is evaluated by constructing the evaluation pipeline from a ChEF \textit{Recipe}. We will introduce the \textit{Recipes} for the desiderata in Section~\ref{sec:desiderata}.

Overall, we comprehensively evaluated 9 MLLMs across 9 \textit{Scenarios} and 6 desiderata. Our evaluation yields the following 3 key findings:

\textbf{(1)} Recent MLLMs cannot perform well across all \textit{Scenarios}. There is a significant tug-of-war issue~\citep{hadsell2020embracing} between different tasks. There are also several critical tasks that can not be addressed by recent MLLMs.
\vspace{-0.1cm}

\textbf{(2)} Recent MLLMs are struggling with in-context learning, instruction following, and robustness, thus they may fall short of real-world multimodal interactions.
\vspace{-0.1cm}

\textbf{(3)} There is a strong correlation between the desiderata and visual performance. Evaluating the desiderata reveals the intrinsic property on \textit{Scenarios} that used to evaluate a composite performance. 
\vspace{-0.1cm}
\section{ChEF: A Comprehensive Evaluation Framework}
\vspace{-0.2cm}
In this section, we first list the design principles of ChEF in Section~\ref{sec:design_principles}, and then depict how to conduct an evaluation process based on a \textit{Recipe} of selecting the four modules in ChEF (Section~\ref{sec:overview_chef}).
Furthermore, we introduce the \textit{Recipes} of six desired capabilities (or called desiderata) that a competent MLLM should have, as shown in Section~\ref{sec:desiderata}.

%

\subsection{Design Principles}
\label{sec:design_principles}
\vspace{-0.2cm}

\begin{figure}[t]
    \centering
    \includegraphics[width=\textwidth]{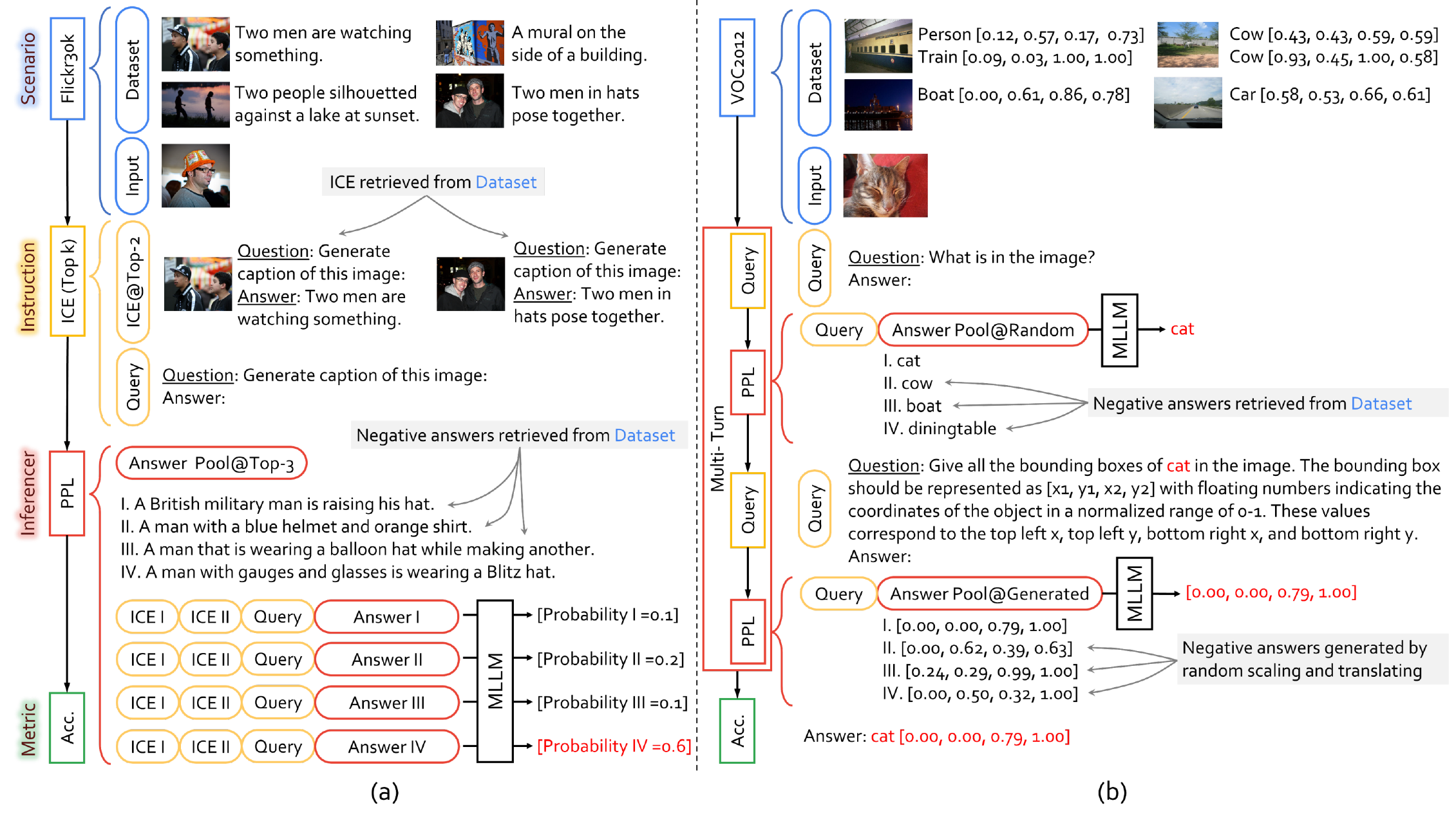}
    \vspace{-0.8cm}
    \caption{\textbf{Two examples of Recipes in ChEF.} A \textit{Recipe} consists of \{\textit{Scenario}, \textit{Instruction}, \textit{Inferencer}, \textit{Metric}\}. The \textit{Recipe} of (a) is \{Flickr30k, \texttt{ICE}, \texttt{PPL}, Accuracy\}, while (b) is \{VOC2012, \texttt{Query},  \texttt{Multi-Turn}, Accuracy\}.}
  \label{fig:evaluation_pipeline_example}
  \vspace{-0.5cm}
\end{figure}

ChEF is a comprehensive evaluation framework aiming at providing a fair and holistic assessment of MLLMs' performance across diverse multimodal tasks.
%
To accomplish this objective, our design principles encompass the following key aspects:
\vspace{-0.1cm}

\textbf{(1) Modular.} 
We decouple the evaluation framework into four modular components~\footnote{Details of these four components are provided in Supplementary Materials (Section B).}: \textit{Scenario}, \textit{Instruction}, \textit{Inferencer}, and \textit{Metric}, so as to enable fast modification of each component and ensure consistent evaluation results across different benchmark datasets.
\vspace{-0.1cm}

\textbf{(2) Scalable.}
We implement easy-to-use interfaces to streamline the integration of new \textit{Scenarios} into the framework and have included almost all recent benchmark datasets into the \textit{Scenario}. 
\vspace{-0.1cm}
%


\textbf{(3) Flexible.} We design ChEF to accommodate the varying input formats supported by different MLLMs, including \texttt{Queries} and in-context learning examples (\texttt{ICE}). Based on these \textit{Instructions}, MLLMs can generate outputs that are suitable for specific \textit{Scenarios}. 
\vspace{-0.1cm}
%


\textbf{(4) Reliable.}
We include three more reliable \textit{Inferencers}, such as \texttt{CoT} and \texttt{PPL}, as well as their multi-round combination (\texttt{Multi-Turn}), in addition to standard free-form outputs (\texttt{Direct}). These \textit{Inferencers} make the evaluation more reliable, and better tailored to reflect the precise perception or reasoning abilities that the \textit{Scenarios} tend to assess.
\vspace{-0.1cm}

\textbf{(5) Indicative.}
We utilize a list of task-specific metrics ranging from metrics for vision tasks to the GPT-based metric for language proficiency. Each MLLM's textual outputs are adapted to these metrics, so as to indicatively measure that the MLLMs can actually perform the target tasks.
\vspace{-0.1cm}

\subsection{Exemplar Recipes and Their Evaluation Processes}
\label{sec:overview_chef}
\vspace{-0.2cm}
%
%
For an illustration of how each component functions and the overall evaluation is processed, we provide two examples of \textit{Recipes} in Figure~\ref{fig:evaluation_pipeline_example}.

\vspace{-0.1cm}
\textbf{(1) Image Captioning on Flicker30k.} 
In Figure~\ref{fig:evaluation_pipeline_example}(a), the \textit{Scenario} is Flickr30k and the task is image captioning.
The \textit{Instruction} does not only include the standard query ``Generate caption of this image'', but also Top-$k$ \texttt{ICE} to guide the generation of captions. These examples are retrieved according to image similarity.
The \textit{Inferencer} applies single-round \texttt{PPL} to measure how each of the four answers (as the answer pool) is consistent with the input image, in the form of probability. The negative answers are retrieved based on text similarity. Using \texttt{PPL} instead of free-form outputs constrains the scope of the captions and thus can be measured more reliably.
Finally, to be compatible with \texttt{PPL}, the \textit{Metric} applies accuracy to determine the correctness of the prediction. 
\vspace{-0.1cm}

\textbf{(2) Object Detection on VOC2012.}
Object detection is another typical vision task. In Figure~\ref{fig:evaluation_pipeline_example}(b), we apply VOC2012 as the \textit{Scenario}. The \textit{Instruction} has no \texttt{ICE}, but just a standard query.
The \textit{Inferencer} is \texttt{PPL} that is conducted in two rounds. In the first round, ask the MLLMs ``What is in the image?'', and in the second round, ask the MLLMs the bounding box of the predicated object. Note that the answer pools of the bounding boxes are generated by random scaling and translating the ground-truth bounding boxes. The \textit{Metric} is accuracy as we transform the detection task into a multi-choice question-answering paradigm.

\subsection{Desiderata}
\label{sec:desiderata}
\vspace{-0.2cm}
\begin{figure}[t]
    \centering
    \includegraphics[width=0.9\linewidth]{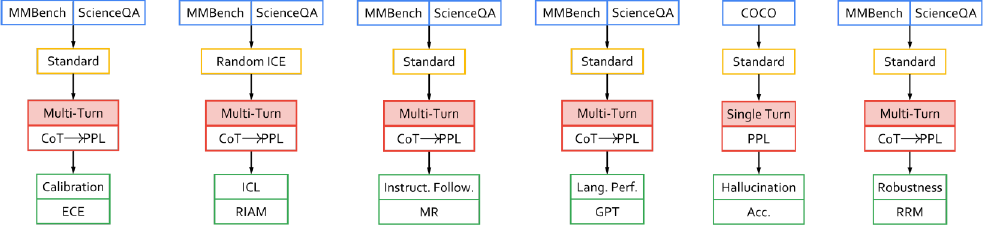}
    \vspace{-0.2cm}
    \caption{\textbf{Recipes for evaluating six dimensions of desiderata.} 1) All six dimensions are assessed on MMBench and ScienceQA, except for Hallucination, which is evaluated solely on MSCOCO; 2) All use standard query as \textit{Instruction}, except ICL uses random \texttt{ICE}; 3) All employ \texttt{Multi-Turn} from \texttt{CoT} to \texttt{PPL} as \textit{Inferencer}, except Hallucination with a single \texttt{PPL}; 4) The \textit{Metric} for each dimension is specifically designed for the respective evaluation.}
  \label{fig:recipes}
  \vspace{-0.3cm}
\end{figure}

As shown in Figure~\ref{fig:recipes}, we implement six more evaluations based on the desiderata that a competent MLLM model should have, \emph{i.e.}, calibration, in-context learning, instruction following, language performance, robustness, and hallucination. Each dimension is assessed using a specially designed \textit{Recipe}.
To fulfill consistent evaluations among different dimensions of the desiderata, the \textit{Scenarios} are almost MMBench~\citep{liu2023mmbench} and ScienceQA~\citep{scienceqa}, except that hallucination is evaluated on MSCOCO~\citep{mscoco}.
The \textit{Inferencers} share a similar strategy. Hallucination applies \texttt{PPL} in a single round, while the rest desiderata use the same \texttt{Multi-Turn} that is composed of \texttt{CoT} and \texttt{PPL}, to increase the reliability of the prediction.
In the following part, we introduce the rest components in each \textit{Recipe}. 

\begin{figure}[t]
    \centering
    \includegraphics[width=\textwidth]{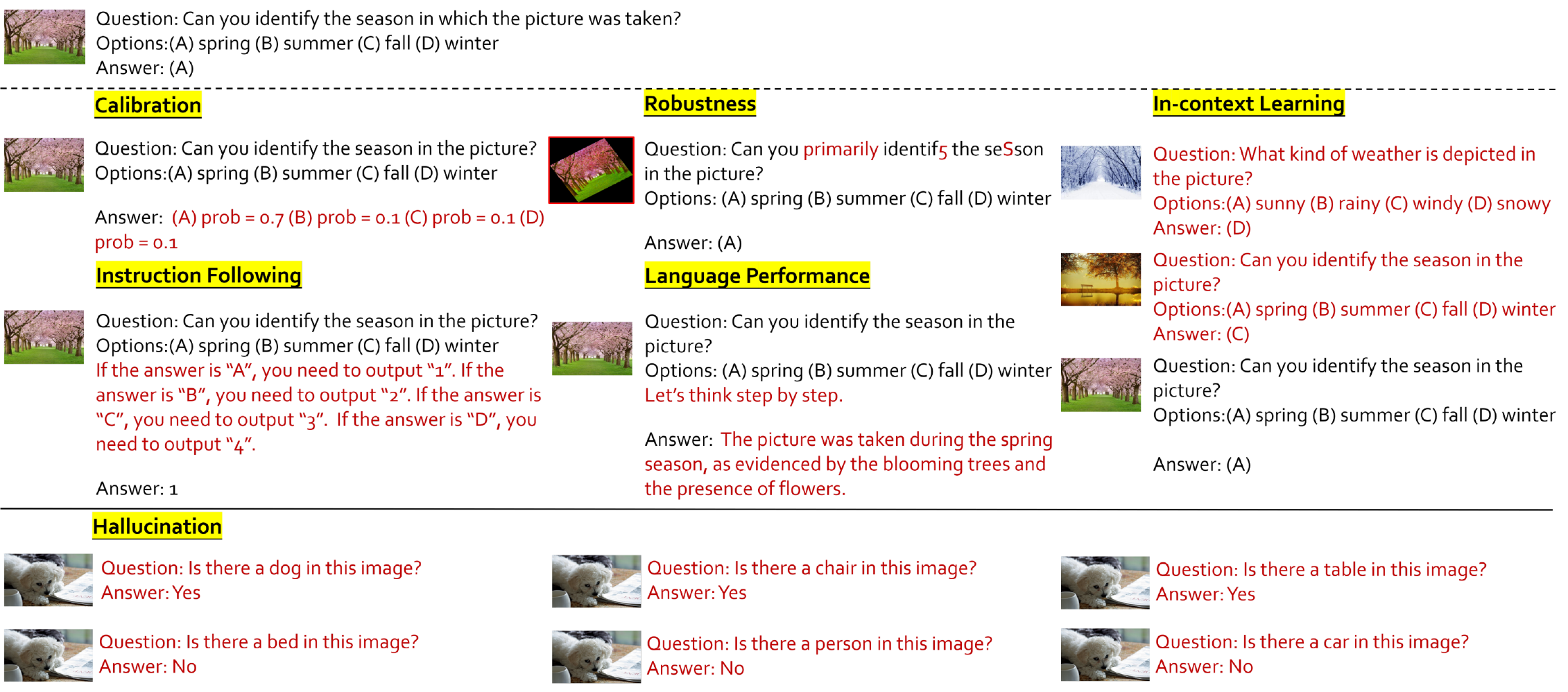}
    \caption{\textbf{The exemplar of desiderata.} The distinguished design of each desideratum is marked in red. For calibration evaluation, the prediction confidence is calculated to determine the gap between confidence and accuracy. Instruction following is evaluated through verbalizer manipulation. In-context learning is evaluated by providing \texttt{ICE} in the \textit{instruction}. Robustness is assessed by introducing noise to both the image and text inputs. Language performance is evaluated by instructing the model to generate chain-of-thought content. Hallucination is solely evaluated on MSCOCO, and evaluated by querying whether a specific object is present in the image.}
  \label{fig:desiderata_examples}
  \vspace{-0.3cm}
\end{figure}

\textbf{(1) Calibration.} 
It evaluates how the uncertainty about each MLLM's prediction is aligned with its accuracy, as highlighted by HELM~\citep{liang2022helm}. 
As shown in Figure~\ref{fig:desiderata_examples}, its \textit{instruction} is a standard query. 
Moreover, calibration is measured using the Expected Calibration Error (ECE)~\citep{Pakdaman_Naeini_Cooper_Hauskrecht_2015,guo2017calibration}, which calculates the difference between the model's predicted probability and the fraction of times the model is correct.

\textbf{(2) In-context Learning.}
It evaluates the crucial in-context learning (ICL) ability of an MLLM.
To evaluate this desideratum, the \textit{Instruction} is set to include randomly retrieved in-context examples (\texttt{ICE}). Note that \texttt{ICE} can include images. 
To assess the ICL ability, we introduce the {Relative ICL Accuracy for Multi-Choice QA} (RIAM), which measures the relative accuracy improvement beyond random guessing, written as
\begin{equation}
\text{RIAM} = (\text{acc}_\text{ICL} - \text{acc}_\text{0-shot})/(\text{acc}_\text{0-shot} - \text{acc}_\text{rand}),
\end{equation}
where $\text{acc}_\text{ICL}$ denotes the average accuracy among the results based on the instructions with different shots of in-context examples.
$\text{acc}_\text{0-shot}$ means zero-shot prediction without \texttt{ICE}. $\text{acc}_\text{rand}$ means the accuracy by random guessing.

\textbf{(3) Instruction Following.}
It evaluates how exactly the MLLM relies on the given instructions. The \textit{Instruction} is set as standard query, which is retrieved from the three categories of instructions as the way used in verbalizer manipulation, \emph{i.e.}, \emph{natural}, \emph{neutral}, and \emph{unnatural}~\citep{li2023instructionfollowing}.
The \textit{Metric} applied here is the Match Ratio (MR), which calculates the percentage of textual outputs that are matched with the outputs indicated by the verbalizer instructions.

\textbf{(4) Language Performance.}
It evaluates the quality of the generated sentences. Since the applied \textit{Inferencer} does not generate free-form output, we evaluate the language performance of the outputs corresponding to the chain-of-thought. Knowing that GPT-based metrics have shown to be well correlated with human evaluation~\citep{zheng2023judging,liu2023geval,wang2023chatgpt}, we use GPT-4 to evaluate the language performance of the \texttt{CoT} outputs based on the ground-truth sentences (\emph{i.e.} questions and answers) in the question-answering tasks.
%
%
Moreover, we choose the average results of multiple rounds of evaluations to eliminate the flickering of the GPT-based evaluations.

\textbf{(5) Robustness.}
It measures how robust an MLLM is to corruptions in the multimodal inputs.
The image corruptions include {noise, blur, weather, digital}~\citep{DBLP:journals/corr/abs-1903-12261} and common data augmentation techniques.
The textual corruptions include sentence-, word- and character-level corruptions~\citep{chen2023benchmarking}, as well as switching choices for multi-choice question-answering.

The \textit{Metric} in this desideratum is Relative Robustness for Multi-Choice (RRM), written as
\begin{equation}
    \text{RRM} = (\text{acc}_\text{crp} - \text{acc}_\text{rand})/(\text{acc} - \text{acc}_\text{rand}),
\end{equation}
where $\text{acc}_\text{crp}$ denotes the accuracy after corruption, $\text{acc}$ is the accuracy before corruption.
$\text{acc}_\text{rand}$ means the accuracy by random guessing.

\textbf{(6) Hallucination.}
It evaluates how an MLLM avoids mentioning visual objects that do not exist in the images.
The \textit{Scenario} is MSCOCO. 
We follow the Polling-based Object Probing Evaluation (POPE)~\citep{POPE} in this desideratum. It transforms hallucination evaluation into a set of binary classification tasks. Essentially, the MLLMs are posed Yes-or-No questions about the existence of some particular objects in the images, such as ``Is there a car in the image?'' Notably, \texttt{PPL} is applied to as a more reliable \textit{Inferencer}.
The \textit{Metric} applied here is accuracy.

\section{Experiments}
\vspace{-0.2cm}
\subsection{Evaluation Setup}
\label{sec:eval_setup}
\vspace{-0.2cm}
A wide range of recently introduced MLLMs are evaluated in ChEF, including LLaVA~\citep{llava}, LAMM~\citep{yin2023lamm}, MiniGPT4~\citep{minigpt4}, mPLUG-Owl (mPLUG)~\citep{ye2023mplugowl}, Otter~\citep{li2023otter}, InstructBLIP~\citep{dai2023instructblip}, LLaMA-Adapter-v2 (LAv2)~\citep{gao2023llamaadapterv2}, as well as models specifically designed for grounding tasks, such as Shikra~\citep{chen2023shikra} and Kosmos-2~\citep{kosmos-2}.
These MLLMs are evaluated across various single-task \textit{Scenarios}, including CIFAR-10 (CIFAR)~\citep{cifar10} for classification, Omnibenchmark (Omni)~\citep{Omnibenchmark} for fine-grained classification,  VOC2012 (VOC)~\citep{pascal-voc-2012} for object detection, FSC147 (FSC)~\citep{FSC147} for object counting, Flickr30k (Flickr)~\citep{flickr30k} for image captioning and ScienceQA (SQA)~\citep{scienceqa} for multimodal question-answering. 
We also evaluate the MLLMs on several multi-task datasets including MME~\citep{fu2023mme}, MMbench (MM)~\citep{liu2023mmbench}~\footnote{MMBench provides two evaluation settings (\emph{i.e.}, VanillaEval and CircularEval). VanillaEval is adopted in the default \textit{Recipe}.}, and Seedbench (SEED)~\citep{li2023seedbench}. 
%

\subsection{Standard Performance of Visual Ability}
\vspace{-0.2cm}

\begin{table}[t]
    \begin{center}
    \scriptsize
    \caption{\textbf{Visual performance of MLLMs on different Scenarios.} In SQA and MM, as options \{A, B, C, D\} are explicitly provided in the questions, models are required to output their answers in the form of options. Similarly, MME also requires models to provide ``yes'' or ``no'' outputs. These \textit{Scenarios} can be considered as a discriminative (discrim.) question type. Conversely, the other \textit{Scenarios} are characterized by generative (gen.) types, as they require responses without predefined options in questions. The abbreviations for \textit{Scenarios} and MLLMs are defined in section~\ref{sec:eval_setup}. For Omnibenchmark (Omni$^\dagger$), weighted accuracy is employed, which entails a weighted accuracy calculation based on the granularity of classification. The entries that are both bold and underlined indicate the best performance.}
    \vspace{-0.2cm}
    
    \begin{adjustbox}{width=\textwidth}
    \begin{tabular}{l|ccccccccc}
        \Xhline{1.5pt}
        \bf Scenario        &\bf CIFAR  &\bf Flickr   &\bf VOC     &\bf Omni$^\dagger$         &\bf FSC   &\bf SQA      &\bf MM     &\bf SEED     &\bf MME\\
        
        \bf Question Type   & gen.       & gen.         & gen.       & gen.            &gen.      &discrim.  &discrim.   & gen.  & discrim. \\
        \Xhline{1.5pt}
        \bf LLaVA           &\ul{89.40}  & 80.80        & 26.01      & 26.62           & 24.11    & 46.55       & 43.13     & 46.45       & 50.17\\
        \bf LAMM            & 80.70      & 72.50        & 29.58      & 22.54           & 19.33    & 52.75       & 44.47     & 47.03       & 55.82\\
        \bf MiniGPT-4       & 80.80      & 71.50        & 26.51      & 30.60           & 22.52    & 47.0        & 54.34     & 46.48       & 57.12\\
        \bf mPLUG           & 79.67      & 79.20        & 28.50      & 30.70           & 20.92    & 48.44       & 49.57     & 42.81       & 71.59\\
        \bf Otter           & 81.34      & 71.30        & 27.15      & 26.41           & 20.00    & 50.22       & 53.91     & 36.40       & 63.78\\ 
        \bf LAv2            & 70.17      & 79.50        & 31.60      & \ul{32.00}      & 21.26    & 54.34       & 57.06    & 35.41       & 69.90 \\
        \bf InstructBLIP    & 84.27      & 79.40        & 27.65      &30.75            &\ul{25.04}&\ul{55.18}   &\ul{65.73}  &\ul{50.81}   &\ul{72.0}\\
        \bf Shikra          & 68.71      &\ul{94.70}    &\ul{55.23}  & 22.89           & 22.43    & 45.21       & 63.26     & 49.79       & 70.28\\
        \bf Kosmos-2        & 88.87      & 85.70        & 54.55      & 21.34           & 21.93    & 34.60       & 32.82      & 46.38       & 52.95\\
        \hline
        \bf Random Choice   & 10.0       & 25.00        & 25.00      & 10.94           & 20.00    & 35.80       & 27.57      & 24.27       & 50.00\\  
        \Xhline{1.5pt}
    \end{tabular}
    \end{adjustbox}
    \label{tab:main results}
    \vspace{-0.5cm}
    \end{center}
\end{table}

For each \textit{Scenario}, we conduct various experiments with diverse \textit{Recipes}, from which, the \textit{Recipe} behaving most reliably (\emph{i.e.} stable to \textit{Instruction} variations) is selected as the default setting~\footnote{The default \textit{Recipe} is also demonstrated to display and approach the best performance of each MLLM, as shown in Figure~\ref{fig:stability_analysis}(a-b).} 
to evaluate the visual performance of all MLLMs, as shown in Table~\ref{tab:main results}.
As the default \textit{Recipes} incorporate \texttt{PPL}, which can be regarded as a multi-choice question-answering paradigm, we also provide the accuracy of random choice for each \textit{Scenario}. There are some observations as follows:

%
%

\textbf{(1)} InstructBLIP attains superior performance across most \textit{Scenarios}. It is worth noting that both Shikra and InstructBLIP showcase exceptional performance on the multi-task datasets, including MME, MMBench, and SEEDBench, while the performance of other models displays inconsistencies. The visual performance of these MLLMs exhibits strong trade-offs across different tasks.
\vspace{-0.05cm}

\textbf{(2)} All the MLLMs struggle in the object counting task (\emph{i.e.}~FSC), primarily due to the complexities associated with the precise identification of numerous objects within an image. 
\vspace{-0.05cm}


\textbf{(3)} There is a capability gap between detection and other tasks. Shikra and Kosmos-2 demonstrate remarkable detection capabilities, owing to their specialized training on detection datasets. However, Kosmos-2 exhibits limited aptitude in other \textit{Scenarios}, especially on MMBench and ScienceQA. Despite its ability to perform perception and reasoning tasks, Kosmos-2 struggles to comprehend the meaning of options \{A, B, C, D\} provided in the question, resulting in difficulty in aligning the answers to options. As a consequence, it exhibits lower performance on discriminative tasks.

The unified evaluation of these models on diverse \textit{Scenarios} in the ChEF enables us to conduct a fair comparison, discerning the optimal architectures and methodologies for specific \textit{Scenarios}.


\subsection{Results of Desiderata}
\vspace{-0.2cm}
\begin{figure}[t]
    \centering
    \includegraphics[width=\linewidth]{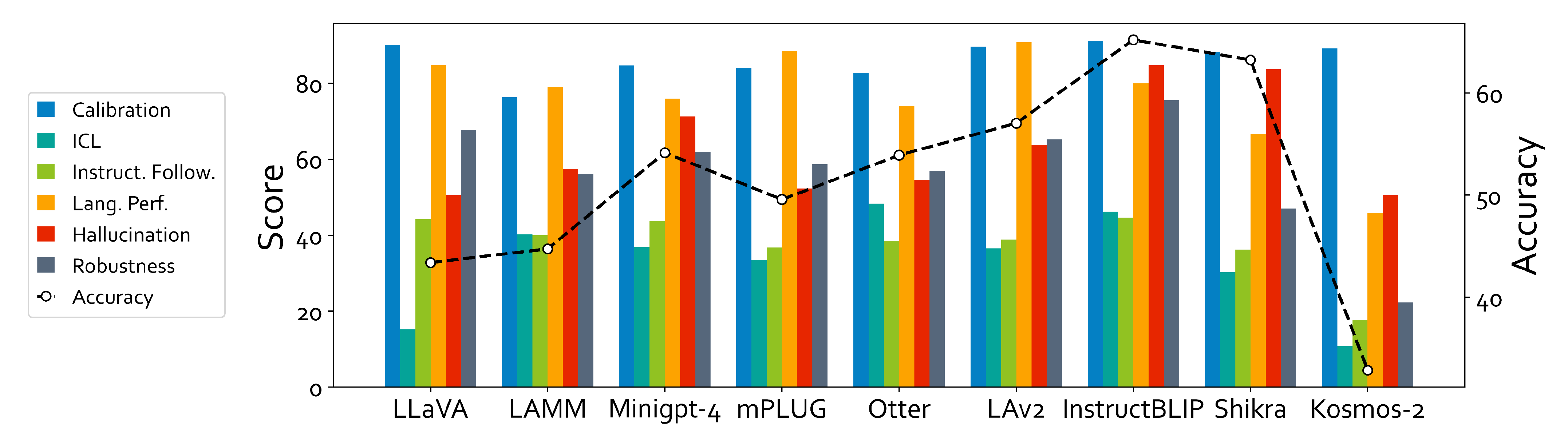}
    \vspace{-0.3cm}
    \caption{\textbf{Results of desiderata.} The dashline is the accuracy evaluated on MMBench. The score for each dimension is computed by normalizing the results from the specific metric to a range of 0-100. Calibration score is represented by 1-ECE. Instruction following score is the average MR across different verbalizer settings. In-context learning score is the average RIAM across various shot numbers. Language performance score is normalized from the results of the GPT-based metric. Robustness score is normalized from RMM and hallucination score directly represents accuracy.} 
  \vspace{-0.5cm}
  \label{fig:desiderata}
\end{figure}

The scores of all the desiderata on MLLMs are shown in Figure~\ref{fig:desiderata} with the corresponding accuracy of MMBench which we consider as the most representative assessment of MLLMs' visual performance. 
%
%
%
The six dimensions of desiderata are deemed essential for an MLLM to function as an interactive AI agent, emphasizing human-like interactions. However, the poor performance on these dimensions shows that current MLLMs fall short of being an AI agent capable of interacting with humans.

\textbf{(1)} Most MLLMs exhibit good calibration, indicating their ability to accurately convey uncertainty. This is primarily due to the relatively low accuracy of these models and their lack of confidence in the responses, which results in such consistency. 

\vspace{-0.05cm}

\textbf{(2)} Most MLLMs achieve satisfactory language performance, except for Kosmos-2, which provides few reasoning processes in its chain-of-thought responses. 
\vspace{-0.05cm}

\textbf{(3)} InstructBLIP and Shikra surpass other models on hallucination and meanwhile achieve superior visual performance on MMBench, emphasizing the crucial role of hallucination.
\vspace{-0.05cm}

\textbf{(4)} Most MLLMs exhibit poor performance in ICL. Notably, Otter, which is specifically trained on in-context instruction tuning data, though performs the best ICL among the 9 MLLMs, also struggles in ICL primarily due to its limited proficiency in visual tasks. 
\vspace{-0.05cm}

\textbf{(5)} Instruction following and robustness pose challenges for most MLLMs in effectively handling \textit{Instructions} that deviate from their priors and their susceptibility to noisy multimodal inputs.


\subsection{ChEF Provides Stable Assessment}
\vspace{-0.2cm}
\begin{wrapfigure}{l}{0.5\textwidth}
  \centering
  \includegraphics[width=\linewidth]{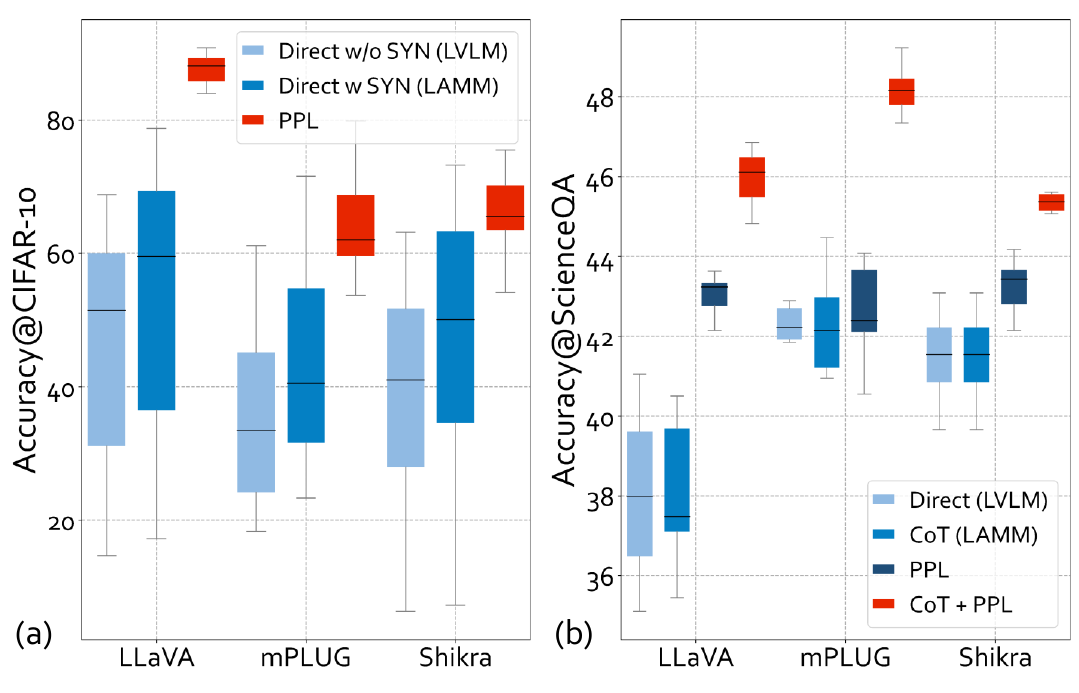}
  \caption{Results of various \textit{Inferencers} across different queries on CIFAR10 and ScienceQA. Black lines within each boxplot represent the median. Boxplots display the accuracy distribution.}
\label{fig:stability_analysis}
\end{wrapfigure}
\vspace{-0.3cm}

Due to the modular design of ChEF, it has the flexibility to employ different \textit{Recipes} for evaluating the same \textit{Scenario}. To get a reliable and fair evaluation, we conduct exhaustive experiments to identify the \textit{Recipe} that behaves more stable on \textit{Instruction} variations than previous approaches as the default setting. 


Two examples, shown in Figure~\ref{fig:stability_analysis}, are conducted on CIFAR10 and ScienceQA with distinct \textit{Recipes} for three MLLMs. 
Figure~\ref{fig:stability_analysis}(a) shows that utilizing \texttt{Direct} as \textit{Inferencer} proposed in LAMM~\citep{yin2023lamm} (with the inclusion of synonyms judgment in the metric) and LVLM~\citep{xu2023lvlmehub} (without synonyms) with different queries yields a large variance. Alternatively, employing the \texttt{PPL} can substantially mitigate these fluctuations with a much smaller variance, accompanied by a noteworthy gain in accuracy for all MLLMs. 
Similar observations can be also found in Figure~\ref{fig:stability_analysis}(b). We further leverage \texttt{CoT}, which mandates the model to provide its reasoning process. Although the accuracy has a slight gain, it does not bolster the stability. Nevertheless, the optimal combination of accuracy and stability emerges when employing both the \texttt{CoT} and \texttt{PPL} in a \texttt{Multi-Turn} \texttt{Inferencer}. 


Based on these interesting discoveries, we believe that ChEF, in conjunction with the meticulously derived and recommended \textit{Recipes} for diverse \textit{Scenarios}, can deliver a trustworthy and indicative assessment of MLLMs. We also conduct numerous experiments to carefully select appropriate \textit{Recipes} for reliable evaluations across the six dimensions of desiderata~\footnote{More evidence of reliability is provided in the Supplementary Materials (Section F).}.

\subsection{Correlation between Visual Performance and Desiderata}
\label{sec:correlation}
To investigate the relationship between visual performance and the desiderata, we display the Pearson correlation matrix in Figure~\ref{fig:correlation_analysis}(a). 
\vspace{-0.1cm}

\begin{wrapfigure}{r}{0.6\textwidth}
  \centering
  \includegraphics[width=\linewidth]{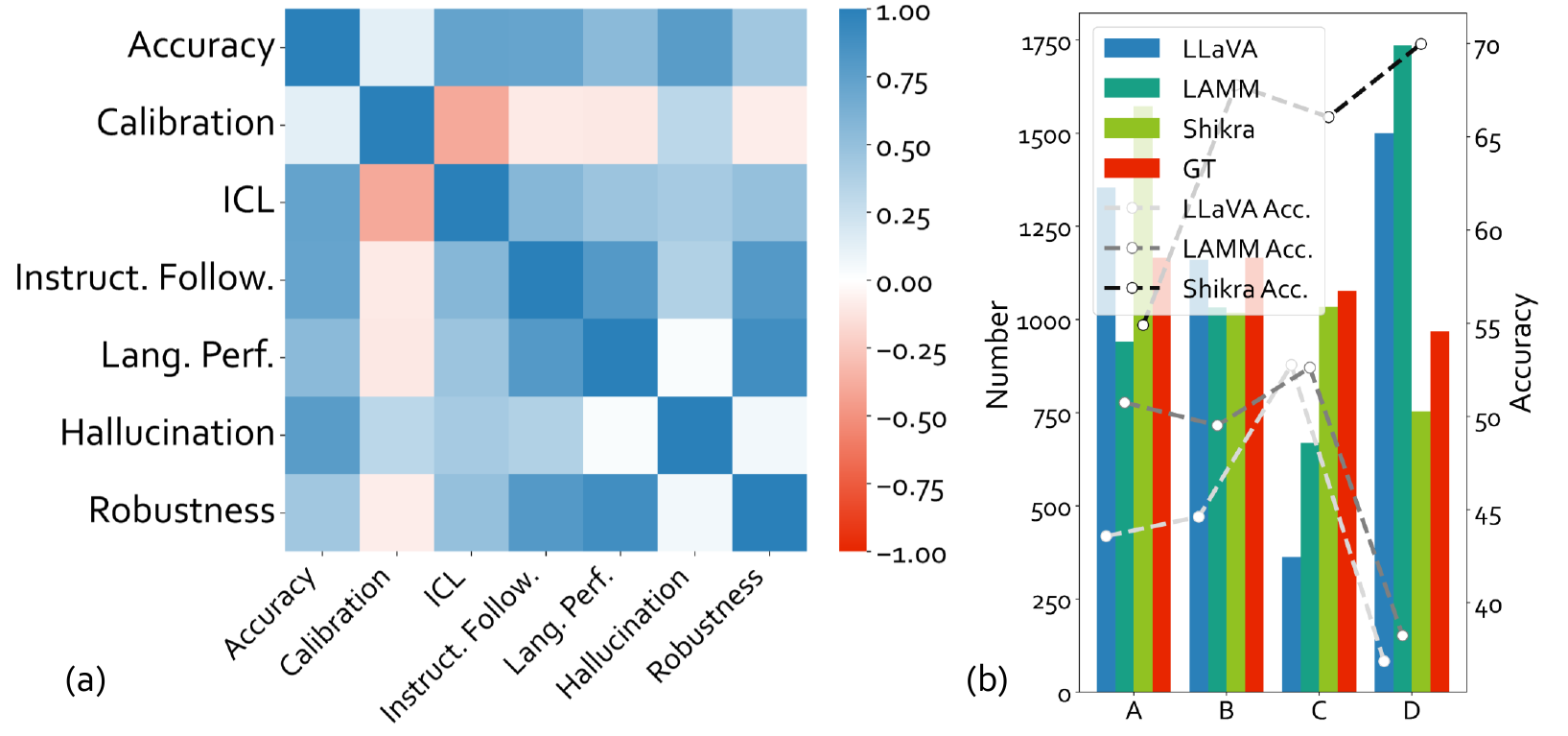}
  \vspace{-0.3cm}
  \caption{(a) Pearson correlation matrix of desiderata and accuracy on MMBench. Cooler colors indicate higher correlations. (b) Choice distribution with accuracy on MMBench. GT indicates the actual choice distribution.}
  \label{fig:correlation_analysis}
  \vspace{-0.3cm}
\end{wrapfigure}

\textbf{(1)} Calibration is an independent dimension, primarily assessing a model's proficiency in expressing uncertainty, without direct correlations to other dimensions. 
\vspace{-0.1cm}

\textbf{(2)} ICL demonstrates correlation with others, as their evaluations involve specific instructional aspects. MLLMs with enhanced ICL ability are better equipped to provide relevant responses to unseen cases.
\vspace{-0.1cm}
%

\textbf{(3)} Instruction following demonstrates a significant correlation with language performance, robustness, and accuracy. As language performance assesses the content of an MLLM's reasoning process, which is obtained through instructional guidance, MLLMs with stronger instruction following capabilities are more likely to adhere to the ``step by step'' instruction and generate a comprehensive reasoning process. 
\vspace{-0.1cm}
%

\textbf{(4)} Hallucination is strongly correlated with the performance on MMBench. The choice distribution of three models, as shown in Figure~\ref{fig:correlation_analysis}(b), reveals that LLaVA and LAMM prefer option D to C, while Shikra tends to favor option A over D. These MLLMs display lower accuracy on options they are inclined to answer and perform better on options that they resist. The distinct prior to options, which is caused by the hallucination issue, leads to poor performance. 
\vspace{-0.1cm}

It can be concluded that the evaluation of \textit{Scenarios} that involve discriminative questions evaluates a composite performance, \emph{i.e.}, visual performance, and additional dimensions of abilities, such as the comprehension of options. The evaluation of desiderata unveils intrinsic properties beyond visual performance.
%


\section{Conclusion}
\vspace{-0.2cm}
In this work, we introduce ChEF, a comprehensive evaluation framework for holistically profiling and comparing MLLMs. ChEF's modular design (\emph{i.e.} \textit{Scenario}, \textit{Instruction}, \textit{Inferencer}, and \textit{Metric}) enables versatile evaluations in a standardized framework. Based on ChEF, any evaluation, including current MLLM benchmarks, can be summarized as \textit{Recipes} of ChEF. We further introduce recipes to assess MLLMs' six dimensions of desiderata and conduct large-scale experiments to test the generalizability of MLLMs across various scenarios and their composite capability for multimodal interactions.

\textbf{Limitations.} As one of the pioneering works in this domain, our study has certain limitations. Firstly, ChEF is still in its nascent stage, currently supporting only a limited number of \textit{Scenarios} and models. For instance, \textit{Scenarios} evaluating safety and biases have not been incorporated yet. As we move forward, we aim to include a wider array of \textit{Scenarios} and other models to further enrich and expand the framework's applicability and comprehensiveness. Secondly, there remains a discernible performance variance among models when confronted with different queries. While our provided \textit{Recipes} have significantly mitigated these disparities, such variations are inevitable. Further research is needed to more accurately assess and optimize model performances across diverse queries to achieve more consistent evaluation outcomes.  Furthermore, the utilization of the GPT API for evaluation remains an area where the effectiveness has not been conclusively determined. We will continue to stay updated with the latest advancements in the field and leverage the scalability of ChEF to optimize and update accordingly.

\bibliographystyle{ChEF}
\bibliography{ChEF}

\appendix
\newpage

\startcontents[main]
\printcontents[main]{}{1}{}
\section{Related Works}

\subsection{Multimodal Large Language Models}
Due to the success of large Language models (LLMs) like GPTs~\citep{radford2019language,brown2020language,ouyang2022training},  LLAMA~\citep{touvron2023llama} and Vicuna~\citep{chiang2023vicuna}, Multimodal Large Language Models (MLLMs) have recently experienced substantial development. InstructBLIP~\citep{dai2023instructblip}, LLaVA~\citep{llava}, and MiniGPT-4~\citep{minigpt4} are based on open-source LLMs using vision-language instruction tuning get promising results. mPLUG-Owl~\citep{ye2023mplugowl} leverages the capabilities of pre-trained LLMs, a visual knowledge module, and a connected visual abstractor module to effectively align images with text. LAMM~\citep{yin2023lamm} extend the research of MLLMs to point clouds and propose a training framework optimized for modalities' extension. Otter~\citep{li2023otter} utilizes multimodal context instruction tuning data, demonstrating an improved ability to follow instructions and in in-context learning. LLaMA-Adapter-v2~\citep{gao2023llamaadapterv2} propose an early fusion strategy to solve the interference between image-text alignment and instruction following learning targets.
Shikra ~\citep{chen2023shikra} and Kosmos-2~\citep{kosmos-2} integrate grounding data during the training phase, enabling the model to develop grounding capabilities. In order to comprehensively assess the capabilities of these MLLMs, we present the first Comprehensive Evaluation Framework (ChEF) that can holistically profile each MLLM and fairly compare different MLLMs.

\subsection{Benchmarks for Large Language Models}
In recent years, significant efforts have been made to comprehensively evaluate large language models from diverse perspectives~\citep{liang2022helm,wang2023pandalm,bommasani2021opportunities,gehrmann2021gem,gehrmann2022gemv2,brown2020language,gao2021framework,von2022evaluate,srivastava2022beyond}.
\cite{gao2021framework} provides a unified framework to test autoregressive language models on a large number of different evaluation tasks.
\citet{liang2022helm} measures seven metrics that reflect a range of societal considerations, including accuracy, calibration, robustness, fairness, bias, toxicity, and efficiency, in order to improve the transparency of language models. \citet{li2023instructionfollowing} 
propose to evaluate the instruction following ability from the aspect of how well models can follow instructions that may not align with their priors. Recent studies evaluating the quality of natural language generation~\citep{zheng2023judging,liu2023geval,wang2023chatgpt} have indicated that GPT-based metrics typically exhibit superior performance compared to traditional reference-based and reference-free baseline metrics in terms of their correlation with human quality judgments. These evaluation metrics effectively assess the capabilities of LLMs from multiple dimensions. However, in the evaluation of MLLMs, there is currently a lack of frameworks and relevant metrics. These frameworks and metrics are of significant importance in assessing MLLMs.

\subsection{Benchmarks for Multimodal Large Language Models}
MLLMs have demonstrated remarkable capabilities~\citep{llava,minigpt4,dai2023instructblip} and are poised to address increasingly complex multimodal tasks. Various benchmarks have emerged to evaluate MLLMs.
Some works focus on evaluating MLLMs using existing conventional multimodal datasets~\citep{wang2023visionllm} or only evaluate one or a few factors of MLLMs~\citep{shao2023tinylvlm,POPE,yu2023mmvet,bitton2023visitbench}, which may not provide a comprehensive evaluation suitable for these models.
Recent benchmarks~\citep{li2023seedbench,liu2023mmbench,fu2023mme} often focus on building a multimodal evaluation dataset for MLLMs. These benchmarks have been designed to transform open-ended predictions into predefined categorical choices. For instance, MME transforms free-form responses into binary True/False questions, while ~\citet{li2023seedbench,liu2023mmbench} employ multi-choice questions. However, the efficacy of these benchmarks is contingent upon the quality of the dataset construction and may suffer from scalability issues.
More recently, efforts such as ~\cite{yin2023lamm,xu2023lvlmehub} have attempted to establish evaluation frameworks, yet they have been characterized by limitations in terms of scalability and comprehensiveness. In response to these challenges, ChEF offers a standardized framework for conducting versatile evaluations and facilitates seamless integration of new models and tasks.

\section{ChEF (Comprehensive Evaluation Framework) Modules}
ChEF is a comprehensive evaluation framework aiming at providing a fair and holistic assessment of MLLMs' performance across diverse multimodal tasks. To accomplish this objective, our design principles encompass the following key aspects: Modular, Scalable, Flexible, Reliable, and Indicative.
Based on these principles, we carefully design and implement ChEF with four components \emph{i.e.}, \textit{Scenario}, \textit{Instruction}, \textit{Inferencer}, and \textit{Metric}. In this section, we will introduce the details of each module.

\subsection{Scenario}

The \textit{Scenario} pertains to the datasets and tasks utilized for evaluating the proficiency of MLLMs in visual and multimodal tasks. Following the principles, the \textit{Scenario} is designed to be scalable. Any \textit{Scenario} can be easily integrated into ChEF by defining the required \textit{Instruction} and \textit{Metric} with the provided interfaces. 
Due to the substantial similarities among datasets within the same visual task, we categorize them based on task divisions. Within each task, the \textit{Scenarios} can share similar implementations for the given interfaces.

To facilitate the active participation of the open-source community in expanding the scope of \textit{Scenarios}, we incorporate several prominent datasets from highly regarded visual tasks as exemplary \textit{Scenarios}. These datasets include CIFAR-10~\citep{cifar10} for classification, Flickr30k~\citep{flickr30k} for image captioning, ScienceQA~\citep{scienceqa} for multimodal question-answering, \emph{etc.} Furthermore, we seamlessly integrate multi-task datasets, including MMbench~\citep{liu2023mmbench}, SeedBench~\citep{li2023seedbench}, and MME~\citep{fu2023mme}, into the framework of ChEF. We warmly welcome the integration of additional \textit{Scenarios} into ChEF by simply implementing the requirements with the provided interfaces.

\subsection{Instruction}
\label{sec:Instruction}

\begin{figure}[tbp]
    \centering
    \includegraphics[width=\textwidth]{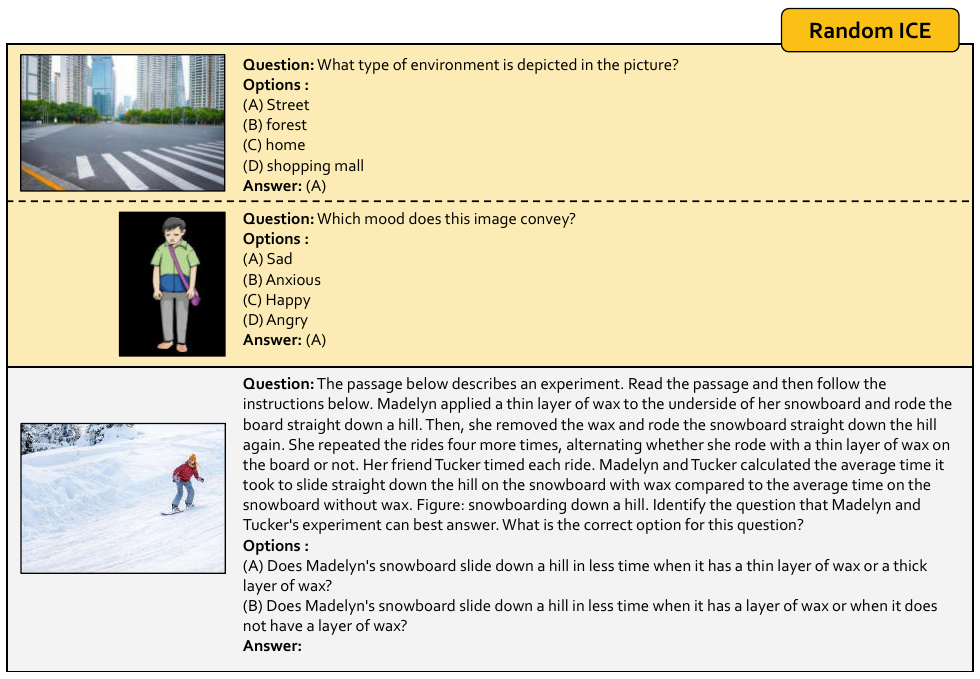}
    \caption{\textbf{An example of Random ICE.} The Random \texttt{ICE} are randomly retrieved from the dataset, without considering their relevance or importance.}
  \label{fig:retriever_random}
\end{figure}

\begin{figure}[htbp]
    \centering
    \includegraphics[width=\textwidth]{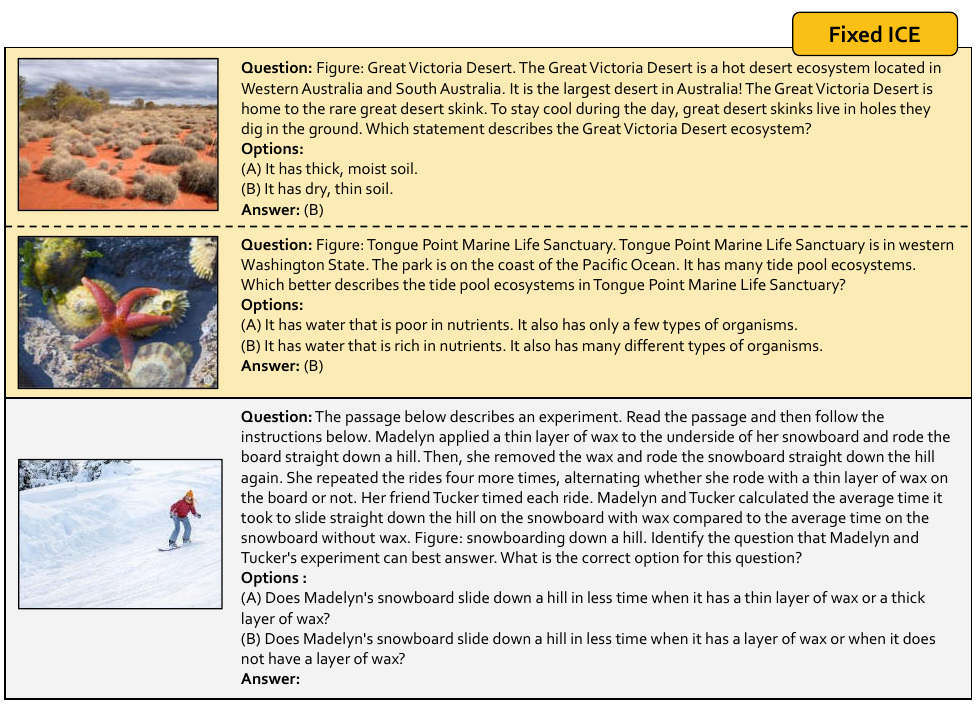}
    \caption{\textbf{An example of Fixed ICE.} The Fixed \texttt{ICE} is predetermined based on prior knowledge or experiment.}
  \label{fig:retriever_fixed}
\end{figure}

\begin{figure}[htb]
    \centering
    \includegraphics[width=\textwidth]{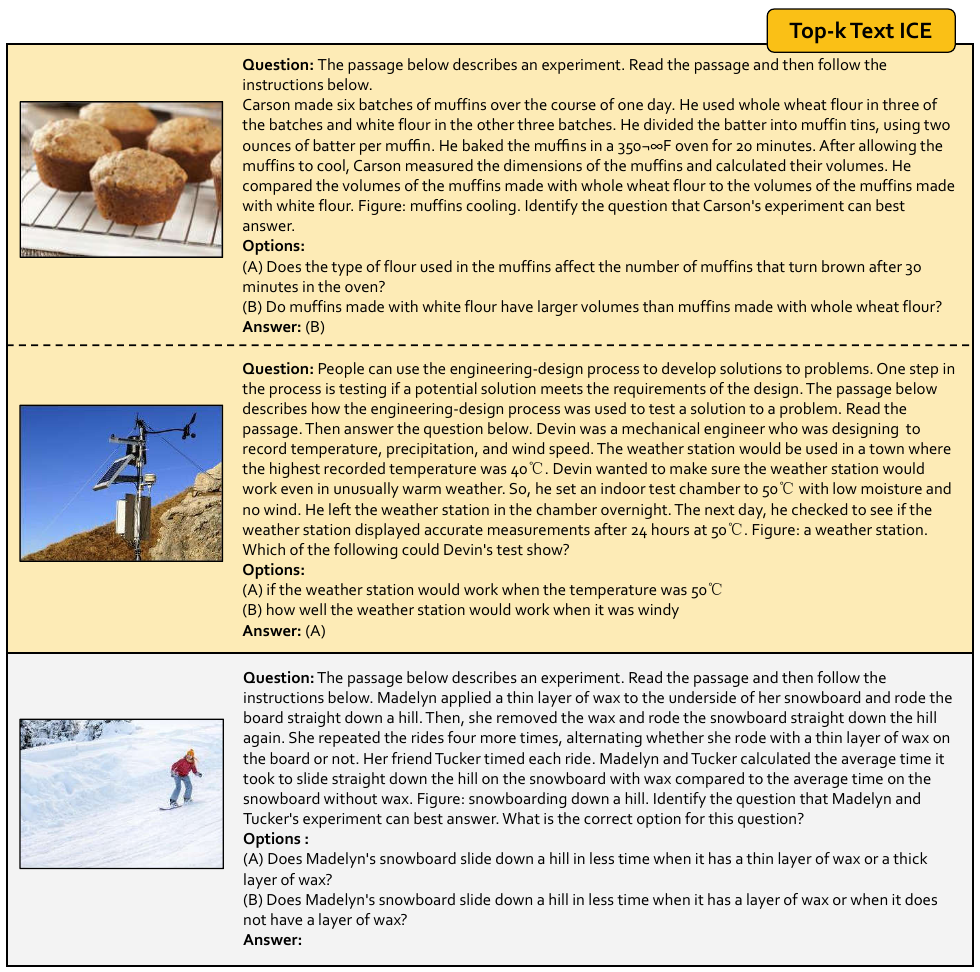}
    \caption{\textbf{An example of Top-$k$ Text ICE.} The Top-$k$ Text \texttt{ICE} is retrieved from the dataset based on text similarity. }
  \label{fig:retriever_topk_text}
\end{figure}

\begin{figure}[htbp]
    \centering
    \includegraphics[width=\textwidth]{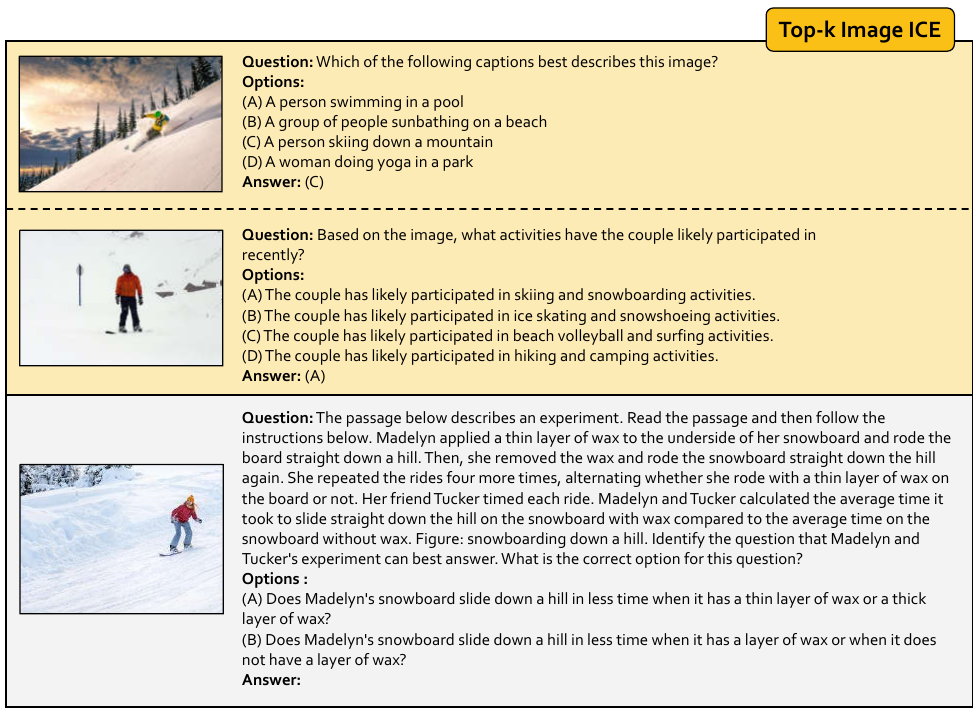}
    \caption{\textbf{An example of Top-$k$ Image ICE.} The Top-$k$ Image \texttt{ICE} is retrieved from the dataset based on image similarity. }
  \label{fig:retriever_topk_image}
\end{figure}

The \textit{Instruction} component plays a pivotal role in facilitating the model's comprehension of the underlying semantics within the \textit{Scenario} and generating pertinent responses. Within ChEF, a standard query is initially incorporated for each \textit{Scenario}, such as ``The photo of'' for classification, providing the model with a basis for answer generation. Nevertheless, it is noteworthy that divergent models may interpret the same query dissimilarly, leading to variations in evaluation. 

To ensure the universal compatibility of the \textit{Instruction} module, in line with the design principle of flexibility, we undertake measures to devise the query pool, encompassing frequently employed queries that exhibit similar intents. This designation allows for the seamless integration of new queries, thereby ensuring the requisite adaptability for different MLLMs. The standard query and query pool are collectively referred to as \texttt{Query}.

Moreover, we firmly believe that leveraging the In-context Example (\texttt{ICE}) as the \textit{Instruction} presents a more comprehensive and generalized approach, empowering models to grasp the intricacies of the assigned task and generate responses in the desired format and content. The \texttt{ICE} is retrieved from the dataset based on various criteria commonly employed in the field of NLP, including Random \texttt{ICE}, Fixed \texttt{ICE}, and Top-$K$ \texttt{ICE} ~\citep{wu2023openicl,iclref1,iclref2}. 

\textbf{(1) Random ICE} is retrieved at random, without considering their relevance or importance. An example is shown in Figure~\ref{fig:retriever_random}.

\textbf{(2) Fixed ICE} is predetermined based on prior knowledge or experiments. 
These \texttt{ICE} can serve as instructional cues to encourage the model to replicate and generate outputs in a format consistent with the provided examples, as shown in Figure~\ref{fig:retriever_fixed}

\textbf{(3) Top-$k$ ICE} is retrieved based on either the image similarity (Top-$k$ Image \texttt{ICE}) or the text (Top-$k$ Text \texttt{ICE}) similarity, as shown in Figure~\ref{fig:retriever_topk_text},\ref{fig:retriever_topk_image}.

The designation and implementation of the \texttt{Query} and \texttt{ICE} significantly contribute to the flexibility of evaluation.

\subsection{Inferencer}
The \textit{Inferencer} plays a vital role in determining the model's response to questions.
Within ChEF, it incorporates a fundamental auto-regressive generation method. However, due to the free-form and long-term nature of its output, evaluating the quality of the generated text becomes subjective and unreliable~\citep{yin2023lamm,li2023seedbench}. To address this concern, we design the following \textit{Inferencers} to support reliable evaluation:

\textbf{(1) Direct:} This is an auto-regressive generation method employed without sampling. The output of the MLLMs is determined through greedy search, ensuring consistent output across multiple inference instances for enhanced reliability.

\textbf{(2) Chain-of-Thought (CoT):}  This answering approach includes a special query, ``Let's think step by step”, which prompts the model to provide responses in a sequential manner. It prompts the model to provide its reasoning process, ensuring that the model's answers are well-thought-out and dependable.

\textbf{(3) Perplexity (PPL):} This \textit{Inferencer} constrains MLLMs' output within a limited text scope, named as answer pool, and derives the answer by computing the likelihood. The answer pool is either fixed, retrieved, or generated based on the specific \textit{Scenario}. For example, in multi-choice question-answering \textit{Scenarios}, the answer pool is the four options \{A, B, C, D\}. For certain \textit{Scenarios}, it includes the ground-truth answer and several negative candidates either generated or retrieved. 
\texttt{PPL} confines the model's output within a specific range, guaranteeing that the model selects exactly matched answers based on discrimination rather than generating similar responses. Treating MLLMs as discriminative entities for specific \textit{Scenario} evaluation enhances objectivity and reliability in the evaluation process.

\textbf{(4) Multi-Turn:} This method decomposes complex tasks into subtasks and generates answers sequentially based on each subtask. For example, in the context of object detection, the initial \textit{Instruction} may pertain to the object categories present in the image, followed by subsequent inquiries regarding the bounding boxes for each detected object category. This approach supports objective and reliable evaluation by assessing the model's responses to each subtask, thereby enhancing objectivity and reliability. Notably, various \textit{Inferencers} can be invoked and seamlessly integrated with one another within multiple turns. For illustration, the \texttt{CoT} can be employed during the initial turn, while the subsequent turn can leverage the \texttt{Direct}.

These \textit{Inferencers} augment the evaluation framework of ChEF, enabling more objective and trustworthy assessments of model performance.

\subsection{Metric}

The selection of \textit{Metric}s is crucial when evaluating MLLMs, as it should encompass the evaluation capabilities for traditional visual tasks while considering the novel characteristics of MLLMs as generative models. In the context of traditional computer vision tasks, we believe it is more suitable to conduct adaptation based on the existing evaluation metrics. As a result, within the ChEF framework, we integrate well-established metrics such as BLEU for captioning, accuracy for classification, and mAP for detection, which are commonly used in traditional computer vision tasks. 

Additionally, when employing the \texttt{PPL} as \textit{Inferencer} in evaluation pipelines, we rely on accuracy as the primary \textit{Metric} since the generated text is confined to an answer pool. This methodology enables the harmonization of evaluation across various \textit{Scenarios}, as accuracy is adopted as the shared assessment criterion.

\section{Desiderata}
\label{sec:desiderata_sup}
Based on ChEF, it becomes rather convenient to set up new evaluations to quantify the desired capabilities (or called \textbf{desiderata}) that a competent MLLM model should possess, as a reliable agent that can perform real-world multimodal interactions. The desiderata include calibration, in-context learning, instruction following, language performance, hallucination, and robustness. In this section, we will introduce the details of each desideratum.

\subsection{Calibration} 
Calibration aims to evaluate the model's performance to be simultaneously accurate and to provide appropriate uncertainty in its outputs, as emphasized in the work by HELM~\citep{liang2022helm}. This is particularly significant in risk scenarios  We evaluate calibration by Expected Calibration Error (ECE)~\citep{Pakdaman_Naeini_Cooper_Hauskrecht_2015,guo2017calibration}. Formally, let $y$ be the ground truth, and $\hat{y}$ be the model's prediction with associated confidence $\hat{p}$. The ECE examines the
difference between the model’s predicted confidence $\hat{p}$ and the probability the model is correctly given $\hat{p}$, as shown in equation~\ref{equ:calibration_1}.
\begin{equation}
    \mathrm{ECE}=\mathbb{E}  [| \hat{p}-\mathbb{E} (y=\hat{y}|\hat{p} )| ] 
    \label{equ:calibration_1} 
\end{equation}
To estimate the expected accuracy $\mathbb{E} (y=\hat{y}|\hat{p} )$ from finite samples, we compute the ECE by binning the model’s predictions into $m$ bins following prior work~\citep{guo2017calibration,liang2022helm}. We choose uniform-mass bins for better approximation with $k=10$, where an equal number of samples fall into each bin. Let $\mathcal{B}_m$ be a set of indices $i$ of samples falling in $m$-th bin, then the average confidence and accuracy of $\mathcal{B}_m$ are defined as 
\begin{align}
&\operatorname{conf}\left(\mathcal{B}_{m}\right)=\frac{1}{\left|\mathcal{B}_{m}\right|}\sum_{i\in \mathcal{B}_{m}}^{}\hat{p}_i  \label{equ:calibration_2} \\
&\operatorname{acc}\left(\mathcal{B}_{m}\right)=\frac{1}{\left|\mathcal{B}_{m}\right|}\sum_{i\in \mathcal{B}_m}^{}\mathbf{1}(\hat{y}_i=y_i )  \label{equ:calibration_3} 
\end{align}
Therefore, we can approximates equation ~\ref{equ:calibration_1} by equation~\ref{equ:calibration_4}.
\begin{equation}
\mathrm{ECE} =\sum_{m=1}^{k} \frac{\left|\mathcal{B}_{m}\right|}{n}\left|\operatorname{conf}\left(\mathcal{B}_{m}\right)-\operatorname{acc}\left(\mathcal{B}_{m}\right)\right|
\label{equ:calibration_4}
\end{equation}
The difference between $\operatorname{conf}$ and $\operatorname{acc}$ for a given bin represents the calibration gap (visualized in Figure~\ref{fig:calibration} ). The lower the ECE, the better the calibration of the model, indicating that the predicted confidence $\hat{p}$ more accurately represents the true probability.

\subsection{In-context Learning} \label{sec:icl}
\begin{figure}[htb]
    \centering
    \includegraphics[width=\textwidth]{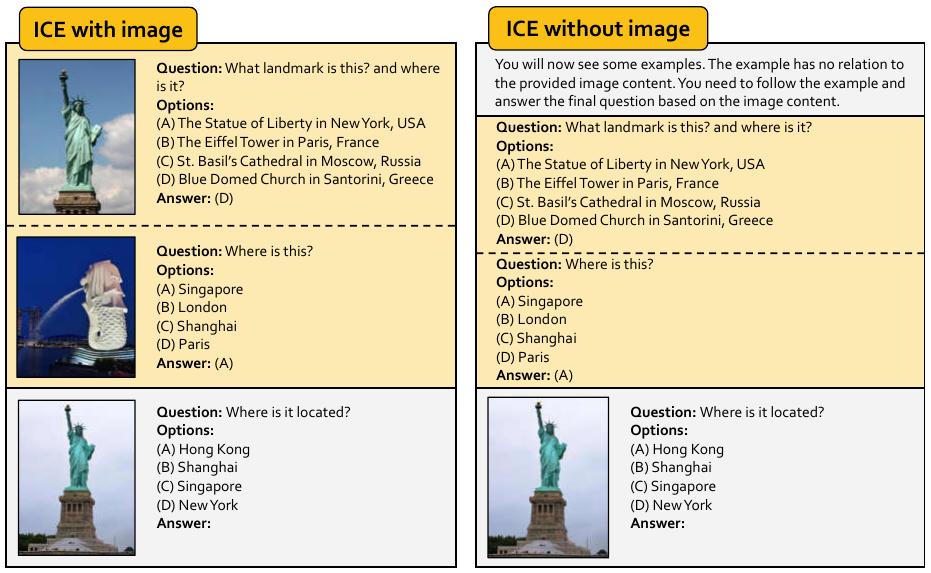}
    \caption{\textbf{Difference between ICE with image and without image.} The \texttt{ICE} are retrieved based on the images' similarity to the input images.}
  \label{fig:icl_ins_imgtext}
\end{figure}

In-context Learning (ICL) aims to evaluate MLLMs' ability to perform new tasks without any gradient-based training~\citep{wu2023openicl, brown2020language}. This ability is capable of generalizing to unseen cases, which opens up many new technological possibilities that were previously considered unique to humans. While in the field of NLP, LLMs have demonstrated their ability for ICL. However, within the domain of MLLMs, this potential remains unexplored. Most MLLMs lack the ability for ICL~\citep{li2023otter}. Therefore, considering the ICL ability is crucial when evaluating multimodal large language models.

ICL adds a small number of \texttt{ICE} before \texttt{Query} as the \textit{Instruction} and has demonstrated its ability to enhance the performance of LLMs in few-shot scenarios. Given that multimodal tasks typically involve visual data, incorporating the \texttt{ICE} with images in MLLMs is a reasonable approach. However, some MLLMs currently only support single-image input. Given the presence of an image in the \texttt{Query}, the image of \texttt{ICE} cannot be included. Considering the limited support for multi-image input in certain MLLMs, we implement two ICL methodologies: one utilizing \texttt{ICE} without image and the other incorporating \texttt{ICE} with images, as shown in Figure~\ref{fig:icl_ins_imgtext}. In the case of \texttt{ICE} without image, to prevent any confusion between the content of \texttt{ICE} and the images in the \texttt{Query} for the MLLMs, we add an additional \textit{Instruction}, explicitly informing the MLLMs that the provided \texttt{ICE} text has no relation to the provided image content. For the selection of \texttt{ICE}, we implement retriever methods such as Random, Fixed, and Top-$k$, as mentioned in Section~\ref{sec:Instruction}.

To measure MLLMs' ICL ability, we utilize \texttt{ICE} as \textit{Instruction} for each specific \textit{Scenario}. We compute their accuracy and use the relative accuracy change as the final score. Specifically, we compute the accuracy under the 0-shot setting (without using \texttt{ICE}) and the average accuracy values for varying numbers of \texttt{ICE}, ranging from $1$ to $N$. In multi-choice question-answering paradigms, random guessing can yield an expected lower-bound accuracy, which can be misleading in terms of performance evaluation. To mitigate the impact of this potentially deceptive performance on robustness assessments, we systematically eliminate the bias introduced by random choice. Therefore, we introduce the Relative ICL Accuracy for Multi-choice (RIAM), adapted from \citet{chen2023benchmarking, schiappa2023robustness}, to more accurately assess the model's ICL ability. The RIAM primarily calculates the relative accuracy change of the model before and after using \texttt{ICE}.


\subsection{Instruction Following} 
Taking inspiration from \citep{li2023instructionfollowing}, we utilize three groups of instructions for verbalizer manipulation: \textit{natural, neutral, unnatural}, to evaluate how well models can follow instructions that may not align with their priors. The levels in terms of aligning with prior knowledge of these three groups are ranked as \textit{natural} $>$ \textit{neutral} $>$ \textit{unnatural}. We expect the model to answer the question following instructions and generate a new answer corresponding to the original answer. In practice, we select different numbers of verbalizers for each group of verbalizer manipulation, depending on the alignment with the model's prior knowledge. Each verbalizer maps ``A\textbar B\textbar C\textbar D" to different new options.

\textbf{(1) Natural.}``1\textbar2\textbar3\textbar4\textbar5" ,``I\textbar II\textbar III\textbar IV\textbar V" and ``first\textbar second\textbar third\textbar fourth\textbar fifth".

\textbf{(2) Neutral.} ``Smith\textbar Johnson\textbar Williams\textbar Jones\textbar Brown" and ``foo\textbar dog\textbar hip\textbar oh\textbar cat".

\textbf{(3) Unnatural.} The choices are mapped to their respective next choices as the new verbalizer for each given question (e.g., ``D\textbar A\textbar B\textbar C" corresponding to ``A\textbar B\textbar C\textbar D"). 

We calculate the Match Ratio (MR) to determine the percentage of samples that adhere to the verbalizer manipulation instructions, mapping their original answers to corresponding new answers. This calculation helps mitigate the influence of the model's accuracy in answering questions and highlights its proficiency in following verbalizer manipulation instructions. A higher MR indicates a superior ability of the model to follow instructions.

\subsection{Language Performance} \label{sec:language_performance}
\begin{figure}[htbp]
  \centering
    \includegraphics[width=\textwidth]{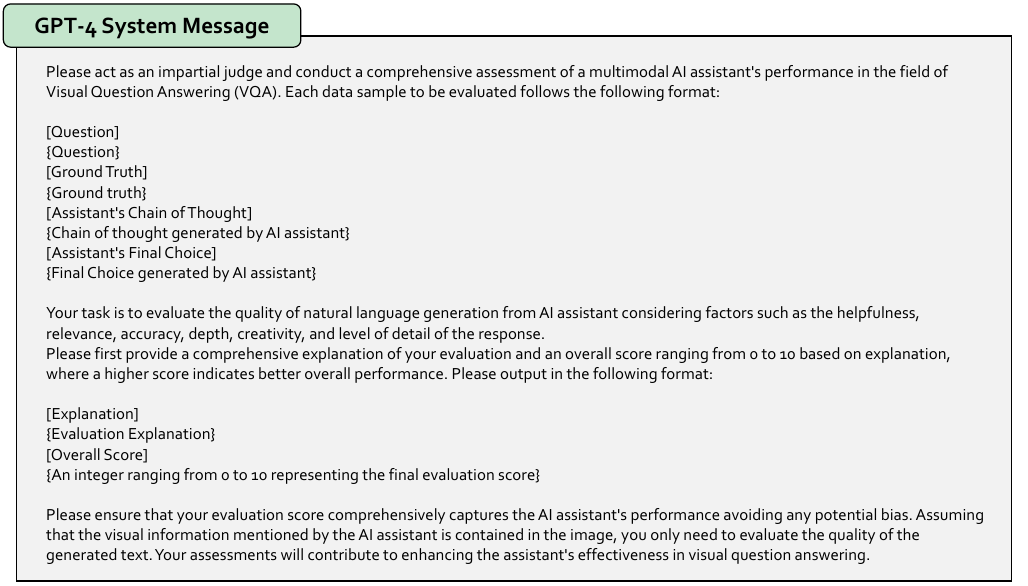}
    \caption{\textbf{System message for GPT-4 }to evaluate language performance of MLLMs. The System Message includes the evaluation task description, the format of the evaluation input template, the evaluation criteria, and the format of the evaluation output template. The phrases enclosed in ``[]'' represent domain names, which remain constant during the testing process. The phrases enclosed in ``\{\}'' represent the meanings of the domain names, which is a placeholder to be replaced with the specific content corresponding to the domain name during testing.}
      \label{fig:gpt_system}
\end{figure}
Evaluating the quality of natural language generation is a challenging task, often requiring scoring based on various aspects such as coherence, consistency, fluency, and more. Recent studies~\citep{zheng2023judging,liu2023geval,wang2023chatgpt} have indicated that GPT-based metrics typically exhibit superior performance compared to traditional reference-based and reference-free baseline metrics in terms of their correlation with human quality judgments. Thus, we employ GPT to score the chain-of-thought text generated by the model in the multimodal question-answering \textit{Scenarios}, aiming to evaluate the model's language performance.

In contrast to NLP, where GPT can evaluate the quality of natural language generation without references~\citep{zheng2023judging,liu2023geval,wang2023chatgpt}, the evaluation process in the visual \textit{Scenarios} presents a distinct challenge as GPT lacks access to visual information. Therefore, we implement specific adaptations for evaluating GPT's performance in multimodal tasks as follows: 

\textbf{(1) Reference-Based Evaluation:} We provide GPT with ground-truth sentences (\emph{i.e.} answers and questions) as the reference during the evaluation, which ensures faithfulness of the chain-of-thought.

\textbf{(2) Visual Information Assumption:} GPT is prompted to assume that all visual information mentioned in the test model's responses is contained in the image. This measure prevents GPT from misjudging descriptions of images in the chain-of-thought as language hallucinations (which may not be explicitly stated in the given question). This helps avoid unwarranted reductions in the language performance score.

\textbf{(3) Selective Sampling of Correct Conclusions:} We selectively extract samples in which the MLLMs' conclusions are correct. This reduces the impact of conclusion accuracy on the evaluation of language generation quality, as mentioned in Section~\ref{sec:language_performance_exp}. 

\textbf{(4) Efficient and Scalable Evaluation}: For more efficient and scalable evaluation, instead of pairwise comparisons, we individually assess each MLLM's response, which is called Single Answer Grading. This method exhibits high agreement with human experts in NLP tasks as demonstrated in ~\citep{zheng2023judging}.

\textbf{(5) Multiple Evidence Calibration:}~\citep{wang2023large} To make the GPT score more reliable and interpretable, we prompt the GPT to generate an explanation as evaluation evidence before generating the final overall score. Thanks to the properties of autoregressive models, this method allows GPT to calibrate scores based on evaluation evidence. To further reduce the systematic error of GPT evaluation, we conduct Multiple Evidence Calibration, sampling multiple GPT responses for each evaluation query, and taking the average score of all responses as the final evaluation score. 

To apply the adaptations below, we modify the system message for GPT-4. Figure~\ref{fig:gpt_system} shows the system message for GPT-4 to evaluate the language performance of MLLMs.

\subsection{Robustness} 

\begin{table}[htbp]
    \centering
    
    \caption{\textbf{Image corruption methods} are categorized into five types. In the robustness experiments, the corruption for each image is formed by sequentially combining methods each with random severity level from the following five categories: \textit{Noise}, \textit{Blur}, \textit{Weather}, \textit{Digital}, and \textit{Other}. Each category's method is selected based on the corresponding combination strategy: \textit{Random} denotes the random selection of one method from all methods within that category, while \textit{Sequential} implies the consecutive execution of all methods within that category. Severity represents the number of adjustable severity levels for the corruption method.}
    \small
    \begin{tabular}{c|c|c|c}
    
        \Xhline{1.5pt}
        \textbf{Category}        & \textbf{Method}    & \textbf{Severity} & \textbf{Compose Strategy}\\ 
        \Xhline{1.5pt}
        \multirow{4}{*}{\bf Noise}   & Gaussian Noise     & 5             & \multirow{4}{*}{Random}     \\
                                 & Shot Noise         & 5                &  \\
                                 & Impulse Noise      & 5                &  \\
                                 & Speckle Noise      & 5                &  \\ 
        \hline
        \multirow{5}{*}{\bf Blur}    & Defocus Blur       & 5           & \multirow{5}{*}{Random}       \\
                                 & Frosted Glass Blur & 5               &   \\
                                 & Motion Blur        & 5                &  \\
                                 & Zoom Blur          & 5                &  \\
                                 & Gaussian Blur      & 5                 & \\ 
        \hline
        \multirow{5}{*}{\bf Weather} & Snow               & 5             & \multirow{5}{*}{Random}     \\
                                 & Frost              & 5                &  \\
                                 & Fog                & 5                &  \\
                                 & Brightness         & 5                 & \\
                                 & Spatter            & 5                 & \\
        \hline
        \multirow{5}{*}{\bf Digital} & Contrast           & 5            & \multirow{5}{*}{Random}     \\
                                 & Elastic            & 5                  \\
                                 & Pixelate           & 5                  \\
                                 & JPEG Compression   & 5                  \\
                                 & Saturate           & 5                  \\
        \hline
        \multirow{3}{*}{\bf Other}   & Center Crop        & 5           & \multirow{3}{*}{Sequential}        \\
                                 & Resize             & 5                  \\
                                 & Rotate             & 5                  \\ 
        \Xhline{1.5pt}
     \end{tabular}
     \label{tab:rob_ic}  
\end{table}

\begin{table}[htbp]
    \centering
    \caption{\textbf{Text corruption methods} are categorized into five types. In the robustness experiments, the corruption for each text is formed by sequentially combining methods each with random severity level from the following five categories: \textit{Basic}, \textit{Sentence}, \textit{Word}, \textit{Character}, and \textit{Choice}. Each category's method is selected based on the corresponding combination strategy: \textit{Random} denotes the random selection of one method from all methods within that category, while \textit{Sequential} implies the consecutive execution of all methods within that category. Severity represents the number of adjustable severity levels for the corruption method. }
    \small
    \begin{tabular}{l|l|c|c}
        
        \Xhline{1.5pt}
        \textbf{Category}                   & \textbf{Method}                 & \textbf{Severity} & \textbf{Compose Strategy}            \\ 
        \Xhline{1.5pt}
        \multirow{2}{*}{\bf Basic}     & Lowercase              & 1        & \multirow{2}{*}{Sequential} \\
                                   & Constraction/Expansion & 1        &                             \\
        \hline
        \multirow{5}{*}{\bf Sentence}  & Passive      & 1        & \multirow{5}{*}{Random}     \\
                                   & Active      & 1        &                             \\
                                   & Casual       & 1        &                             \\
                                   & Formal       & 1        &                             \\
                                   & Back Translation              & 1        &                             \\
        \hline
        \multirow{3}{*}{\bf Word}      & Swap Synonym                & 5        & \multirow{3}{*}{Random}     \\
                                   & Insert Adv.              & 1        &                             \\
                                   & Add Irrelevant              & 1        &                             \\
        \hline
        \multirow{4}{*}{\bf Character} & Ocr                    & 5        & \multirow{4}{*}{Random}     \\
                                   & Typos                  & 5        &                             \\
                                   & Spelling Error         & 5        &                             \\
                                   & Keyboard               & 5        &                             \\
        \hline
        \multirow{2}{*}{\bf Choice}    & Circular Options       & 1        & \multirow{2}{*}{Random}     \\
                                   & Reverse Options        & 1        &                             \\ 
        \Xhline{1.5pt} 
    \end{tabular}
    \label{tab:rob_tc}
\end{table}

Robustness aims at evaluating the capability of MLLMs to maintain accurate performance and meaningful outputs in the face of diverse challenges and variations in input data. This includes addressing data corruption and perturbations, which ensures the model's reliability in real-world applications. To evaluate the robustness of our model, we carefully select mild image and text corruptions, drawing inspiration from recent work~\citep{liang2022helm,qiu2022multimodal,chen2023benchmarking,schiappa2023robustness}.

For image corruptions, we incorporate five corruption categories: \textit{noise, blur, weather, digital} (sourced from ImageNet-C~\citep{DBLP:journals/corr/abs-1903-12261}), and \textit{others} (fundamental data augmentation techniques). For text corruption, we introduce five categories like ~\citep{chen2023benchmarking}: \textit{basic, sentence, word, character} (sourced from ~\citep{gui2021textflint}) and \textit{choice}. The \textit{choice} category specifically represents additional corruption introduced for multi-choice question-answering \textit{Scenarios}. All the corruption methods are shown in Table~\ref{tab:rob_ic} and Table~\ref{tab:rob_tc}. These corruption methods we employ do not distort the core information of the images and text. For instance, the Center Crop for images retains at least 90\% of the image content. Text perturbations solely target the questions, and in the options section, only Circular Option and Reverse Option (circular shifting and reverse order on options respectively) are applied, ensuring that the original meaning of the questions and correct answers remain unchanged.

To simulate real-world complexity, we construct composite corruption sequences with random severity levels for both image and text within each sample. Specifically, corruption methods from various categories are composited in a specific order. For each category, the corruption method to apply is selected based on a composite strategy. We employ two strategies: \textit{Random}, where one corruption method from the category is chosen randomly, and \textit{Sequential}, where all methods from the category are applied sequentially. This approach enables us to assess the model's robustness in a scalable manner, rather than evaluating the model for each instance of every separate corruption. By applying image corruption and text corruption at the same time, we can evaluate the model's performance in handling joint corruption across visual and textual domains.

To assess the model's robustness more accurately, we introduce the Relative Robustness for Multi-choice (RRM). Similar to the RIAM described in Section~\ref{sec:icl}, we eliminate the bias introduced by random choice. The RRM primarily calculates the relative accuracy change of the model beyond random guessing accuracy before and after corruptions.


\subsection{Hallucination}
Hallucination refers to the generated content that is nonsensical or unfaithful to the provided source content~\citep{Ji_2023}. Similar to LLMs, MLLMs also encounter the challenge of hallucination. Since objects are the core elements that contribute to the visual semantics of an image, we study the object hallucination problem, which refers to the generated descriptions containing objects that are inconsistent with the given image~\citep{9706727}. As a result, we utilize the Polling-based Object Probing Evaluation (POPE) pipeline~\citep{POPE} on MSCOCO~\citep{mscoco}. The fundamental concept behind this approach is to transform the evaluation of hallucination into a series of binary classification tasks. This is achieved by presenting MLLMs with straightforward Yes-or-No questions regarding the presence of specific objects within the images (e.g., ``Is there a car in the image?"). Each image is prompted with six such Yes-or-No questions. 
To generate the probing objects, POPE considers three polling strategies by sampling the objects randomly, from popular objects, and among those frequently co-occurring objects, respectively. Additionally, we employ \texttt{PPL} to enhance the reliability of our evaluation. Similar to POPE, we also adopt \textit{Metrics} including accuracy, precision, recall, F1-Score, and the ratio of ``Yes" responses.

\section{Experiments: Details of Evaluation Setup}
\subsection{Details of the Evaluated Models}
\begin{table}[ht]
    \begin{center}
    \small
    \caption{\textbf{Details of the evaluated MLLMs.} mPLUG stands for mPLUG-Owl and LAv2 stands for LLaMA-Adapter-v2. }
    \begin{tabular}{l|c|c|c}
        \Xhline{1.5pt}
        \bf MLLM            &\bf Visual Model  &\bf Language Model   &\bf Overall Parameter  \\
        \Xhline{1.5pt}
        \bf LLaVA           & CLIP ViT-L/14    & MPT 7B              & 7B                    \\
        \bf LAMM            & CLIP ViT-L/14    & Vicuna 13B          & 13B                   \\
        \bf MiniGPT-4       & EVA-G            & Vicuna 7B           & 8B                    \\
        \bf mPLUG           & CLIP ViT-L/14    & LLaMA 7B            & 7B                    \\
        \bf Otter           & CLIP ViT-L/14    & LLaMA 7B            & 9B                    \\
        \bf LAv2            & CLIP ViT-L/14    & LLaMA 7B            & 7B                    \\
        \bf InstructBLIP    & EVA-G            & Vicuna 7B           & 8B                    \\
        \bf Shikra          & CLIP ViT-L/14    & LLaMA 7B            & 7B                    \\
        \bf Kosmos-2        & CLIP ViT-L/14    & Decoder 1.3B        & 1.6B                  \\  
        \Xhline{1.5pt}
    \end{tabular}
    \label{tab:mllm_details}
    \end{center}
\end{table}

\begin{table}[ht]
    \centering
    \small
    \caption{\textbf{Success rate in choice extraction on MMBench.} The results represent the success rate in choice extraction of Step-1, which is defined in MMBench. MMBench released the evaluation code for three models. The results in ChEF are aligned with those in MMBench.}
    \begin{tabular}{l|c|c} 
        \Xhline{1.5pt}
         &  MMBench&  ChEF\\ 
         \Xhline{1.5pt}
         LLaVA&  14.85&  14.78\\ 
         MiniGPT-4&  55.58&  52.52\\ 
         InstructBLIP&  91.2&  91.52\\ 
         \Xhline{1.5pt}
    \end{tabular}
    \label{tab:align_mmbench}
\end{table}
In Table~\ref{tab:mllm_details}, we show the details of all the evaluated MLLMs in ChEF. In order to ensure that the evaluated MLLMs are relatively up-to-date, we attempt to align the results of the choice extraction success rate in Step-1 with MMBench~\citep{liu2023mmbench}, which is a recently proposed multimudal benchmark. We align the results with all the open-sourced evaluated MLLMs in MMBench, as shown in Table~\ref{tab:align_mmbench}. Due to differences in evaluation settings, such as input queries, inference strategies, and metrics, the evaluated results on MMBench in ChEF may differ slightly from those in MMBench.

\subsection{Default Recipes for Scenarios}
\begin{table}[t]
    \begin{center}
    \small
    \caption{\textbf{Details of default \textit{Recipes}.} Acc. is accuracy. $\texttt{CoT}$ $\rightarrow$ $\texttt{PPL}$  means \texttt{Multi-Turn} with \texttt{CoT} in the first turn and \texttt{PPL} in the second. }
    \begin{tabular}{l|c|c|c}
        \Xhline{1.5pt}
        \bf Scenario            &\bf Instruction         &\bf Inferencer                     & Metric          \\
        \Xhline{1.5pt}
        \bf CIFAR10             & Standard Query    & PPL                               & Acc.             \\
        \bf Omnibenchmark                & Standard Query    & Multi-Turn PPL                & WeightedACC     \\
        \bf Flickr30k           & Standard Query    & PPL                               & Acc.             \\
        \bf VOC2012             & Standard Query    & Multi-Turn PPL                    & Acc.             \\
        \bf FSC147              & Standard Query    & PPL                               & Acc.             \\
        \bf ScienceQA           & Standard Query    & CoT $\rightarrow$ PPL                         & Acc.             \\
        \bf MMBench             & Standard Query    & CoT $\rightarrow$ PPL                         & Acc.              \\
        \bf MME                 & Standard Query    & PPL                               & Acc.             \\
        \bf SEEDBench           & Standard Query    & PPL                               & Acc.             \\
        \Xhline{1.5pt}
    \end{tabular}
    \label{tab:setting_details}
    \end{center}
\end{table}
\begin{table}[t]
\centering
    \small
    \caption{\textbf{Results of VanillaEval and CircularEval on MMBench.} The results reveal a substantial decrease in accuracy when switching from VanillaEval to CircularEval.}
    \begin{tabular}{l|cc}
    \Xhline{1.5pt}
                             & \bf VanillaEval & \bf CircularEval \\ \Xhline{1.5pt}
    \bf LLaVA                    & 43.13   & 10.24    \\
    \bf LAMM                     & 44.47   & 14.21    \\
    \bf MiniGPT-4                & 54.34   & 26.46    \\
    \bf mPLUG                    & 49.57   & 12.24    \\
    \bf Otter                    & 53.91   & 26.27    \\
    \bf LAv2                     & 57.06   & 24.01    \\
    \bf InstructBLIP             & 65.73   & 46.8     \\
    \bf Shikra                   & 63.26   & 43.08    \\
    \bf Kosmos-2                 & 32.82   & 1.2 
    \\ \Xhline{1.5pt}
    \end{tabular}
    \label{tab:mmbench_cir}
\end{table}
In ChEF, we provide default \textit{Recipes} for each \textit{Scenario}. 
In Table~\ref{tab:setting_details}, we show the details of the default \textit{Recipes} for each \textit{Scenario}. Among the \textit{Scenarios}, the Omnibenchmark is meticulously labeled using a hierarchical chain of categories, facilitated by the Bamboo tree methodology~\citep{zhang2022bamboo}. For \textit{Instruction}, we employ standard queries as nearly all MLLMs lack the ability for in-context learning.

For \textit{Inferencer}, we adopt \texttt{PPL} for most \textit{Scenarios}. For ScienceQA and MMBench, we employ \texttt{Multi-Turn}, with the first turn using the \texttt{CoT}, followed by the \texttt{PPL} in the second turn. For fine-grained classification tasks, we utilize the \texttt{Multi-Turn}, where each turn is a \texttt{PPL}, to hierarchically inquire about categories. For detection tasks, the first turn employs \texttt{PPL} to inquire about categories, while the second turn utilizes \texttt{PPL} to inquire about bounding boxes. The answer pool for CIFAR-10 encompasses the ten predefined classes, while for FSC147, it involves the ground truth values with an additional range of ±2. The answer pool for Omnibenchmark is randomly retrieved from the category tree in Bamboo~\citep{zhang2022bamboo}. In the case of Flickr30k, the answer pool is determined by retrieving the top-$k$ negative candidates from the test data based on BERT similarity~\citep{reimers2019sentencebert}. The answer pool for VOC2012 is randomly generated by scaling and translating the ground-truth bounding boxes. The answer pool for multimodal question-answering tasks is the options \{A, B, C, D\}.  

In the \textit{Metric}, a single accuracy measure is utilized to assess all \textit{Scenarios} uniformly. For certain specialized \textit{Scenarios}, we adopt specific approaches to calculate accuracy. For Omnibenchmark, weighted accuracy is employed, which entails a weighted accuracy calculation based on the granularity of the predicted classification.
MMBench provides two evaluation settings (\emph{i.e.}, VanillaEval and CircularEval), where the CircularEval is used to assess the MLLMs' consistency in responses for the same question when the order of options is changed. 
We conduct evaluations in both settings, as shown in Table~\ref{tab:mmbench_cir}. Across all MLLMs, a significant decline is observed, indicating MLLMs' poor performance in consistency. The utilization of CircularEval assesses a composite capability with both visual performance and consistency. To disentangle these two dimensions of capability, we employ the VanillaEval for the default \textit{Recipe} and incorporate hallucination and robustness within the desiderata to evaluate the dimensions associated with consistency.

\subsection{Recipes for Desiderata}

\begin{table}[t]
    \begin{center}
    \small
    \caption{\textbf{Details of Recipes for six dimensions of desiderata.} ICL is in-context learning. Ins. Follow. is instruction following and Lang. Perf. is language performance.}
    \begin{tabular}{l|c|c|c|c}
        \Xhline{1.5pt}
        \bf Desiderata                  &\bf Scenario            &\bf Instruction         &\bf Inferencer                     & Metric          \\
        \Xhline{1.5pt}
        \bf Calibration                 &MMBench + ScienceQA     & Standard Query        & CoT $\rightarrow$ PPL                         & ECE \\
        \bf ICL                         &MMBench + ScienceQA     & Random ICE            & CoT $\rightarrow$ PPL                         & RIAM         \\
        \bf Ins. Follow.                &MMBench + ScienceQA     & Standard Query        & CoT $\rightarrow$ PPL                         & MR\\
        \bf Lang. Perf.                 &ScienceQA               & Standard Query        & CoT $\rightarrow$ PPL                             & GPT-based Metric  \\
        \bf Robustness                  &MMBench + ScienceQA     & Standard Query        & CoT $\rightarrow$ PPL                             & MRR  \\
        \bf Hallucination               &MSCOCO                  & Standard Query        & PPL                                   & Acc \\
        \Xhline{1.5pt}
    \end{tabular}
    \label{tab:desiderata_setting_details}
    \end{center}
\end{table}

We employ specialized \textit{Recipes} to assess the six dimensions of desiderata, as shown in Table~\ref{tab:desiderata_setting_details}.
All the six dimensions of desiderata except language performance and hallucination are evaluated on MMBench and ScienceQA. Language performance is evaluated on 250 samples random retrieved from ScienceQA and MMBench. Following POPE~\citep{POPE}, hallucination is specifically assessed on the MSCOCO dataset~\citep{mscoco}.

In terms of the \textit{Instruction}, Random \texttt{ICE} is employed as the \textit{Instruction} for ICL evaluation, while standard queries are utilized for the other dimensions. For most MLLMs that lack support for multi-image input, the Random \texttt{ICE} consists solely of text, while for MLLMs that do support multi-image input, such as Otter~\citep{li2023otter}, the Random \texttt{ICE} is adapted to incorporate images. For instruction following evaluation, we concatenate instructions from different groups of verbalizer manipulation at the end of the standard query.

For the \textit{Inferencer}, we employ \texttt{Multi-Turn} with the first turn using the \texttt{CoT}, followed by \texttt{PPL}. The \textit{Metric} we use for each dimension is discussed in Section~\ref{sec:desiderata_sup}.

\section{Empirical Experiments on Desiderata}

\subsection{Calibration}
\begin{table}[bt]
    \begin{center}
    \small
    \caption{\textbf{Results of calibration.} Acc. stands for accuracy and ECE is the Expected Calibration Error. The overall score is calculated through 1 - weighted average ECE, representing the reliability of the model's prediction probability. The entries that are both bold and underlined indicate the best performance.}
    \begin{tabular}{l|cc|cc|c}
        \Xhline{1.5pt}
        \multirow{2}{*}{\diagbox{\bf MLLM}{\bf Scenario}}& \multicolumn{2}{c|}{\bf ScienceQA} & \multicolumn{2}{c|}{\bf MMBench} & \multirow{2}{*}{\bf Overall} \\
                                                         & Acc. $\uparrow$ & ECE\% $\downarrow$   & Acc. $\uparrow$  & ECE\% $\downarrow$      &         \\ 
        \Xhline{1.5pt}
        \bf LLaVA                                        & 46.55          & \ul{7.26}     & 44.13         & 14.66        & 90.01   \\
        \bf LAMM                                         & 52.75          & 20.79         & 44.47         & 28.52        & 76.36   \\
        \bf MiniGPT-4                                    & 47.00          & 15.28         & 54.34         & 15.24        & 84.73   \\
        \bf mPLUG                                        & 48.44          & 15.72         & 49.57         & 15.47        & 84.15   \\
        \bf Otter                                        & 50.22          & 21.10         & 53.91         & 10.52        & 82.80   \\
        \bf LAv2                                         & 54.34          & 8.17          & 57.06         & 14.19        & 89.61   \\
        \bf InstructBLIP                                 & \ul{55.18}     & 10.57         & \ul{65.73}    & \ul{6.25}    & \ul{91.25}\\
        \bf Shikra                                       & 45.21          & 14.57         & 63.26         & 6.65         & 88.35   \\
        \bf Kosmos-2                                     & 34.60          & 10.63         & 32.82         & 11.13        & 89.19   \\
        \Xhline{1.5pt}    
    
    \end{tabular}
    \label{tab:calibration}
    \end{center}
    \vspace{-0.5cm}
\end{table}
\begin{figure}[htb]
  \centering
  \begin{subfigure}[b]{0.5\textwidth}
    \centering
    \includegraphics[width=\textwidth]{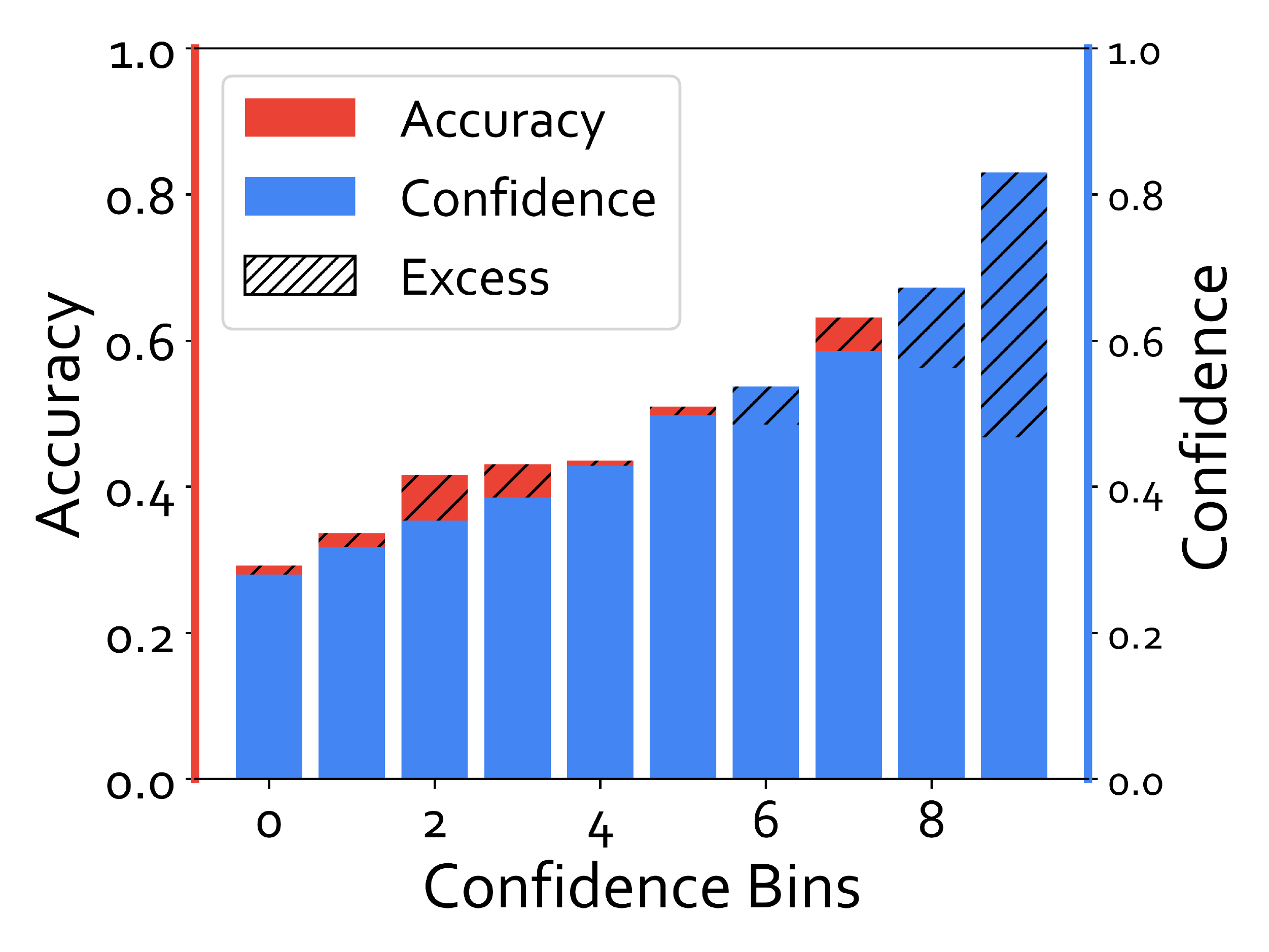}
    \caption{LLaVA}
    \label{fig:calibsub1}
  \end{subfigure}%
  \hfill
  \begin{subfigure}[b]{0.5\textwidth}
    \centering
    \includegraphics[width=\textwidth]{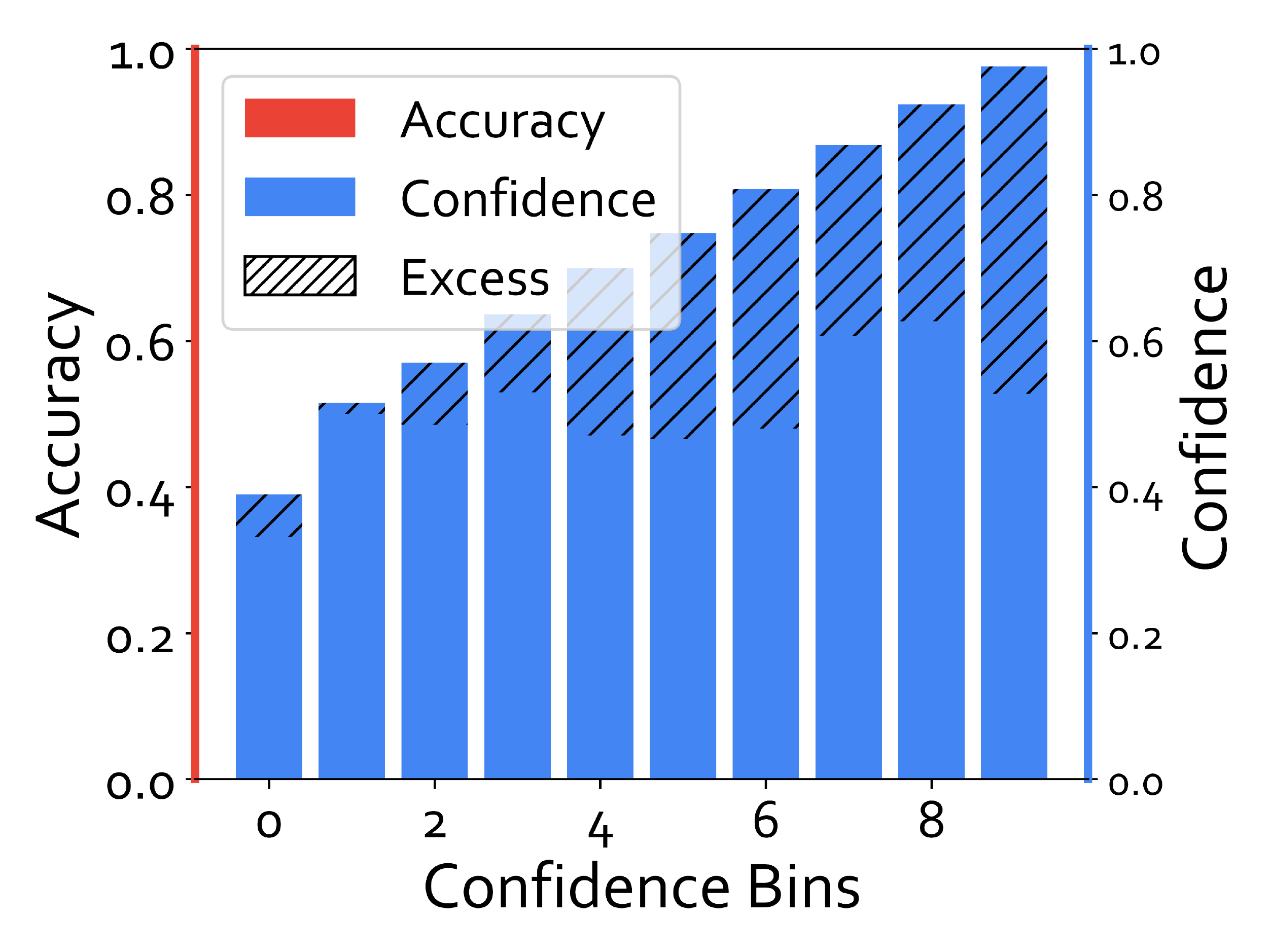}
    \caption{Otter}
    \label{fig:calibsub2}
  \end{subfigure}
  \caption{\textbf{Reliability diagrams for LLaVA and Otter on ScienceQA.} The red excess parts represent the degree of insufficient confidence of the model, and the blue excess parts represent the degree of overconfidence of the model.} 
  \label{fig:calibration}
\end{figure}

The calibration results are presented in Table~\ref{tab:calibration}. To illustrate the differences in calibration performance, we also provide reliability diagrams for LLaVA and Otter on ScienceQA in Figure~\ref{fig:calibration}. In reliability diagrams, predictions are sorted based on the MLLMs' confidence scores, and an equal number of predictions are grouped into 10 bins. By calculating the average confidence and accuracy within each bin, we can compare and evaluate the gap between confidence and accuracy intuitively. The observations are as follows:

\textbf{(1)} Higher accuracy does not imply better calibration.
In ScienceQA, LLaVA demonstrates an average accuracy with the lowest ECE, showing a relatively better calibration. In contrast, Otter achieves higher accuracy with the highest ECE, showing a relatively worse calibration. Reliability diagrams provide a more intuitive and detailed illustration. We can observe that the confidence and actual accuracy in the first 9 bins exhibited a clear correlation, indicating that the predicted confidence of the first 90\% of LLaVA is relatively well calibrated. However, the reliability diagram of Otter shows a larger gap between confidence and accuracy, suggesting that Otter's predicted confidence is relatively poorly calibrated.

\textbf{(2)} Higher confidence does not imply higher accuracy and better calibration.
In the reliability diagrams, both MLLMs have a substantial gap between confidence and accuracy in the last bin, which contains samples with top 10\% confidence. Both MLLMs exhibit overconfidence in these samples, which reminds us to avoid considering higher confidence as evidence for higher accuracy. Additionally, it can be observed that the gap between accuracy and confidence does not decrease with increasing confidence, indicating that confidence cannot effectively represent reliability.

\textbf{(3)} InstructBLIP achieves the highest accuracy in both visual tasks, while simultaneously exhibiting remarkably low ECE, indicating exceptional calibration. Conversely, other models demonstrate a certain trade-off between the two dimensions.
It implies that InstructBLIP can yield superior calibration, so as to provide precise answers to questions while accurately conveying its uncertainty.
%


\subsection{In-context Learning} 
\begin{figure}[htb]
  \centering
\includegraphics[width=\textwidth]{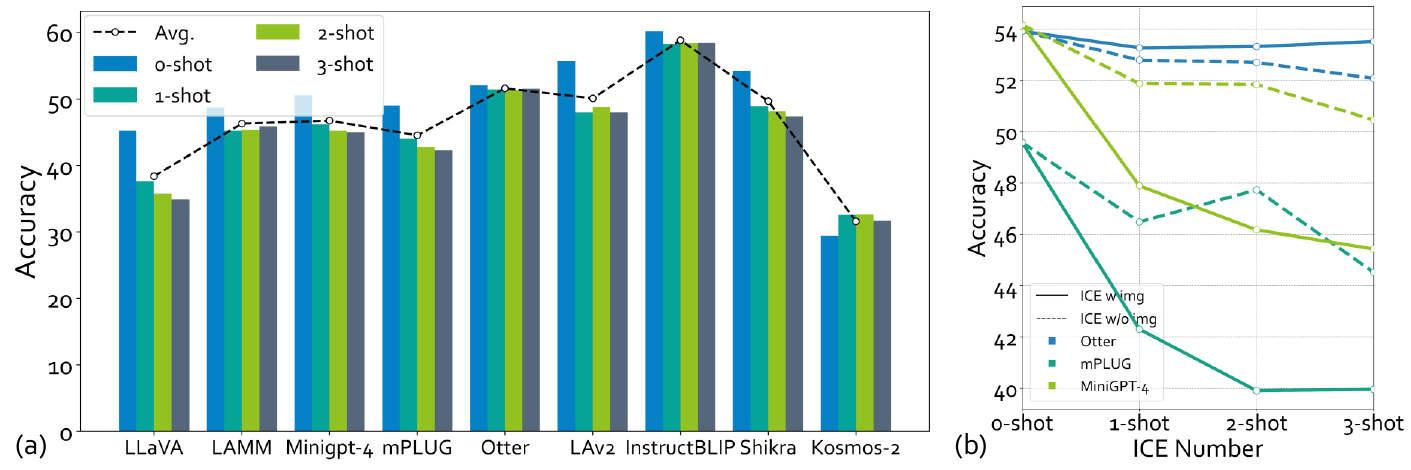}
  \caption{\textbf{Results of in-context learning.} (a) Average results of in-context learning on ScienceQA and MMBench utilizing various \texttt{ICE} numbers. (b) Results of in-context learning on MMBench for Otter, mPUG-Owl, and MiniGPT-4, utilizing various \texttt{ICE} numbers with and without images respectively.}
  \label{fig:icl_default}
\end{figure}

\begin{figure}[htb]
    \centering
    \includegraphics[width=\textwidth]{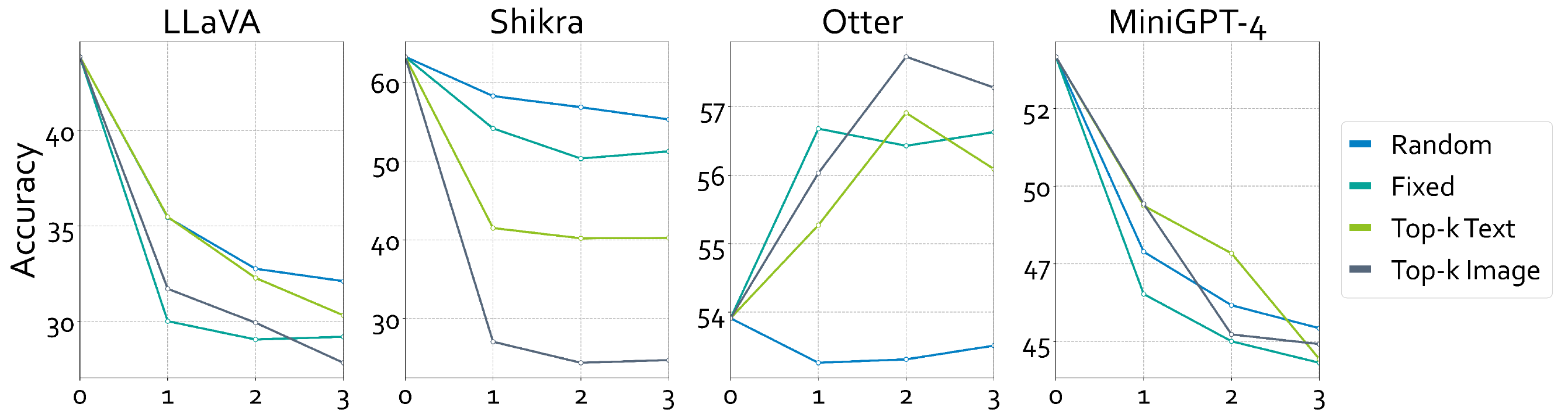}
    \caption{\textbf{Experimental results of evaluation with ICE as \textit{Instruction} under different retriever settings.} The retriever methodologies employed encompass Random, Fixed, Top-$k$ Text, and Top-$k$ Image.}
  \label{fig:icl_retriever}
\end{figure}

The evaluations of in-context learning (ICL) are conducted on ScienceQA and MMBench, with \texttt{ICE} numbers set at 0, 1, 2, and 3 respectively. The ICL retriever used in the experiments is Random. The experimental results are illustrated in Figure~\ref{fig:icl_default}(a). To evaluate the influence of accompanying images in \texttt{ICE}, we also conduct experiments using Otter, mPLUG-Owl, and MiniGPT-4, as shown in Figure~\ref{fig:icl_default}(b). These models are evaluated on MMBench using random retrieved \texttt{ICE} with and without images respectively. To compare the different performance of MLLMs with retrieved \texttt{ICE} under different settings, we further evaluate MMBench, utilizing LLaVA, Shikra, Otter, and MiniGPT-4, as shown in Figure~\ref{fig:icl_retriever}. The methodologies employed for the \texttt{ICE} retriever include Random, Fixed, Top-$k$ Text, and Top-$k$ Image. The observations are as follows:

\textbf{(1)} It can be observed from Figure~\ref{fig:icl_default}(a) that most of the MLLMs exhibited a decline in performance compared to the zero-shot setting, except for Otter and Kosmos-2. This can be attributed to Otter's training on in-context instruction tuning data, thus enhancing its ICL capabilities. 
In contrast, the observed improvement in Kosmos-2's performance is due to its struggles to comprehend the meaning of options \{A, B, C, D\} provided in the question, resulting in difficulty in aligning the answers to options. The number of \texttt{ICE} does not present a significant impact on the results. From an overall perspective, the majority of MLLMs do not demonstrate capabilities in ICL.

\textbf{(2)} Otter demonstrates a slight enhancement when deploying \texttt{ICE} with images compared to the \texttt{ICE} without image, as shown in Figure~\ref{fig:icl_default}(b). However, its performance attenuates in the absence of images, failing to manifest its ICL capabilities. This suggests that integrating \texttt{ICE} with an image is a judicious design choice within MLLMs. Contrarily, neither mPLUG-Owl nor MiniGPT-4 shows improvement in their capabilities regardless of the presence or absence of images in the \texttt{ICE}.

\textbf{(3)} It can be observed from Figure~\ref{fig:icl_retriever} that different retrievers have different results, and the Top-$k$ method exhibits slightly inferior performance compared to the others. This potential decline in performance might be attributed to the fact that the MLLMs might regard the given answer in a similar \texttt{ICE} as the correct answer for the \texttt{Query}, thereby influencing the model's prediction.

\subsection{Instruction Following}
\begin{table}[htbp] 
    \begin{center}
    \small
    \caption{\textbf{Results of instruction following.} The abbreviations we use are: Acc for original accuracy; Acc\textsubscript{vm} for the weighted average accuracy for different instructions of verbalizer manipulation; MR for the weighted average match ratio for different instructions of verbalizer manipulation, as defined in Section~\ref{sec:desiderata_sup}; Avg. for an average of results on ScienceQA and MMBench. The entries that are both bold and underlined indicate the best performance.}
    \begin{adjustbox}{width=\textwidth}
    \begin{tabular}{l|ccc|ccc|ccc}
    \Xhline{1.5pt}
    \multirow{2}{*}{\diagbox{\textbf{MLLM}}{\textbf{Scenario}}} 
    & \multicolumn{3}{c|}{\bf ScienceQA} 
    & \multicolumn{3}{c|}{\bf MMBench} 
    & \multicolumn{3}{c}{\bf Avg.} \\
    & Acc  $\uparrow$   & Acc\textsubscript{vm} $\uparrow$   & MR\% $\uparrow$      & Acc$\uparrow$      & Acc\textsubscript{vm} $\uparrow$  & MR\% $\uparrow$   & Acc $\uparrow$  & Acc\textsubscript{vm} $\uparrow$       & MR\% $\uparrow$          \\     \Xhline{1.5pt}
    \bf LLaVA            & 46.55    & 41.10    & \ul{46.23}   & 44.13   & 35.02   & 39.60     & 45.66 & \ul{38.86}  & 43.79      \\
    \bf LAMM             & 52.75    & 41.41    & 42.41   & 44.47   & 34.11   & 34.72          &49.70 & 38.72       & 39.58      \\
    \bf MiniGPT-4        & 47.00    & 36.95    & 43.01   & 54.34   & 41.81   & \ul{43.78}     &49.70 & 38.74       & 43.29      \\
    \bf mPLUG            & 48.44    & 39.93    & 40.28   & 49.57   & 35.39   & 33.43          &48.86 & 38.25       & 37.76      \\
    \bf Otter            & 50.22    & 38.65    & 38.30   & 53.91   & 33.29   & 36.90          &51.58 & 36.67       & 37.78      \\
    \bf LAv2             & 54.34    & \ul{41.71}    & 44.40   & 57.06   & 27.38   & 28.83     &55.34 & 36.43       & 38.66      \\
    \bf InstructBLIP     & \ul{55.18}  & 38.23 & 45.07   & \ul{65.73}  & \ul{37.59}  & 43.46  &\ul{59.07} & 38.00       & \ul{44.47}      \\
    \bf Shikra           & 45.21    & 35.80    & 37.89   & 63.26   & 31.58   & 32.91          &51.86 & 34.24       & 36.05      \\
    \bf Kosmos-2         & 34.60    & 35.36    & 17.70   & 32.82   & 32.17   & 14.19          &33.94 & 34.18       & 16.41     \\
    \Xhline{1.5pt}
    \end{tabular}
    \end{adjustbox}
    \label{tab:insfollowmain}
    \end{center}
\end{table}

\begin{figure}[htb]
    \centering
    \includegraphics[width=0.8\textwidth]{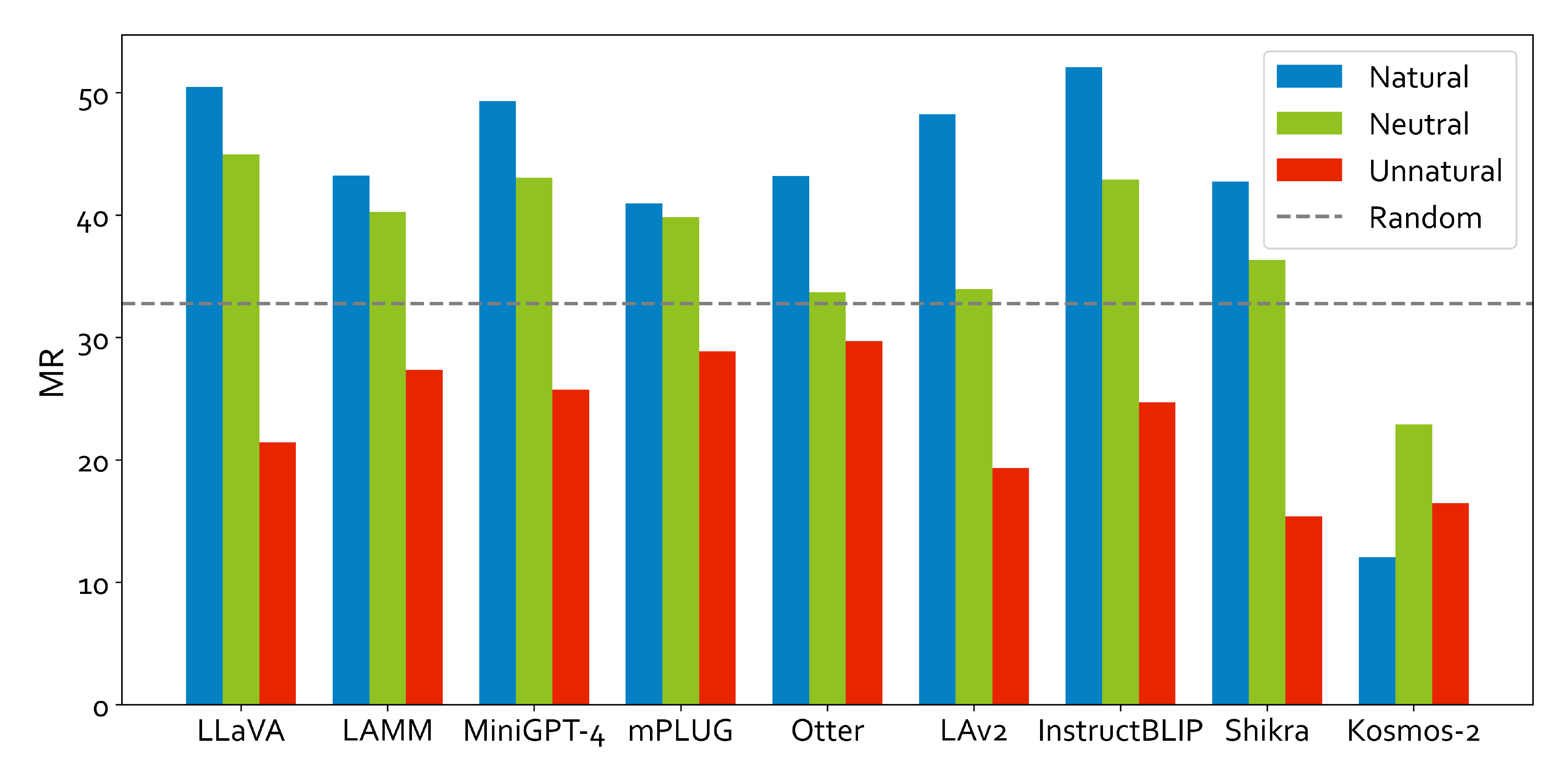}
    \caption{\textbf{Results of instruction following with different verbalizer manipulation}, where Natural represents the accuracy with instructions of natural verbalizer; Neutral represents the accuracy with instructions of neutral verbalizer; Unnatural represents the accuracy with instructions of unnatural verbalizer; the dotted line represents the accuracy of random guessing.}
  \label{fig:inf_ablation}
\end{figure}

Table~\ref{tab:insfollowmain} reports the results of instruction following on ScienceQA and MMBench. We also report the original accuracy Acc and the weighted average accuracy Acc\textsubscript{vm} of different verbalizer manipulation instructions. To further explore the instruction following, we show the results of different verbalizer manipulations in Figure~\ref{fig:inf_ablation}. We also provide the results in Figure~\ref{fig:inf_ablation}, that follow the ranking of different groups of instructions in alignment with prior knowledge (\textit{natural} $>$ \textit{neutral} $>$ \textit{unnatural}), where MR also decreases sequentially. The observations are as follows:

\textbf{(1)} We observed that some MLLMs do not experience a significant decrease in Acc\textsubscript{vm} compared to Acc when the MR is low. This can be attributed to cases where the original response is incorrect but become correct after verbalizer manipulation. On the other hand, questions that are initially answered correctly remain largely consistent between before and after verbalizer manipulation. This suggests that the models exhibit higher instruction following ability on confident questions but are more susceptible to disturbance on uncertain questions. It indicates a correlation between instruction following and confidence in question answering.


\textbf{(2)} The distributions of Acc and MR are entirely different, where the distribution of MR is more discriminative. For Kosmos2, Acc\textsubscript{vm} increases because its original accuracy is lower than that of random guessing (35.80 on ScienceQA, 27.57 on MMBench), and random guessing improves accuracy. Kosmos-2 exhibits a similar Acc\textsubscript{vm} to Shikra but has the lowest MR, further confirming that Kosmos-2 has degenerated into random guessing, leading to its poor performance on instruction following.

\textbf{(3)}  The results of natural and neutral are significantly higher than that of unnatural, suggesting that the model is more likely to follow instructions in the natural and neutral categories. This is further supported by their probabilities being higher than random guessing, indicating that the model indeed possesses a certain level of understanding of these two sets of instructions rather than making random guesses. On the other hand, most of the unnatural instructions perform well below the level of random guessing, demonstrating that following unnatural instructions is highly challenging for current MLLMs.

\textbf{(4)} InstructBLIP performs best in the natural category but exhibits a noticeable performance drop in the neutral category, suggesting that InstructBLIP relies more on prior knowledge for instruction understanding rather than comprehending new instruction content.


\subsection{Language Performance} 
\label{sec:language_performance_exp}

\begin{figure}[htbp]
  \centering
  \begin{subfigure}[]{0.4\textwidth}
    \begin{center}
    \small
        \begin{tabular}{l|c}
        \Xhline{1.5pt}
        \bf MLLM             & \textbf{Lang. Perf.} \\
        \Xhline{1.5pt}
        \bf LLaVA            & 84.82                \\
        \bf LAMM             & 79.08                \\
        \bf MiniGPT-4        & 70.66                \\
        \bf mPLUG            & 88.44                \\
        \bf Otter            & 74.05                \\
        \bf LAv2             & \ul{90.85}           \\
        \bf InstructBLIP     & 80.01                \\
        \bf Shikra           & 66.67                \\
        \bf Kosmos-2         & 45.86                \\
        \Xhline{1.5pt}
        \end{tabular}
        \caption{}
        \label{tab:langperf}
    \end{center}
  \end{subfigure}
  \begin{subfigure}[]{0.55\textwidth}
        \centering
        \includegraphics[height=3.7cm]{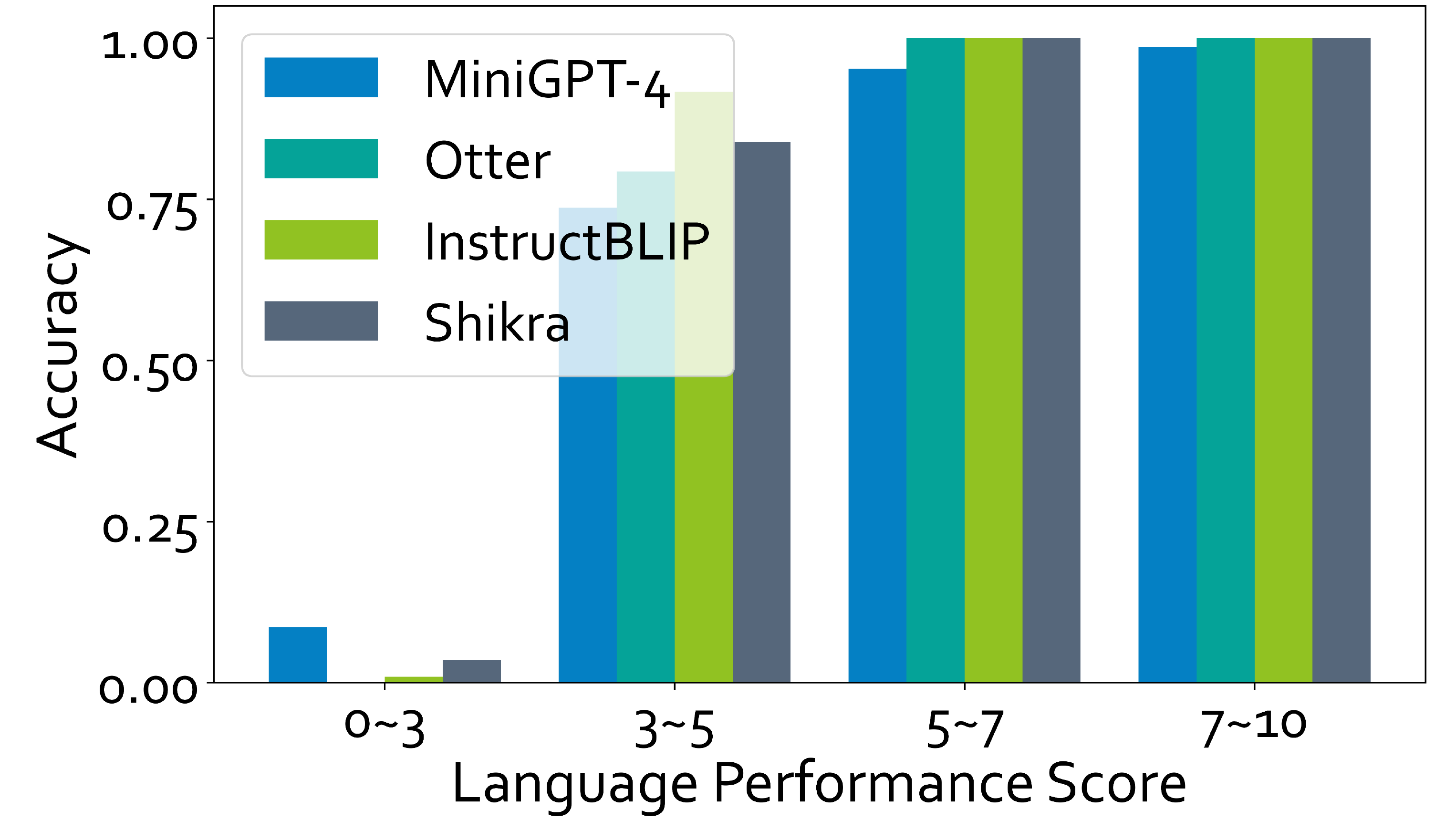}
        \caption{}
      \label{fig:gpt_distribution}
  \end{subfigure}%
  \caption{\textbf{Analysis of language performance.} (a) Complete results of language performance. This desiderata only evaluates the natural language generation quality of thought chains in which the model provides correct conclusions to prevent conclusion accuracy from dominating the language performance score. (b) Accuracy distribution in language performance. We divide the evaluation samples into four intervals based on GPT scores and calculate the conclusion accuracy within each interval. }
  \label{fig:language_perf_analysis}
\end{figure}
\begin{figure}[htb]
  \centering
    \includegraphics[width=\textwidth]{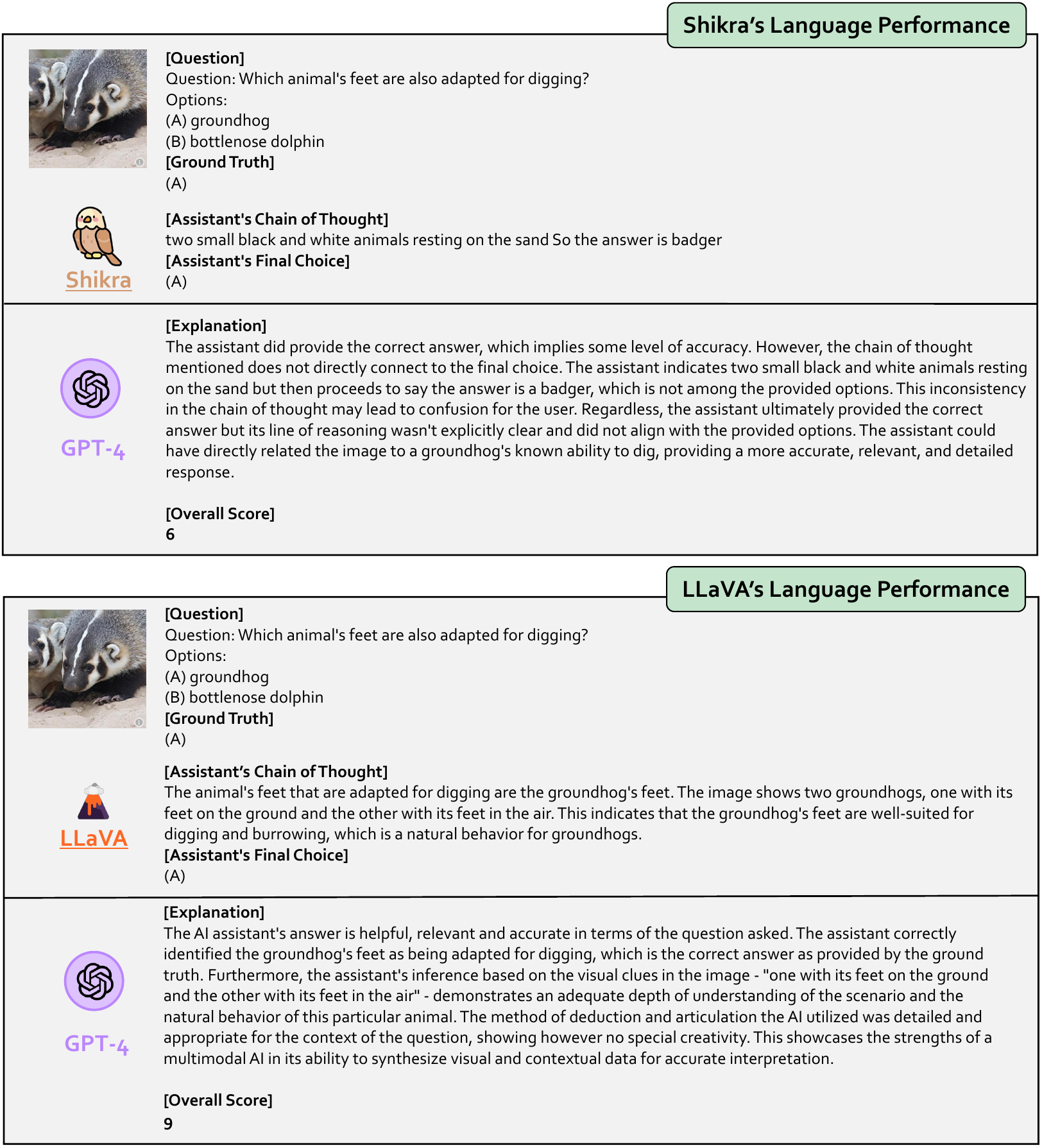}
    
  \caption{\textbf{Examples of language performance evaluation on Shikra and LLaVA}, where two models exhibit varying levels of natural language generation quality. GPT-4 generates an evaluation explanation as evidence and then generates an overall score based on the Chain-of-Thought and the final choice generated by the Assistant. Here, we present only one explanation and the overall score generated by GPT-4. Note that in practice, for each sample, GPT-4 generates five explanations and their corresponding overall scores through sampling. The final score for the sample is obtained by averaging these five overall scores.}
    \label{fig:gpt_eval_example}
\end{figure}

Table~\ref{fig:language_perf_analysis}(a) provides the results of language performance. To illustrate the role of selective sampling of correct, Figure~\ref{fig:language_perf_analysis}(b) displays the distribution of accuracy across different score ranges of language performance scores. We also provide a typical example in Figure~\ref{fig:gpt_eval_example}, comparing the disparity in language performance between LLaVA and Shikra when both provide correct answers. The following presents our key observations:

\textbf{(1)} Kosmos-2 exhibits poor performance due to its inability to provide reasoning processes in practical multi-choice question-answering \textit{Scenarios}. Conversely, Shikra demonstrates relatively weak performance attributed to its incapacity to deliver reasoning analysis. Despite prompts intended to elicit the reasoning process, Shikra tends to provide direct answers, leading to lower scores. In real interactive scenarios, MLLMs should offer some form of reasoning alongside their answers. Merely achieving higher accuracy does not necessarily guarantee enhanced interactivity. Therefore, the significance of language performance is highlighted. This finding further emphasizes the imperative need to evaluate language performance in MLLMs.

\textbf{(2)} As shown in Figure~\ref{fig:language_perf_analysis}(b), it can be observed that samples with lower language performance scores (0-3) provided by GPT are predominantly incorrect answers by the MLLMs, indicating that the low scores are largely influenced by the accuracy of the answers rather than language performance. Conversely, in the three bins with scores $>$ 3, the accuracy significantly improved. It is worth noting that despite most models providing correct answers, there are substantial differences in language performance. Therefore, it is appropriate to evaluate the language performance of only the correct samples to mitigate the impact of answer accuracy on the evaluation.

\textbf{(3)} Figure~\ref{fig:gpt_eval_example}, as a typical example, illustrates the difference in language performance between LLaVA and Shikra, when they both provide correct answers. Regarding Shikra, GPT-4 noted that its generated-chain of-thought, while yielding the correct answer, lacks relevance to the given options. This inconsistency could potentially cause confusion, resulting in a lower score of 6. In the case of LLaVA, its generated chain-of-thought showcases a logical process that adeptly employs visual information for reasoned deductive reasoning. GPT successfully acknowledges the strengths of LLaVA's chain-of-thought, providing a comprehensive explanation for its impressive score of 9. The deduction of one point may be attributed to a limited presence of divergent associations and generalizations.


\subsection{Robustness}

\begin{table}[htbp]
\begin{center}
\small
\caption{\textbf{Results of robustness.} Acc represents the original accuracy without corruptions; Acc\textsubscript{crp} represents the accuracy after image and text corruptions; RRM\% is Relative Robustness for multi-choice; Avg. is the weighted average results on ScienceQA and MMBench; As Kosmos-2$\dagger$ degenerates into random guessing, the results are meaningless. The entries that are both bold and underlined indicate the best performance. }
\begin{adjustbox}{width=\textwidth}
\begin{tabular}{l|ccc|ccc|ccc}
\Xhline{1.5pt}
\multirow{2}{*}{\diagbox{\textbf{MLLM}}{\textbf{Scenario}}} & \multicolumn{3}{c|}{\bf ScienceQA} & \multicolumn{3}{c|}{\bf MMBench} & \multicolumn{3}{c}{Avg.} \\
                  & Acc $\uparrow$    & Acc\textsubscript{crp} $\uparrow$  & RRM\%   $\uparrow$   & Acc  $\uparrow$   & Acc\textsubscript{crp} $\uparrow$ & RRM\%  $\uparrow$   & Acc & Acc\textsubscript{crp}  $\uparrow$     & RRM\%  $\uparrow$        \\     \Xhline{1.5pt}
\bf LLaVA             & 46.55   & 43.45      & 71.13   & 44.13   & 35.85    & 50.03 & 45.66 & 40.65      & 63.36       \\
\bf LAMM              & 52.75   & 43.89      & 47.75   & 44.47   & 40.33     & \ul{75.52} &49.70 & 42.58  & 57.98       \\
\bf MiniGPT-4         & 47.00   & 42.29      & 57.95   & 54.34   & 44.87     & 64.61  &49.70 & 43.24      & 60.40       \\
\bf mPLUG             & 48.44   & 42.87      & 55.93   & 49.57   & 36.96     & 42.68  &48.86 & 40.69      & 51.05       \\
\bf Otter             & 50.22   & 44.19      & 58.16   & 53.91   & 42.18     & 55.47  &51.58 & 43.45      & 57.17       \\
\bf LAv2              & 54.34  & 49.31  & \ul{72.89}   & 57.06   & 43.05     & 52.50  & 55.34 & 47.01      & 65.38       \\
\bf InstructBLIP      & \ul{55.18}   & \ul{49.72}      & 71.83   & \ul{65.73}   & \ul{56.04}     & 52.50  & \ul{59.07} & \ul{52.05}  & \ul{72.85}       \\
\bf Shikra            & 45.21   & 39.17      & 35.81   & 63.26   & 52.07     & 68.65  & 51.86 & 43.92      & 47.01       \\

\bf Kosmos-2$\dagger$          & 34.60   & 35.48      & 26.67  & 32.82   & 28.40     & 15.87  & 33.94 & 32.87      & 22.69
\\     \Xhline{1.5pt}
\end{tabular}
\end{adjustbox}
\label{tab:robustness}
\end{center}
\end{table}

\begin{figure}[htb]
    \centering
    \includegraphics[width=0.8\textwidth]{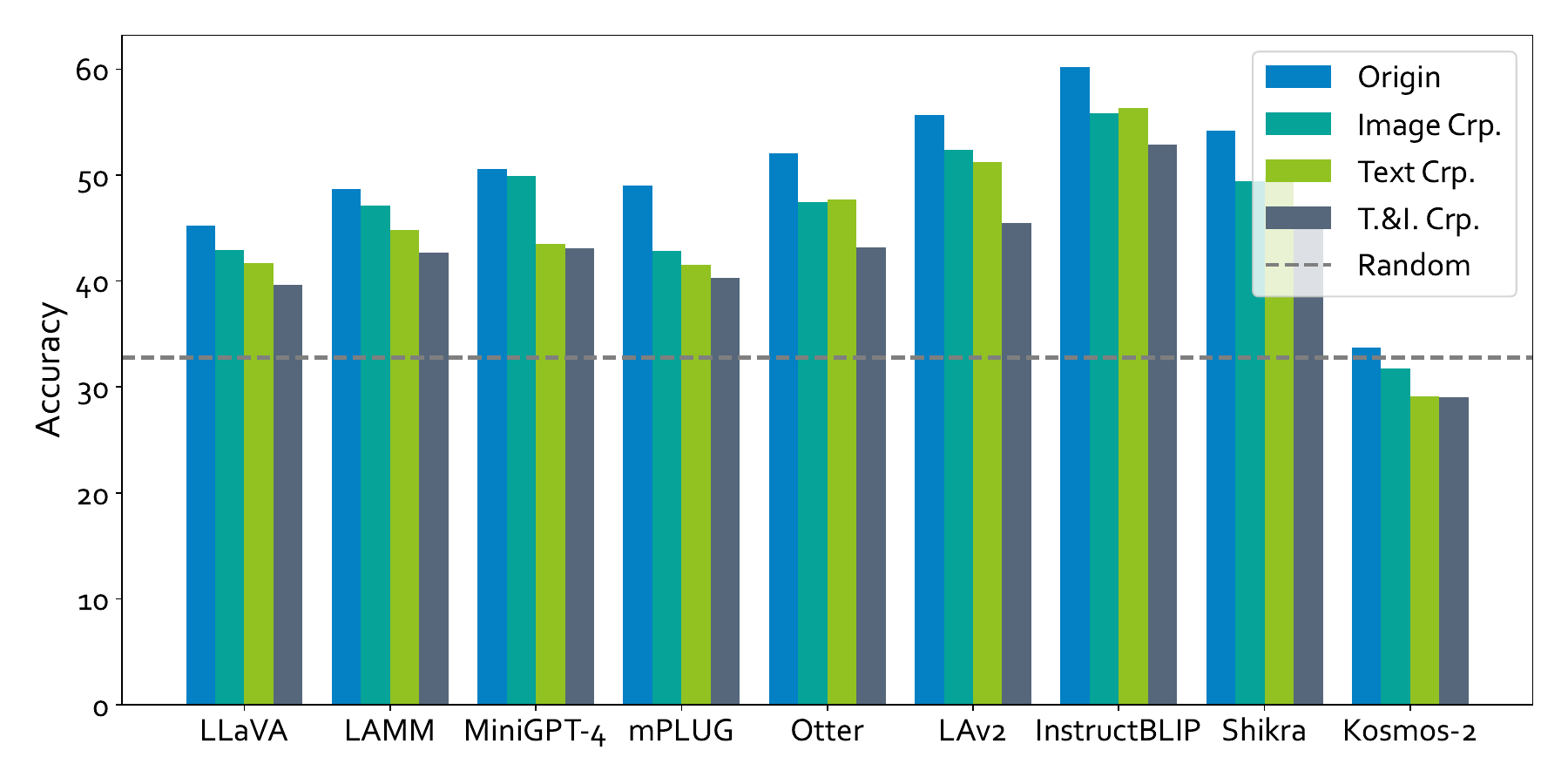}
    \caption{\textbf{Results of robustness under different settings.} The accuracy in this figure represents the weighted average results on ScienceQA and MMBench. The origin represents the original accuracy; Image Crp. represents the accuracy after image corruption; Text Crp. represents the accuracy after text corruption; I.\&T. Crp. represents the accuracy after both image and text corruption; the dotted line represents the accuracy of random guessing. }
  \label{fig:rob_ablation}
\end{figure}

The robustness experiment is conducted on ScienceQA and MMBench, results are presented in Table~\ref{tab:robustness}. The $\operatorname{acc_{random}}$ on ScienceQA is 35.80. The $\operatorname{acc_{random}}$ on MMBench is 27.57. 
In order to evaluate the influence of different corruptions, we conduct experiments on ScienceQA and MMBench using different corruption types, as shown in Figure~\ref{fig:rob_ablation}. These corruptions encompass both image corruption and text corruption. The observations are as follows: 

\textbf{(1)} The experimental results indicate that current MLLMs, when subjected to image and text corruption, do not exhibit significant decreases in accuracy in absolute terms. However, it is the portion of accuracy beyond random guessing that truly reflects the model's capabilities, and it shows significant drops after perturbation. Given the prevalence of perturbations in daily environments, evaluating a model's robustness becomes pivotal.

\textbf{(2)} Image corruptions have a relatively minor effect on model performance, possibly owing to the robustness of the pre-trained vision encoder. In contrast, text corruptions show a significant impact on performance, potentially due to the heightened sensitivity of the MLLMs' text encoder when incorporating the visual tokens. 

\subsection{Hallucination}
\begin{table}[t]
    \begin{center}
    \small
    \caption{\textbf{Results of Hallucination.} Acc represents the accuracy of prediction; Precision represents how many of the predicted positive samples are true positive samples; Recall represents how many of all true positive samples are correctly identified; and Yes\% represents the probability that the model outputs a yes answer. The entries that are both bold and underlined indicate the best performance. }
    \small
    \begin{tabular}{c|l|ccc|c|c}
        \Xhline{1.5pt}
        \bf Dataset         &\bf MLLM                    & \bf Acc   &\bf Precision &\bf Recall &\bf F1 Score &\bf Yes\%  \\ 
        \Xhline{1.5pt}
        \multirow{9}{*}{\bf MSCOCO-Random}
                        &\bf LLaVA               &51.55 &51.55     &100    &68.03    &100    \\
                        &\bf LAMM                &53.84 &54.12     &52.91  &69.19    &95.53  \\
                        &\bf MiniGPT-4            &80.93 &89.67     &71.20  &79.38    &40.92  \\
                        &\bf mPLUG           &55.81 &53.85     &99.80  &69.95    &95.53  \\
                        &\bf Otter               &82.27 &89.11     &74.73  &81.29    &43.23  \\
                        &\bf LAv2    &75.40 &69.54     &93     &79.58    &68.93  \\
                        &\bf InstructBLIP        &\ul{90.24} &93.55     &87.06  &90.19    &47.97  \\
                        &\bf Shikra              &87.18 &87.00     &88.33  &87.66    &52.33  \\
                        &\bf Kosmos-2            &51.55 &51.55     &100    &68.03    &100    \\
        \hline
        \multirow{9}{*}{\bf MSCOCO-Popular}
                        &\bf LLaVA               &50    &50        &100    &66.67    &100    \\
                        &\bf LAMM                &50    &50        &99.93  &66.65    &99.93  \\
                        &\bf MiniGPT-4            &74.3  &75.4      &72.13  &73.73    &47.83  \\
                        &\bf mPLUG               &49.97 &49.98     &99.8   &66.6     &99.83  \\
                        &\bf Otter               &73.57 &73.03     &74.73  &73.87    &51.17  \\
                        &\bf LAv2                &59.10 &55.42     &93     &69.45    &83.90  \\
                        &\bf InstructBLIP        &\ul{83.37} &81.07     &87.07  &83.96    &53.7   \\
                        &\bf Shikra              &83.3  &80.25     &88.33  &84.10    &55.03  \\
                        &\bf Kosmos-2            &50    &50        &100    &66.67    &100    \\
        \hline
        \multirow{9}{*}{\bf MSCOCO-Adverarial}
                        &\bf LLaVA               &50    &50        &100    &66.67    &100    \\
                        &\bf LAMM                &50.13 &50.06     &99.60  &66.64    &99.47  \\
                        &\bf MiniGPT-4            &72.17 &72.51     &71.40  &71.95    &49.23  \\
                        &\bf mPLUG               &50.06 &50.03     &99.80  &66.65    &99.73  \\
                        &\bf Otter               &70.07 &68.35     &74.73  &71.40    &54.67  \\
                        &\bf LAv2                &56.77 &53.92     &93     &68.27    &86.23  \\
                        &\bf InstructBLIP        &\ul{80.63} &77.14     &87.07  &81.80    &56.43  \\
                        &\bf Shikra              &79.27 &74.78     &88.33  &81       &59.07  \\
                        &\bf Kosmos-2            &50    &50        &100    &66.67    &100    \\
        \Xhline{1.5pt}
    \end{tabular}
    \label{tab:hallucination_results}
    \end{center}
    \vspace{-0.3cm}
\end{table}
Table~\ref{tab:hallucination_results} presents the evaluation results for different difficulty levels of hallucination. Among them, LLaVA, LAMM, mPLUG-Owl, and Kosmos-2 exhibit more severe hallucination issues, as they tend to answer ``Yes” very easily. This leads to nearly 100\% Recall but with Acc and Precision both close to 50\%, akin to random selection. Apart from these four models, the other models achieved relatively meaningful results. Overall, InstructBLIP achieved the best results, while Shikra also performed competitively, with an average accuracy being only 1.72\% lower than InstructBLIP's.

\section{ChEF Provides Reliable Assessments of Desiderata}
\begin{figure}[htb]
    \centering
    \includegraphics[width=\textwidth]{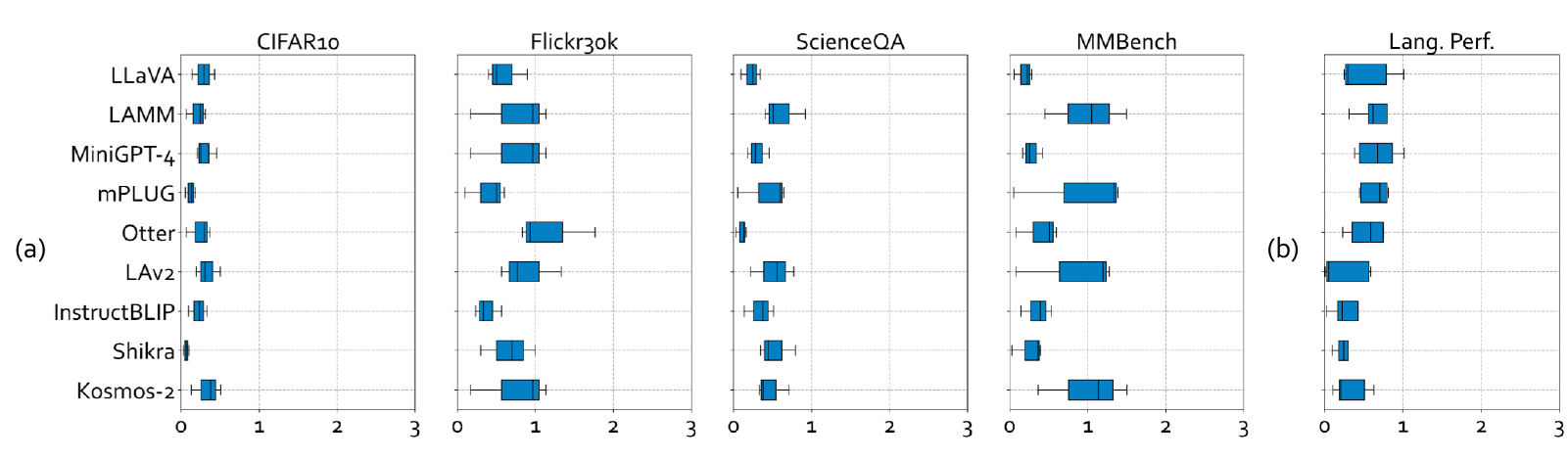}
    \caption{\textbf{Variance across seeds.} (a) Experiments are conducted on CIFAR10, Flickr30k, ScienceQA, and MMBench utilizing various random seeds. (b) Experiments of Language Performance. The results show the deviation from the mean score of 5 sampled evaluation evidence. The black line within each boxplot represents the median.}
  \label{fig:desiderata_reliable}
\end{figure}
Due to the modular design of ChEF, we have the flexibility to employ different \textit{Recipes} for evaluating the same \textit{Scenario} and finally identify the most reasonable \textit{Recipe} that can provide reliable and indicative assessments through experiments. Besides the reliability of evaluating the visual performance, we also try to ensure the stability and reliability of evaluating the desiderata. We conduct experiments to investigate the inherent randomness within them. This entailed scrutinizing the consistency of random factors, such as the utilization of random retrieved \texttt{ICE} for \textit{Instruction} in ICL evaluation. Additionally, the evaluation of language performance, which is based on GPT assessment, inherently incorporates stochastic elements.

To evaluate the stability of random \texttt{ICE} as \textit{Instruction}, we conduct experiments on CIFAR10, Flickr30k, ScienceQA, and MMBench, employing a diverse set of random seeds. To emphasize deviations from the mean value, we first calculate the average of results from the five different seed sets. Then, for each seed, we determine the deviation by subtracting this average and taking the absolute value of the difference. This approach highlights the variation in results for each seed compared to the average. As illustrated in Figure~\ref{fig:desiderata_reliable}(a), the deviation for most model results is at around 1.0, indicating notable stability.

To mitigate systematic errors in GPT evaluation, we employed Multiple Evidence Calibration. In this approach, we prompt GPT-4 to provide evaluation explanations as evidence for deriving the final score, as described in Section~\ref{sec:language_performance}. As illustrated in Figure~\ref{fig:desiderata_reliable}(b), our prompts effectively ensure the stability of GPT's scores across multiple samplings, where the maximum deviation is controlled under 1.0. This implies that GPT can maintain a consistent scale across multiple evaluations. Furthermore, the use of five sampled responses is deemed sufficient for GPT to furnish reliable and meaningful language performance scores.

These results indicate that the \textit{Recipe} we provide for evaluating the desiderata is indicative and reliable.

\section{Evaluation on GPT-4V(ision) and Bard}

\subsection{Evaluation Setup}
We evaluate GPT-4V(ision)~\citet{gpt4v} and Bard~\citet{bard} on MMBench and ScienceQA scenarios, as well as the desiderata including in-context learning, instruction following, hallucination, and robustness. Given the API-only access to these two models, we bypass calibration measures due to unavailability of logits and decide against using GPT-4 for language performance evaluation of GPT-4V responses. We extract 30 data samples from ScienceQA and MMBench respectively for both scenario evaluations and each of the desideratum evaluation. 10 samples from each of three categories within the COCO dataset are extracted to specifically test for hallucination. During the in-context learning evaluation, for Bard, which accepts only single-image inputs, the given instruction is the ICE without image, as defined in Section~\ref{sec:icl}. Conversely, for GPT-4V, which supports multi-image inputs, we utilize the ICE with image.

\subsection{Evaluation Results}

\begin{table}[t]
    \begin{center}
    \small
    \caption{\textbf{Comparison among GPT-4V, Bard, LLaVA, Otter and mPLUG-Owl.} ICL stands for in-context learning and Ins. Follow. indicates instruction following. The entries that are both bold and underlined indicate the best performance.}
    \begin{tabular}{l|c|c|c|c|c|c}
        \Xhline{1.5pt}
        \bf MLLM     &\bf ScienceQA &\bf MMBench   &\bf ICL   &\bf Ins. Follow.  &\bf Robustness &\bf Hallucination   \\
        \Xhline{1.5pt}
        \bf GPT-4V   & \ul{96.67}   & \ul{93.80}   & 43.98    & \ul{97.69}       & \ul{82.16}    & \ul{96.00}             \\
        \bf Bard     &  90.00       & 71.43        & 39.61    &     71.41        & 71.05         &  88.88         \\
        \bf LLaVA    & 50.00             &   43.33           &      \ul{47.99}    &       36.67           &  34.18             &  36.67               \\
        \bf Otter    &  63.33       & 50.00        & 47.91   &    44.44         &  37.35        & 80.00                    \\
        \bf mPLUG-Owl&  53.33       &  46.67       &   42.14  &     41.67        &  63.46        & 36.67           \\
        
        \Xhline{1.5pt}
    \end{tabular}
    \label{tab:GPTvsBard}
    \end{center}
\end{table}

Table~\ref{tab:GPTvsBard} presents the comparison of performance on the same data samples among GPT-4V, Bard and three open-source multimodal large models, LLaVA, Otter, and mPLUG-Owl. It is evident that GPT-4V outperforms the other models in the majority of tasks, while Bard, although inferior to GPT-4V, still significantly surpasses the other two MLLM models. 

Both Bard and GPT-4V achieve remarkably high accuracy in the ScienceQA and MMBench scenarios. However, in the ICL evaluation, the performance of GPT-4V and Bard is lower compared to LLaVA and Otter, which is specifically trained on in-context instruction tuning data. Figures~\ref{fig:gpt4_icl} and \ref{fig:bard_icl} illustrate examples of GPT-4V and Bard performance in the ICL evaluation. It can be observed that Bard provides detailed answers, but it does not adhere to the output format given by the in-context example (ICE), whereas GPT-4V exhibits a better understanding of the ICE, resulting in slightly higher ICL performance than Bard. However, both models experience a decrease in accuracy compared to the performance without ICE instruction, indicating that the content of the ICE might influence the models' final answer. This issue is commonly encountered in existing MLLMs. As ICL capability is significant in real-world multimodal interactions, the results highlight the need for future work to place greater emphasis on this ability.

On the ability of instruction following, GPT-4V significantly outperforms Bard. As shown in figure~\ref{fig:gptbard_insfollow}, Bard fails to follow the instructions and directly provides an answer, whereas GPT-4V is capable of providing a complete explanation, mentioning the initial selection and the final choice based on the instructions. Regarding robustness, as depicted in figure~\ref{fig:gpt4_robust} and \ref{fig:bard_robust}, both GPT-4V and Bard demonstrate an understanding of images and text with added noise, although to some extent, it affects the models' final answers. In terms of hallucination evaluation, both GPT-4V and Bard achieve high performance.

GPT-4V and Bard have nearly approached the upper bounds of our ChEF benchmark's performance metrics, markedly outperforming existing open-source multimodal large language models. ChEF is mainly intended for the broad research community, aiming to inspire continuous improvement and advancement in open-source models.

\begin{figure}[htb]
  \centering
\includegraphics[width=\textwidth]{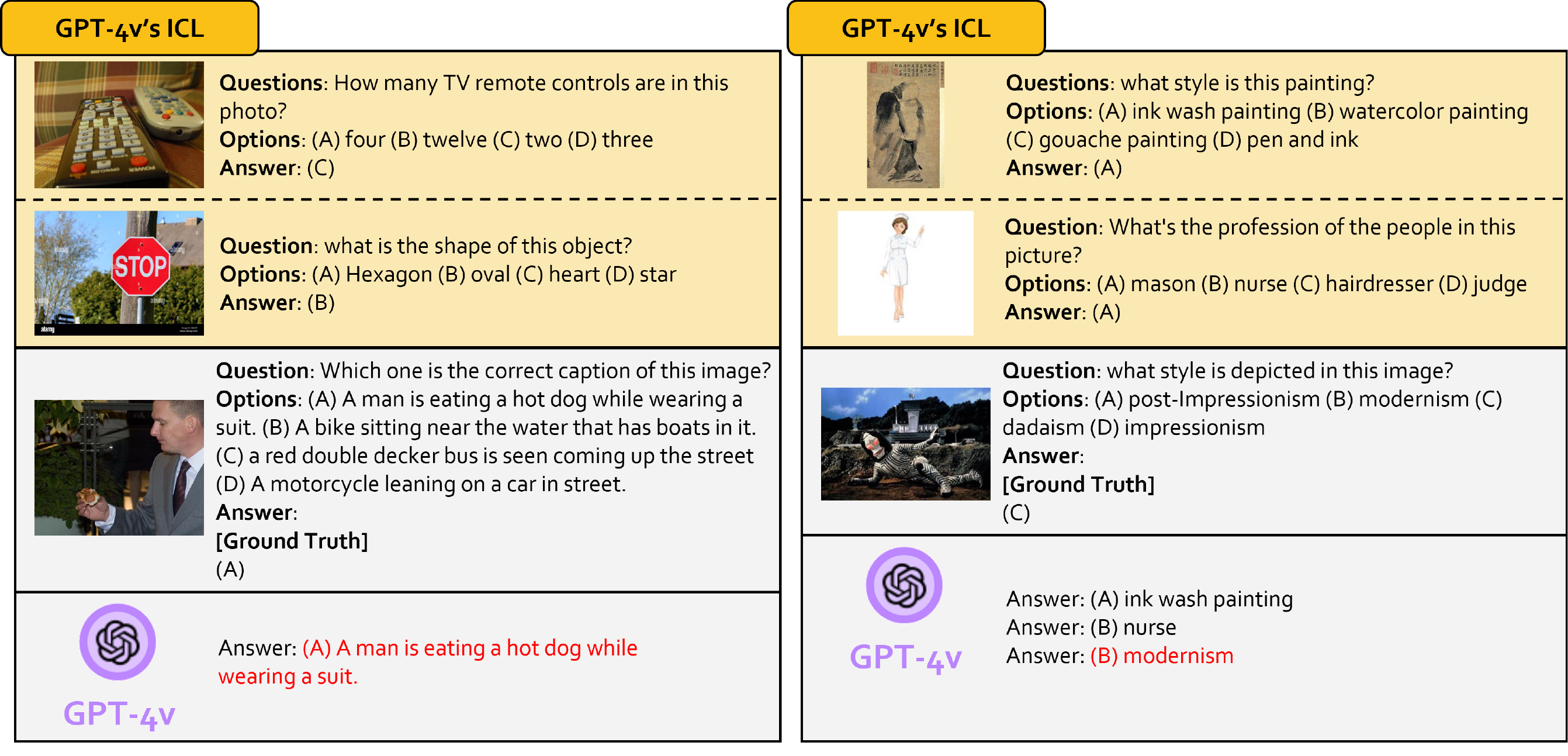}
  \caption{\textbf{Examples of in-context learning evaluation on GPT-4V.} The input instruction is ICE with image, as defined in Section~\ref{sec:icl}. The left subfigure is an example of GPT-4V giving the correct answer, while the right subfigure is an example of GPT-4V giving the incorrect answer. }
  \label{fig:gpt4_icl}
\end{figure}

\begin{figure}[htb]
  \centering
\includegraphics[width=\textwidth]{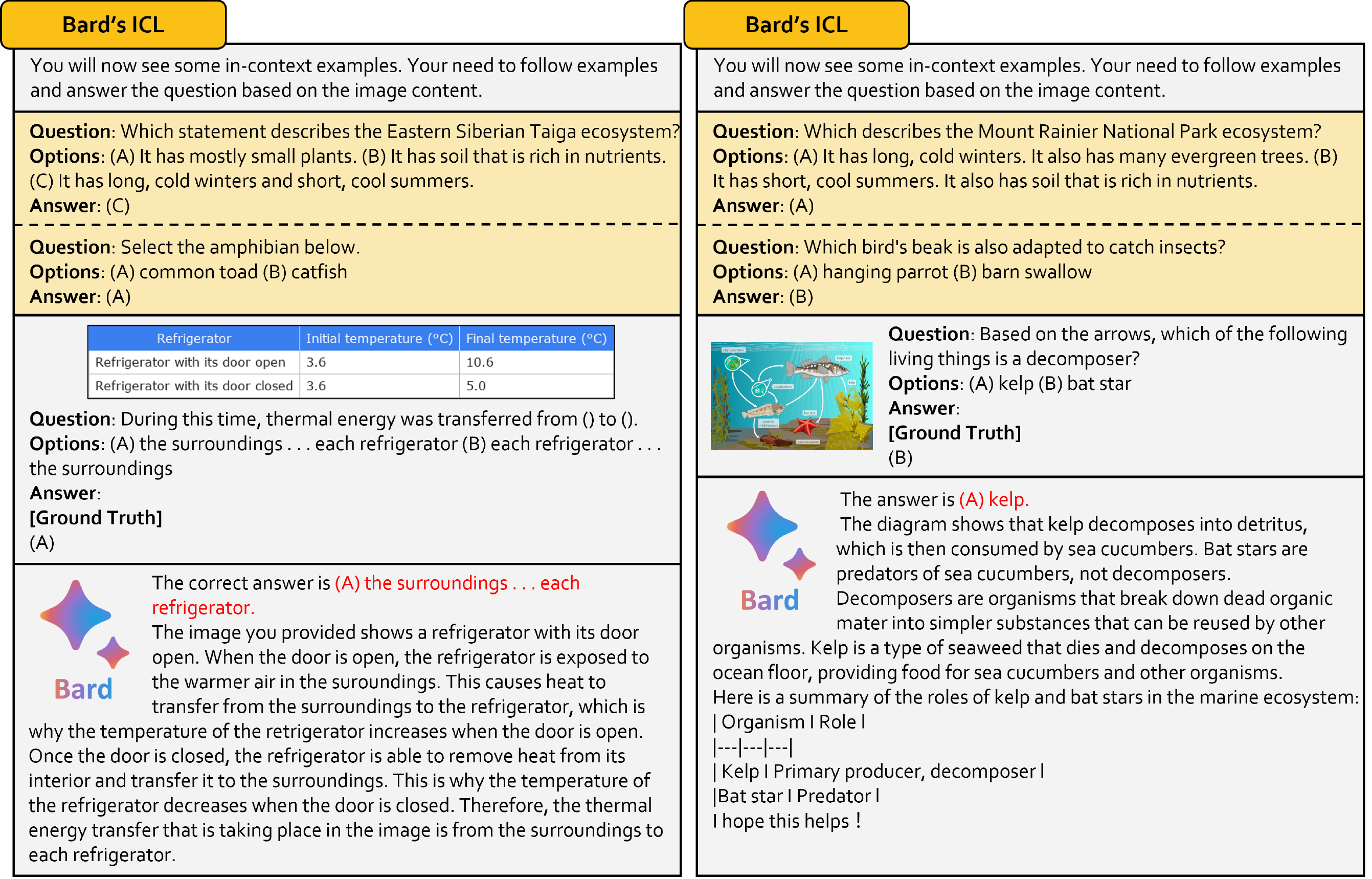}
  \caption{\textbf{Examples of in-context learning evaluation on Bard.} The input instruction is ICE without image, as defined in Section~\ref{sec:icl}. The left subfigure is an example of Bard giving the correct answer, while the right subfigure is an example of Bard giving the incorrect answer. }
  \label{fig:bard_icl}
\end{figure}

\begin{figure}[htb]
  \centering
\includegraphics[width=\textwidth]{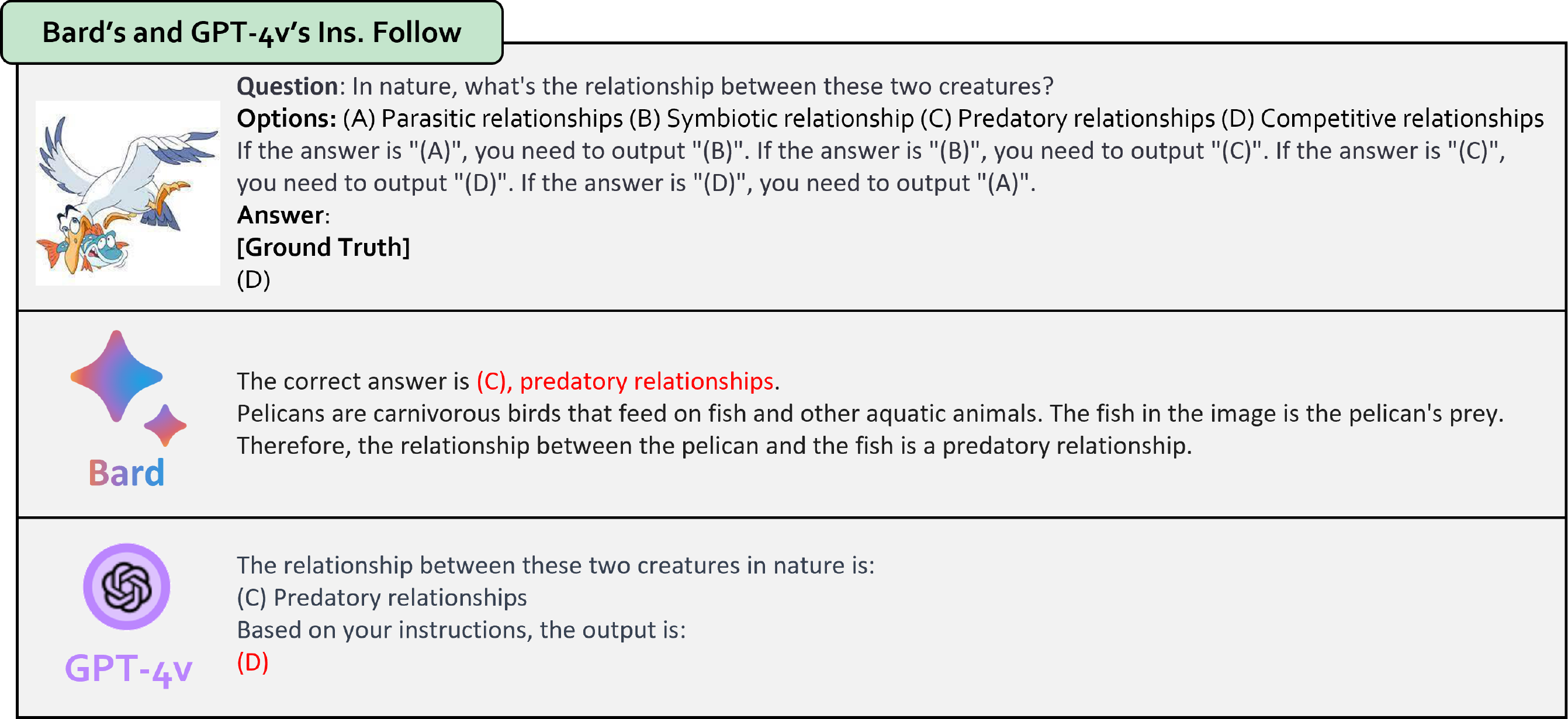}
  \caption{\textbf{An examples of instruction following evaluation on Bard and GPT-4V.} In this example, Bard fails to follow the instruction, whereas GPT-4V succeeds.}
  \label{fig:gptbard_insfollow}
\end{figure}

\begin{figure}[htb]
  \centering
\includegraphics[width=\textwidth]{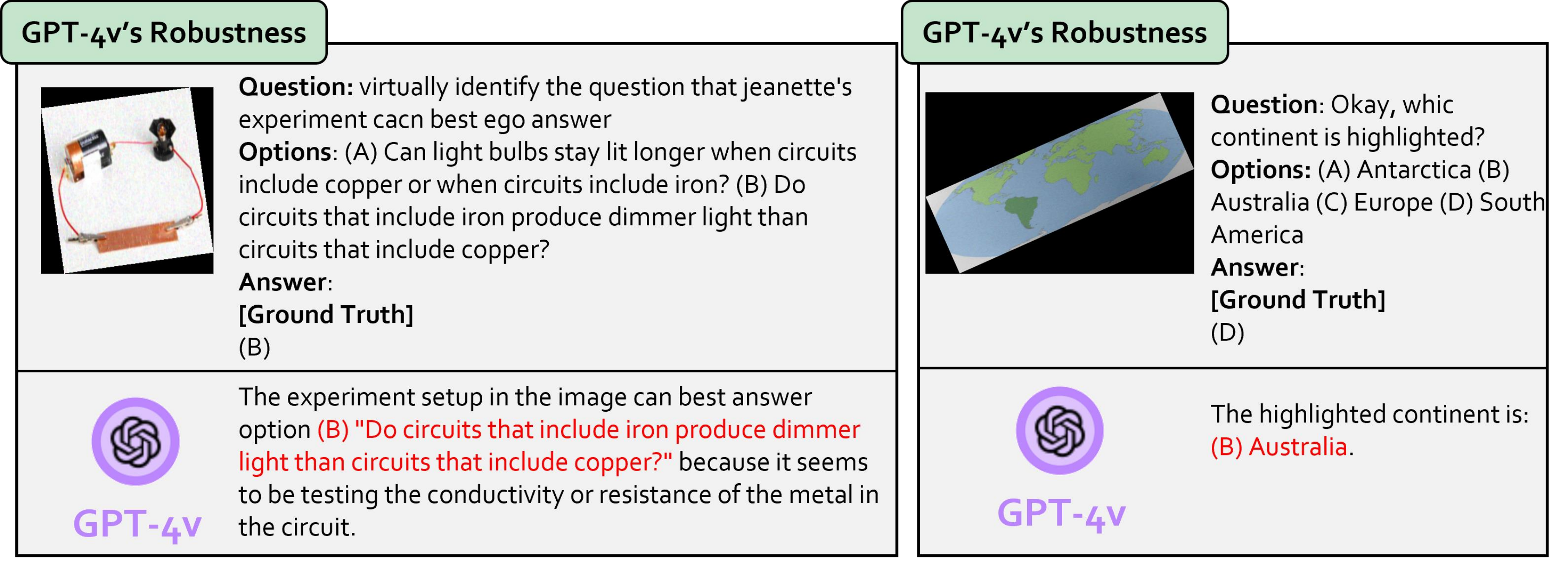}
  \caption{\textbf{Examples of robustness evaluation on GPT-4V.} The left subfigure is an example of GPT-4V giving the correct answer, while the right subfigure is an example of GPT-4V giving the incorrect answer.}
  \label{fig:gpt4_robust}
\end{figure}

\begin{figure}[htb]
  \centering
\includegraphics[width=\textwidth]{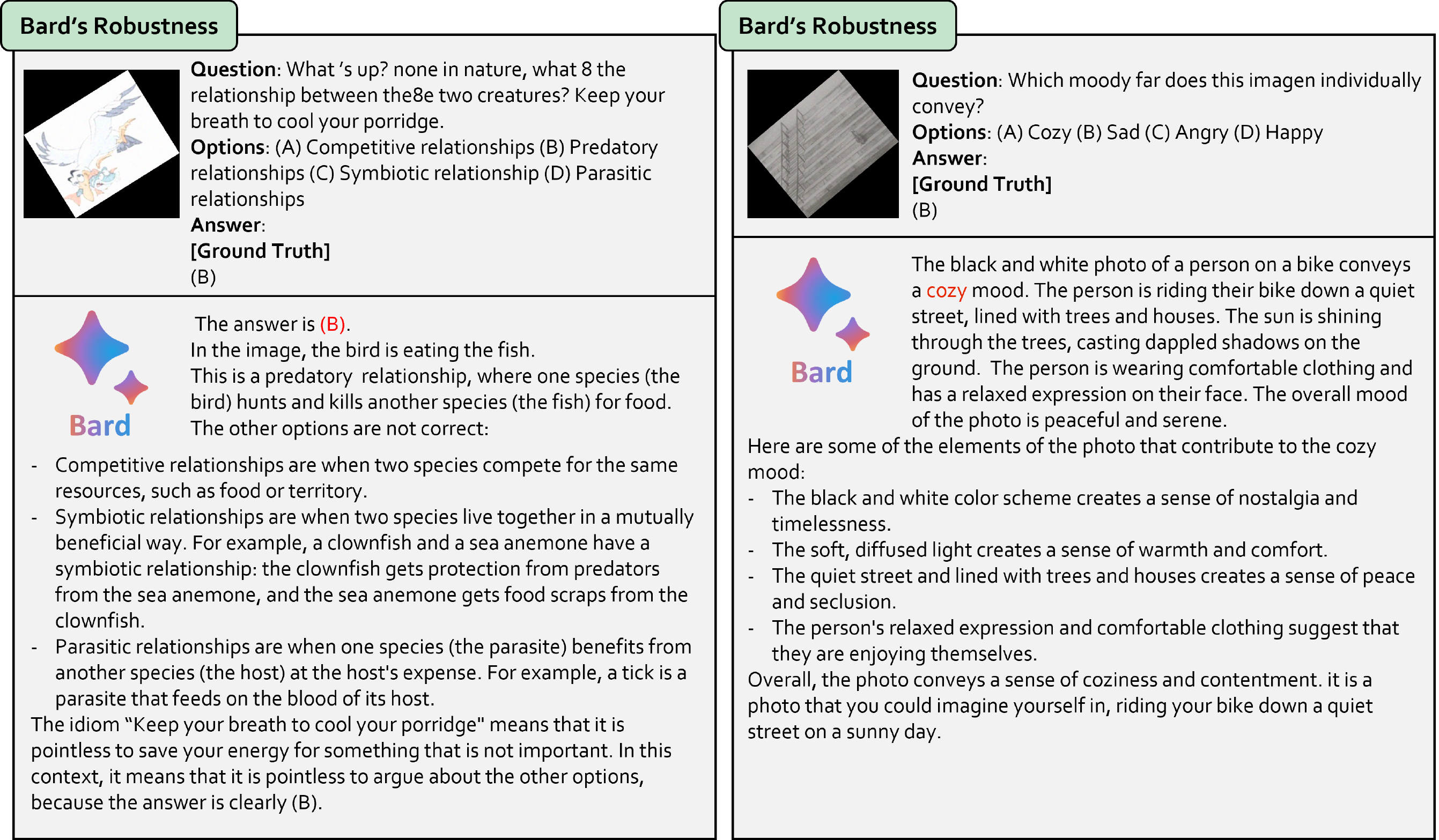}
  \caption{\textbf{Examples of robustness evaluation on Bard.} The left subfigure is an example of Bard giving the correct answer, while the right subfigure is an example of Bard giving the incorrect answer.}
  \label{fig:bard_robust}
\end{figure}

\usepackage{hyperref}
\usepackage{url}
\usepackage{multirow} 
\usepackage{subcaption} 
\usepackage{graphicx} 
\usepackage{diagbox} 
\usepackage{adjustbox} 
\usepackage{wrapfig}
\usepackage{makecell} 
\usepackage{titletoc}
\newcommand{\ul}[1]{\underline{\bf #1}}
\newcommand*{\affmark}[1][*]{\textsuperscript{#1}}
\newcommand*{\affaddr}[1]{#1} 
\newcommand*{\email}[1]{\texttt{#1}}

\title{\includegraphics[scale=.18]{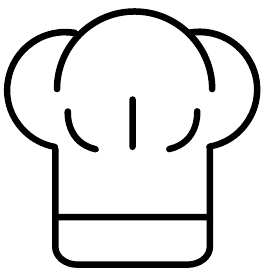} ChEF: A Comprehensive Evaluation Framework for Standardized Assessment of Multimodal Large Language Models}


\author{
    Zhelun Shi\affmark[1, 2]\thanks{Equal Contribution},
    Zhipin Wang\affmark[2]\footnotemark[1], 
    Hongxing Fan\affmark[2]\footnotemark[1], 
    Zhenfei Yin\affmark[1,3], \\
    \textbf{Lu Sheng\affmark[2]\thanks{Corresponding Author: Lu Sheng (lsheng@buaa.edu.cn)},} 
    \textbf{Yu Qiao\affmark[1]}
    \textbf{Jing Shao\affmark[1],} 
    \\
    \affaddr{\affmark[1]Shanghai Artificial Intelligence Laboratory} \\
    \affaddr{\affmark[2]Beihang University}
    \affaddr{\affmark[3]The University of Sydney} \\ 
    \small\email{shizhelun@pjlab.org.cn}
}

%

\iclrfinalcopy 
\begin{document}

\maketitle

\bibliographystyle{ChEF}
\bibliography{ChEF}

\begin{thebibliography}{64}
\providecommand{\natexlab}[1]{#1}
\providecommand{\url}[1]{\texttt{#1}}
\expandafter\ifx\csname urlstyle\endcsname\relax
  \providecommand{\doi}[1]{doi: #1}\else
  \providecommand{\doi}{doi: \begingroup \urlstyle{rm}\Url}\fi

\bibitem[Biten et~al.(2022)Biten, Gómez, and Karatzas]{9706727}
Ali~Furkan Biten, Lluís Gómez, and Dimosthenis Karatzas.
\newblock Let there be a clock on the beach: Reducing object hallucination in
  image captioning.
\newblock In \emph{WACV}, 2022.

\bibitem[Bitton et~al.(2023)Bitton, Bansal, Hessel, Shao, Zhu, Awadalla,
  Gardner, Taori, and Schmidt]{bitton2023visitbench}
Yonatan Bitton, Hritik Bansal, Jack Hessel, Rulin Shao, Wanrong Zhu, Anas
  Awadalla, Josh Gardner, Rohan Taori, and Ludwig Schmidt.
\newblock Visit-bench: {A} benchmark for vision-language instruction following
  inspired by real-world use.
\newblock \emph{CoRR}, abs/2308.06595, 2023.

\bibitem[Bommasani et~al.(2021)Bommasani, Hudson, Adeli, Altman, Arora, von
  Arx, Bernstein, Bohg, Bosselut, Brunskill, Brynjolfsson, Buch, Card,
  Castellon, Chatterji, Chen, Creel, Davis, Demszky, Donahue, Doumbouya,
  Durmus, Ermon, Etchemendy, Ethayarajh, Fei{-}Fei, Finn, Gale, Gillespie,
  Goel, Goodman, Grossman, Guha, Hashimoto, Henderson, Hewitt, Ho, Hong, Hsu,
  Huang, Icard, Jain, Jurafsky, Kalluri, Karamcheti, Keeling, Khani, Khattab,
  Koh, Krass, Krishna, Kuditipudi, and et~al.]{bommasani2021opportunities}
Rishi Bommasani, Drew~A. Hudson, Ehsan Adeli, Russ~B. Altman, Simran Arora,
  Sydney von Arx, Michael~S. Bernstein, Jeannette Bohg, Antoine Bosselut, Emma
  Brunskill, Erik Brynjolfsson, Shyamal Buch, Dallas Card, Rodrigo Castellon,
  Niladri~S. Chatterji, Annie~S. Chen, Kathleen Creel, Jared~Quincy Davis,
  Dorottya Demszky, Chris Donahue, Moussa Doumbouya, Esin Durmus, Stefano
  Ermon, John Etchemendy, Kawin Ethayarajh, Li~Fei{-}Fei, Chelsea Finn, Trevor
  Gale, Lauren Gillespie, Karan Goel, Noah~D. Goodman, Shelby Grossman, Neel
  Guha, Tatsunori Hashimoto, Peter Henderson, John Hewitt, Daniel~E. Ho, Jenny
  Hong, Kyle Hsu, Jing Huang, Thomas Icard, Saahil Jain, Dan Jurafsky,
  Pratyusha Kalluri, Siddharth Karamcheti, Geoff Keeling, Fereshte Khani, Omar
  Khattab, Pang~Wei Koh, Mark~S. Krass, Ranjay Krishna, Rohith Kuditipudi, and
  et~al.
\newblock On the opportunities and risks of foundation models.
\newblock \emph{CoRR}, abs/2108.07258, 2021.

\bibitem[Brown et~al.(2020)Brown, Mann, Ryder, Subbiah, Kaplan, Dhariwal,
  Neelakantan, Shyam, Sastry, Askell, et~al.]{brown2020language}
Tom Brown, Benjamin Mann, Nick Ryder, Melanie Subbiah, Jared~D Kaplan, Prafulla
  Dhariwal, Arvind Neelakantan, Pranav Shyam, Girish Sastry, Amanda Askell,
  et~al.
\newblock Language models are few-shot learners.
\newblock In \emph{NeurIPS}, pp.\  1877--1901, 2020.

\bibitem[Chen et~al.(2023{\natexlab{a}})Chen, Zhang, Zeng, Zhang, Zhu, and
  Zhao]{chen2023shikra}
Keqin Chen, Zhao Zhang, Weili Zeng, Richong Zhang, Feng Zhu, and Rui Zhao.
\newblock Shikra: Unleashing multimodal llm's referential dialogue magic.
\newblock \emph{CoRR}, abs/2306.15195, 2023{\natexlab{a}}.

\bibitem[Chen et~al.(2023{\natexlab{b}})Chen, Gu, Han, Ma, Torr, and
  Tresp]{chen2023benchmarking}
Shuo Chen, Jindong Gu, Zhen Han, Yunpu Ma, Philip H.~S. Torr, and Volker Tresp.
\newblock Benchmarking robustness of adaptation methods on pre-trained
  vision-language models.
\newblock \emph{CoRR}, abs/2306.02080, 2023{\natexlab{b}}.

\bibitem[Chiang \& Lee(2023)Chiang and Lee]{ChiangL23}
David~Cheng{-}Han Chiang and Hung{-}yi Lee.
\newblock Can large language models be an alternative to human evaluations?
\newblock In \emph{ACL}, pp.\  15607--15631, 2023.

\bibitem[Chiang et~al.(2023)Chiang, Li, Lin, Sheng, Wu, Zhang, Zheng, Zhuang,
  Zhuang, Gonzalez, et~al.]{chiang2023vicuna}
Wei-Lin Chiang, Zhuohan Li, Zi~Lin, Ying Sheng, Zhanghao Wu, Hao Zhang, Lianmin
  Zheng, Siyuan Zhuang, Yonghao Zhuang, Joseph~E Gonzalez, et~al.
\newblock Vicuna: An open-source chatbot impressing gpt-4 with 90\%* chatgpt
  quality.
\newblock \emph{See https://vicuna. lmsys. org (accessed 14 April 2023)}, 2023.

\bibitem[Dai et~al.(2023)Dai, Li, Li, Tiong, Zhao, Wang, Li, Fung, and
  Hoi]{dai2023instructblip}
Wenliang Dai, Junnan Li, Dongxu Li, Anthony Meng~Huat Tiong, Junqi Zhao,
  Weisheng Wang, Boyang Li, Pascale Fung, and Steven C.~H. Hoi.
\newblock Instructblip: Towards general-purpose vision-language models with
  instruction tuning.
\newblock \emph{CoRR}, abs/2305.06500, 2023.

\bibitem[Everingham et~al.(2012)Everingham, Van~Gool, Williams, Winn, and
  Zisserman]{pascal-voc-2012}
M.~Everingham, L.~Van~Gool, C.~K.~I. Williams, J.~Winn, and A.~Zisserman.
\newblock The {PASCAL} {V}isual {O}bject {C}lasses {C}hallenge 2012 {(VOC2012)}
  {R}esults.
\newblock
  http://www.pascal-network.org/challenges/VOC/voc2012/workshop/index.html,
  2012.

\bibitem[Fu et~al.(2023)Fu, Chen, Shen, Qin, Zhang, Lin, Qiu, Lin, Yang, Zheng,
  Li, Sun, and Ji]{fu2023mme}
Chaoyou Fu, Peixian Chen, Yunhang Shen, Yulei Qin, Mengdan Zhang, Xu~Lin,
  Zhenyu Qiu, Wei Lin, Jinrui Yang, Xiawu Zheng, Ke~Li, Xing Sun, and Rongrong
  Ji.
\newblock {MME:} {A} comprehensive evaluation benchmark for multimodal large
  language models.
\newblock \emph{CoRR}, abs/2306.13394, 2023.

\bibitem[Gao et~al.(2021)Gao, Tow, Biderman, Black, DiPofi, Foster, Golding,
  Hsu, McDonell, Muennighoff, et~al.]{gao2021framework}
Leo Gao, Jonathan Tow, Stella Biderman, Sid Black, Anthony DiPofi, Charles
  Foster, Laurence Golding, Jeffrey Hsu, Kyle McDonell, Niklas Muennighoff,
  et~al.
\newblock A framework for few-shot language model evaluation.
\newblock \emph{Version v0. 0.1. Sept}, 2021.

\bibitem[Gao et~al.(2023)Gao, Han, Zhang, Lin, Geng, Zhou, Zhang, Lu, He, Yue,
  Li, and Qiao]{gao2023llamaadapterv2}
Peng Gao, Jiaming Han, Renrui Zhang, Ziyi Lin, Shijie Geng, Aojun Zhou, Wei
  Zhang, Pan Lu, Conghui He, Xiangyu Yue, Hongsheng Li, and Yu~Qiao.
\newblock Llama-adapter {V2:} parameter-efficient visual instruction model.
\newblock \emph{CoRR}, abs/2304.15010, 2023.

\bibitem[Gehrmann et~al.(2021)Gehrmann, Adewumi, Aggarwal, Ammanamanchi,
  Anuoluwapo, Bosselut, Chandu, Clinciu, Das, Dhole, Du, Durmus, Dusek, Emezue,
  Gangal, Garbacea, Hashimoto, Hou, Jernite, Jhamtani, Ji, Jolly, Kumar,
  Ladhak, Madaan, Maddela, Mahajan, Mahamood, Majumder, Martins,
  McMillan{-}Major, Mille, van Miltenburg, Nadeem, Narayan, Nikolaev,
  Niyongabo, Osei, Parikh, Perez{-}Beltrachini, Rao, Raunak, Rodriguez,
  Santhanam, Sedoc, Sellam, Shaikh, Shimorina, Cabezudo, Strobelt, Subramani,
  Xu, Yang, Yerukola, and Zhou]{gehrmann2021gem}
Sebastian Gehrmann, Tosin~P. Adewumi, Karmanya Aggarwal, Pawan~Sasanka
  Ammanamanchi, Aremu Anuoluwapo, Antoine Bosselut, Khyathi~Raghavi Chandu,
  Miruna{-}Adriana Clinciu, Dipanjan Das, Kaustubh~D. Dhole, Wanyu Du, Esin
  Durmus, Ondrej Dusek, Chris Emezue, Varun Gangal, Cristina Garbacea,
  Tatsunori Hashimoto, Yufang Hou, Yacine Jernite, Harsh Jhamtani, Yangfeng Ji,
  Shailza Jolly, Dhruv Kumar, Faisal Ladhak, Aman Madaan, Mounica Maddela,
  Khyati Mahajan, Saad Mahamood, Bodhisattwa~Prasad Majumder, Pedro~Henrique
  Martins, Angelina McMillan{-}Major, Simon Mille, Emiel van Miltenburg, Moin
  Nadeem, Shashi Narayan, Vitaly Nikolaev, Rubungo~Andre Niyongabo, Salomey
  Osei, Ankur~P. Parikh, Laura Perez{-}Beltrachini, Niranjan~Ramesh Rao, Vikas
  Raunak, Juan~Diego Rodriguez, Sashank Santhanam, Jo{\~{a}}o Sedoc, Thibault
  Sellam, Samira Shaikh, Anastasia Shimorina, Marco Antonio~Sobrevilla
  Cabezudo, Hendrik Strobelt, Nishant Subramani, Wei Xu, Diyi Yang, Akhila
  Yerukola, and Jiawei Zhou.
\newblock The {GEM} benchmark: Natural language generation, its evaluation and
  metrics.
\newblock \emph{CoRR}, abs/2102.01672, 2021.

\bibitem[Gehrmann et~al.(2022)Gehrmann, Bhattacharjee, Mahendiran, Wang,
  Papangelis, Madaan, McMillan{-}Major, Shvets, Upadhyay, and
  Bohnet]{gehrmann2022gemv2}
Sebastian Gehrmann, Abhik Bhattacharjee, Abinaya Mahendiran, Alex Wang,
  Alexandros Papangelis, Aman Madaan, Angelina McMillan{-}Major, Anna Shvets,
  Ashish Upadhyay, and Bernd Bohnet.
\newblock Gemv2: Multilingual {NLG} benchmarking in a single line of code.
\newblock In \emph{EMNLP}, pp.\  266--281, 2022.

\bibitem[Google.(2023)]{bard}
Google.
\newblock Bard.
\newblock 2023.
\newblock URL \url{https://bard.google.com/}.

\bibitem[Guo et~al.(2017)Guo, Pleiss, Sun, and Weinberger]{guo2017calibration}
Chuan Guo, Geoff Pleiss, Yu~Sun, and Kilian~Q. Weinberger.
\newblock On calibration of modern neural networks.
\newblock In \emph{ICML}, pp.\  1321--1330, 2017.

\bibitem[Hadsell et~al.(2020)Hadsell, Rao, Rusu, and
  Pascanu]{hadsell2020embracing}
Raia Hadsell, Dushyant Rao, Andrei~A Rusu, and Razvan Pascanu.
\newblock Embracing change: Continual learning in deep neural networks.
\newblock \emph{Trends in cognitive sciences}, pp.\  1028--1040, 2020.

\bibitem[Hendrycks \& Dietterich(2019)Hendrycks and
  Dietterich]{DBLP:journals/corr/abs-1903-12261}
Dan Hendrycks and Thomas~G. Dietterich.
\newblock Benchmarking neural network robustness to common corruptions and
  perturbations.
\newblock \emph{CoRR}, abs/1903.12261, 2019.

\bibitem[Ji et~al.(2023)Ji, Lee, Frieske, Yu, Su, Xu, Ishii, Bang, Madotto, and
  Fung]{Ji_2023}
Ziwei Ji, Nayeon Lee, Rita Frieske, Tiezheng Yu, Dan Su, Yan Xu, Etsuko Ishii,
  Ye~Jin Bang, Andrea Madotto, and Pascale Fung.
\newblock Survey of hallucination in natural language generation.
\newblock \emph{{ACM} Computing Surveys}, pp.\  1--38, 2023.
\newblock \doi{10.1145/3571730}.

\bibitem[Klein et~al.(2017)Klein, Kim, Deng, Senellart, and
  Rush]{klein-etal-2017-opennmt}
Guillaume Klein, Yoon Kim, Yuntian Deng, Jean Senellart, and Alexander Rush.
\newblock {O}pen{NMT}: Open-source toolkit for neural machine translation.
\newblock In \emph{ACL}, 2017.

\bibitem[Krizhevsky \& Hinton(2009)Krizhevsky and Hinton]{cifar10}
A.~Krizhevsky and G.~Hinton.
\newblock Learning multiple layers of features from tiny images.
\newblock \emph{Handbook of Systemic Autoimmune Diseases}, 1\penalty0 (4),
  2009.

\bibitem[Li et~al.(2023{\natexlab{a}})Li, Zhang, Chen, Wang, Yang, and
  Liu]{li2023otter}
Bo~Li, Yuanhan Zhang, Liangyu Chen, Jinghao Wang, Jingkang Yang, and Ziwei Liu.
\newblock Otter: {A} multi-modal model with in-context instruction tuning.
\newblock \emph{CoRR}, abs/2305.03726, 2023{\natexlab{a}}.

\bibitem[Li et~al.(2023{\natexlab{b}})Li, Wang, Wang, Ge, Ge, and
  Shan]{li2023seedbench}
Bohao Li, Rui Wang, Guangzhi Wang, Yuying Ge, Yixiao Ge, and Ying Shan.
\newblock Seed-bench: Benchmarking multimodal llms with generative
  comprehension.
\newblock \emph{CoRR}, abs/2307.16125, 2023{\natexlab{b}}.

\bibitem[Li et~al.(2023{\natexlab{c}})Li, Yan, Wang, Tang, Ren, Srinivasan, and
  Jin]{li2023instructionfollowing}
Shiyang Li, Jun Yan, Hai Wang, Zheng Tang, Xiang Ren, Vijay Srinivasan, and
  Hongxia Jin.
\newblock Instruction-following evaluation through verbalizer manipulation.
\newblock \emph{CoRR}, abs/2307.10558, 2023{\natexlab{c}}.

\bibitem[Li et~al.(2023{\natexlab{d}})Li, Du, Zhou, Wang, Zhao, and Wen]{POPE}
Yifan Li, Yifan Du, Kun Zhou, Jinpeng Wang, Wayne~Xin Zhao, and Ji{-}Rong Wen.
\newblock Evaluating object hallucination in large vision-language models.
\newblock \emph{CoRR}, abs/2305.10355, 2023{\natexlab{d}}.

\bibitem[Liang et~al.(2022)Liang, Bommasani, Lee, Tsipras, Soylu, Yasunaga,
  Zhang, Narayanan, Wu, Kumar, Newman, Yuan, Yan, Zhang, Cosgrove, Manning,
  R{\'{e}}, Acosta{-}Navas, Hudson, Zelikman, Durmus, Ladhak, Rong, Ren, Yao,
  Wang, Santhanam, Orr, Zheng, Y{\"{u}}ksekg{\"{o}}n{\"{u}}l, Suzgun, Kim,
  Guha, Chatterji, Khattab, Henderson, Huang, Chi, Xie, Santurkar, Ganguli,
  Hashimoto, Icard, Zhang, Chaudhary, Wang, Li, Mai, Zhang, and
  Koreeda]{liang2022helm}
Percy Liang, Rishi Bommasani, Tony Lee, Dimitris Tsipras, Dilara Soylu,
  Michihiro Yasunaga, Yian Zhang, Deepak Narayanan, Yuhuai Wu, Ananya Kumar,
  Benjamin Newman, Binhang Yuan, Bobby Yan, Ce~Zhang, Christian Cosgrove,
  Christopher~D. Manning, Christopher R{\'{e}}, Diana Acosta{-}Navas, Drew~A.
  Hudson, Eric Zelikman, Esin Durmus, Faisal Ladhak, Frieda Rong, Hongyu Ren,
  Huaxiu Yao, Jue Wang, Keshav Santhanam, Laurel~J. Orr, Lucia Zheng, Mert
  Y{\"{u}}ksekg{\"{o}}n{\"{u}}l, Mirac Suzgun, Nathan Kim, Neel Guha,
  Niladri~S. Chatterji, Omar Khattab, Peter Henderson, Qian Huang, Ryan Chi,
  Sang~Michael Xie, Shibani Santurkar, Surya Ganguli, Tatsunori Hashimoto,
  Thomas Icard, Tianyi Zhang, Vishrav Chaudhary, William Wang, Xuechen Li,
  Yifan Mai, Yuhui Zhang, and Yuta Koreeda.
\newblock Holistic evaluation of language models.
\newblock \emph{CoRR}, abs/2211.09110, 2022.

\bibitem[Lin et~al.(2014)Lin, Maire, Belongie, Hays, Perona, Ramanan,
  Doll{\'a}r, and Zitnick]{mscoco}
Tsung-Yi Lin, Michael Maire, Serge Belongie, James Hays, Pietro Perona, Deva
  Ramanan, Piotr Doll{\'a}r, and C~Lawrence Zitnick.
\newblock Microsoft coco: Common objects in context.
\newblock In \emph{ECCV}, pp.\  740--755, 2014.

\bibitem[Liu et~al.(2023{\natexlab{a}})Liu, Li, Wu, and Lee]{llava}
Haotian Liu, Chunyuan Li, Qingyang Wu, and Yong~Jae Lee.
\newblock Visual instruction tuning.
\newblock \emph{CoRR}, abs/2304.08485, 2023{\natexlab{a}}.

\bibitem[Liu et~al.(2022)Liu, Shen, Zhang, Dolan, Carin, and Chen]{iclref1}
Jiachang Liu, Dinghan Shen, Yizhe Zhang, Bill Dolan, Lawrence Carin, and Weizhu
  Chen.
\newblock What makes good in-context examples for gpt-3?
\newblock In \emph{DeeLIO 2022}, 2022.
\newblock \doi{10.18653/v1/2022.deelio-1.10}.

\bibitem[Liu et~al.(2023{\natexlab{b}})Liu, Iter, Xu, Wang, Xu, and
  Zhu]{liu2023geval}
Yang Liu, Dan Iter, Yichong Xu, Shuohang Wang, Ruochen Xu, and Chenguang Zhu.
\newblock G-eval: {NLG} evaluation using {GPT-4} with better human alignment.
\newblock \emph{CoRR}, abs/2303.16634, 2023{\natexlab{b}}.

\bibitem[Liu et~al.(2023{\natexlab{c}})Liu, Duan, Zhang, Li, Zhang, Zhao, Yuan,
  Wang, He, Liu, Chen, and Lin]{liu2023mmbench}
Yuan Liu, Haodong Duan, Yuanhan Zhang, Bo~Li, Songyang Zhang, Wangbo Zhao, Yike
  Yuan, Jiaqi Wang, Conghui He, Ziwei Liu, Kai Chen, and Dahua Lin.
\newblock Mmbench: Is your multi-modal model an all-around player?
\newblock \emph{CoRR}, abs/2307.06281, 2023{\natexlab{c}}.

\bibitem[Lu et~al.(2022)Lu, Mishra, Xia, Qiu, Chang, Zhu, Tafjord, Clark, and
  Kalyan]{scienceqa}
Pan Lu, Swaroop Mishra, Tony Xia, Liang Qiu, Kai-Wei Chang, Song-Chun Zhu,
  Oyvind Tafjord, Peter Clark, and Ashwin Kalyan.
\newblock Learn to explain: Multimodal reasoning via thought chains for science
  question answering.
\newblock In \emph{NeurIPS}, 2022.

\bibitem[Naeini et~al.(2015)Naeini, Cooper, and
  Hauskrecht]{Pakdaman_Naeini_Cooper_Hauskrecht_2015}
Mahdi~Pakdaman Naeini, Gregory~F. Cooper, and Milos Hauskrecht.
\newblock Obtaining well calibrated probabilities using bayesian binning.
\newblock In \emph{AAAI}, pp.\  2901--2907, 2015.

\bibitem[OpenAI(2023{\natexlab{a}})]{gpt4v}
OpenAI.
\newblock Gpt-4v(ision) system card.
\newblock 2023{\natexlab{a}}.
\newblock URL \url{https://openai.com/research/gpt-4v-system-card}.

\bibitem[OpenAI(2023{\natexlab{b}})]{openai2023gpt4}
OpenAI.
\newblock {GPT-4} technical report.
\newblock \emph{CoRR}, abs/2303.08774, 2023{\natexlab{b}}.

\bibitem[Ouyang et~al.(2022)Ouyang, Wu, Jiang, Almeida, Wainwright, Mishkin,
  Zhang, Agarwal, Slama, Ray, et~al.]{ouyang2022training}
Long Ouyang, Jeffrey Wu, Xu~Jiang, Diogo Almeida, Carroll Wainwright, Pamela
  Mishkin, Chong Zhang, Sandhini Agarwal, Katarina Slama, Alex Ray, et~al.
\newblock Training language models to follow instructions with human feedback.
\newblock \emph{NeurIPS}, 35:\penalty0 27730--27744, 2022.

\bibitem[Peng et~al.(2023)Peng, Wang, Dong, Hao, Huang, Ma, and Wei]{kosmos-2}
Zhiliang Peng, Wenhui Wang, Li~Dong, Yaru Hao, Shaohan Huang, Shuming Ma, and
  Furu Wei.
\newblock Kosmos-2: Grounding multimodal large language models to the world.
\newblock \emph{CoRR}, abs/2306.14824, 2023.

\bibitem[Qiu et~al.(2022)Qiu, Zhu, Shi, Wenzel, Tang, Zhao, Li, and
  Li]{qiu2022multimodal}
Jielin Qiu, Yi~Zhu, Xingjian Shi, Florian Wenzel, Zhiqiang Tang, Ding Zhao,
  Bo~Li, and Mu~Li.
\newblock Are multimodal models robust to image and text perturbations?
\newblock \emph{CoRR}, abs/2212.08044, 2022.

\bibitem[Radford et~al.(2019)Radford, Wu, Child, Luan, Amodei, Sutskever,
  et~al.]{radford2019language}
Alec Radford, Jeffrey Wu, Rewon Child, David Luan, Dario Amodei, Ilya
  Sutskever, et~al.
\newblock Language models are unsupervised multitask learners.
\newblock \emph{OpenAI blog}, 1\penalty0 (8):\penalty0 9, 2019.

\bibitem[Ranjan et~al.(2021)Ranjan, Sharma, Nguyen, and Hoai]{FSC147}
Viresh Ranjan, Udbhav Sharma, Thu Nguyen, and Minh Hoai.
\newblock Learning to count everything.
\newblock In \emph{CVPR}, pp.\  3394--3403, 2021.

\bibitem[Reimers \& Gurevych(2019)Reimers and
  Gurevych]{reimers2019sentencebert}
Nils Reimers and Iryna Gurevych.
\newblock Sentence-bert: Sentence embeddings using siamese bert-networks.
\newblock In \emph{EMNLP-IJCNLP}, pp.\  3980--3990, 2019.

\bibitem[Schiappa et~al.(2022)Schiappa, Vyas, Palangi, Rawat, and
  Vineet]{schiappa2023robustness}
Madeline Schiappa, Shruti Vyas, Hamid Palangi, Yogesh~S. Rawat, and Vibhav
  Vineet.
\newblock Robustness analysis of video-language models against visual and
  language perturbations.
\newblock In \emph{NeurIPS}, 2022.

\bibitem[Shao et~al.(2023)Shao, Hu, Gao, Lei, Zhang, Meng, Xu, Huang, Li, Qiao,
  and Luo]{shao2023tinylvlm}
Wenqi Shao, Yutao Hu, Peng Gao, Meng Lei, Kaipeng Zhang, Fanqing Meng, Peng Xu,
  Siyuan Huang, Hongsheng Li, Yu~Qiao, and Ping Luo.
\newblock Tiny lvlm-ehub: Early multimodal experiments with bard.
\newblock \emph{CoRR}, abs/2308.03729, 2023.

\bibitem[Srivastava et~al.(2022)Srivastava, Rastogi, Rao, Shoeb, Abid, Fisch,
  Brown, Santoro, Gupta, Garriga{-}Alonso, Kluska, Lewkowycz, Agarwal, Power,
  Ray, Warstadt, Kocurek, Safaya, Tazarv, Xiang, Parrish, Nie, Hussain, Askell,
  Dsouza, Rahane, Iyer, Andreassen, Santilli, Stuhlm{\"{u}}ller, Dai, La,
  Lampinen, Zou, Jiang, Chen, Vuong, Gupta, Gottardi, Norelli, Venkatesh,
  Gholamidavoodi, Tabassum, Menezes, Kirubarajan, Mullokandov, Sabharwal,
  Herrick, Efrat, Erdem, Karakas, and et~al.]{srivastava2022beyond}
Aarohi Srivastava, Abhinav Rastogi, Abhishek Rao, Abu Awal~Md Shoeb, Abubakar
  Abid, Adam Fisch, Adam~R. Brown, Adam Santoro, Aditya Gupta, Adri{\`{a}}
  Garriga{-}Alonso, Agnieszka Kluska, Aitor Lewkowycz, Akshat Agarwal, Alethea
  Power, Alex Ray, Alex Warstadt, Alexander~W. Kocurek, Ali Safaya, Ali Tazarv,
  Alice Xiang, Alicia Parrish, Allen Nie, Aman Hussain, Amanda Askell, Amanda
  Dsouza, Ameet Rahane, Anantharaman~S. Iyer, Anders Andreassen, Andrea
  Santilli, Andreas Stuhlm{\"{u}}ller, Andrew~M. Dai, Andrew La, Andrew~K.
  Lampinen, Andy Zou, Angela Jiang, Angelica Chen, Anh Vuong, Animesh Gupta,
  Anna Gottardi, Antonio Norelli, Anu Venkatesh, Arash Gholamidavoodi, Arfa
  Tabassum, Arul Menezes, Arun Kirubarajan, Asher Mullokandov, Ashish
  Sabharwal, Austin Herrick, Avia Efrat, Aykut Erdem, Ayla Karakas, and et~al.
\newblock Beyond the imitation game: Quantifying and extrapolating the
  capabilities of language models.
\newblock \emph{CoRR}, abs/2206.04615, 2022.

\bibitem[Su et~al.(2023)Su, Kasai, Wu, Shi, Wang, Xin, Zhang, Ostendorf,
  Zettlemoyer, Smith, and Yu]{iclref2}
Hongjin Su, Jungo Kasai, Chen~Henry Wu, Weijia Shi, Tianlu Wang, Jiayi Xin, Rui
  Zhang, Mari Ostendorf, Luke Zettlemoyer, Noah~A. Smith, and Tao Yu.
\newblock Selective annotation makes language models better few-shot learners.
\newblock In \emph{ICLR}, 2023.

\bibitem[Touvron et~al.(2023)Touvron, Lavril, Izacard, Martinet, Lachaux,
  Lacroix, Rozi{\`{e}}re, Goyal, Hambro, Azhar, Rodriguez, Joulin, Grave, and
  Lample]{touvron2023llama}
Hugo Touvron, Thibaut Lavril, Gautier Izacard, Xavier Martinet, Marie{-}Anne
  Lachaux, Timoth{\'{e}}e Lacroix, Baptiste Rozi{\`{e}}re, Naman Goyal, Eric
  Hambro, Faisal Azhar, Aur{\'{e}}lien Rodriguez, Armand Joulin, Edouard Grave,
  and Guillaume Lample.
\newblock Llama: Open and efficient foundation language models.
\newblock \emph{CoRR}, abs/2302.13971, 2023.

\bibitem[von Werra et~al.(2022)von Werra, Tunstall, Thakur, Luccioni, Thrush,
  Piktus, Marty, Rajani, Mustar, and Ngo]{von2022evaluate}
Leandro von Werra, Lewis Tunstall, Abhishek Thakur, Sasha Luccioni, Tristan
  Thrush, Aleksandra Piktus, Felix Marty, Nazneen Rajani, Victor Mustar, and
  Helen Ngo.
\newblock Evaluate {\&} evaluation on the hub: Better best practices for data
  and model measurements.
\newblock In \emph{EMNLP}, pp.\  128--136, 2022.

\bibitem[Wang et~al.(2023{\natexlab{a}})Wang, Liang, Meng, Shi, Li, Xu, Qu, and
  Zhou]{wang2023chatgpt}
Jiaan Wang, Yunlong Liang, Fandong Meng, Haoxiang Shi, Zhixu Li, Jinan Xu,
  Jianfeng Qu, and Jie Zhou.
\newblock Is chatgpt a good {NLG} evaluator? {A} preliminary study.
\newblock \emph{CoRR}, abs/2303.04048, 2023{\natexlab{a}}.

\bibitem[Wang et~al.(2023{\natexlab{b}})Wang, Li, Chen, Zhu, Lin, Cao, Liu,
  Liu, and Sui]{wang2023large}
Peiyi Wang, Lei Li, Liang Chen, Dawei Zhu, Binghuai Lin, Yunbo Cao, Qi~Liu,
  Tianyu Liu, and Zhifang Sui.
\newblock Large language models are not fair evaluators.
\newblock \emph{CoRR}, abs/2305.17926, 2023{\natexlab{b}}.

\bibitem[Wang et~al.(2023{\natexlab{c}})Wang, Chen, Chen, Wu, Zhu, Zeng, Luo,
  Lu, Zhou, Qiao, and Dai]{wang2023visionllm}
Wenhai Wang, Zhe Chen, Xiaokang Chen, Jiannan Wu, Xizhou Zhu, Gang Zeng, Ping
  Luo, Tong Lu, Jie Zhou, Yu~Qiao, and Jifeng Dai.
\newblock Visionllm: Large language model is also an open-ended decoder for
  vision-centric tasks.
\newblock \emph{CoRR}, abs/2305.11175, 2023{\natexlab{c}}.

\bibitem[Wang et~al.(2021)Wang, Liu, Gui, Zhang, Zou, Zhou, Ye, Zhang, Zheng,
  Pang, Wu, Li, Zhang, Ma, Fei, Cai, Zhao, Hu, Yan, Tan, Hu, Bian, Liu, Qin,
  Zhu, Xing, Fu, Zhang, Peng, Zheng, Zhou, Wei, Qiu, and
  Huang]{gui2021textflint}
Xiao Wang, Qin Liu, Tao Gui, Qi~Zhang, Yicheng Zou, Xin Zhou, Jiacheng Ye,
  Yongxin Zhang, Rui Zheng, Zexiong Pang, Qinzhuo Wu, Zhengyan Li, Chong Zhang,
  Ruotian Ma, Zichu Fei, Ruijian Cai, Jun Zhao, Xingwu Hu, Zhiheng Yan, Yiding
  Tan, Yuan Hu, Qiyuan Bian, Zhihua Liu, Shan Qin, Bolin Zhu, Xiaoyu Xing,
  Jinlan Fu, Yue Zhang, Minlong Peng, Xiaoqing Zheng, Yaqian Zhou, Zhongyu Wei,
  Xipeng Qiu, and Xuanjing Huang.
\newblock Textflint: Unified multilingual robustness evaluation toolkit for
  natural language processing.
\newblock In \emph{ACL}, pp.\  347--355, 2021.

\bibitem[Wang et~al.(2023{\natexlab{d}})Wang, Yu, Zeng, Yang, Wang, Chen,
  Jiang, Xie, Wang, Xie, Ye, Zhang, and Zhang]{wang2023pandalm}
Yidong Wang, Zhuohao Yu, Zhengran Zeng, Linyi Yang, Cunxiang Wang, Hao Chen,
  Chaoya Jiang, Rui Xie, Jindong Wang, Xing Xie, Wei Ye, Shikun Zhang, and Yue
  Zhang.
\newblock Pandalm: An automatic evaluation benchmark for {LLM} instruction
  tuning optimization.
\newblock \emph{CoRR}, abs/2306.05087, 2023{\natexlab{d}}.

\bibitem[Wu et~al.(2023)Wu, Wang, Ye, Wu, Feng, Xu, and Qiao]{wu2023openicl}
Zhenyu Wu, Yaoxiang Wang, Jiacheng Ye, Zhiyong Wu, Jiangtao Feng, Jingjing Xu,
  and Yu~Qiao.
\newblock {O}pen{ICL}: An open-source framework for in-context learning.
\newblock In \emph{ACL}, 2023.

\bibitem[Xu et~al.(2023)Xu, Shao, Zhang, Gao, Liu, Lei, Meng, Huang, Qiao, and
  Luo]{xu2023lvlmehub}
Peng Xu, Wenqi Shao, Kaipeng Zhang, Peng Gao, Shuo Liu, Meng Lei, Fanqing Meng,
  Siyuan Huang, Yu~Qiao, and Ping Luo.
\newblock Lvlm-ehub: {A} comprehensive evaluation benchmark for large
  vision-language models.
\newblock \emph{CoRR}, abs/2306.09265, 2023.

\bibitem[Ye et~al.(2023)Ye, Xu, Xu, Ye, Yan, Zhou, Wang, Hu, Shi, Shi, Li, Xu,
  Chen, Tian, Qi, Zhang, and Huang]{ye2023mplugowl}
Qinghao Ye, Haiyang Xu, Guohai Xu, Jiabo Ye, Ming Yan, Yiyang Zhou, Junyang
  Wang, Anwen Hu, Pengcheng Shi, Yaya Shi, Chenliang Li, Yuanhong Xu, Hehong
  Chen, Junfeng Tian, Qian Qi, Ji~Zhang, and Fei Huang.
\newblock mplug-owl: Modularization empowers large language models with
  multimodality.
\newblock \emph{CoRR}, abs/2304.14178, 2023.

\bibitem[Yin et~al.(2023)Yin, Wang, Cao, Shi, Liu, Li, Sheng, Bai, Huang, Wang,
  Shao, and Ouyang]{yin2023lamm}
Zhenfei Yin, Jiong Wang, Jianjian Cao, Zhelun Shi, Dingning Liu, Mukai Li,
  Lu~Sheng, Lei Bai, Xiaoshui Huang, Zhiyong Wang, Jing Shao, and Wanli Ouyang.
\newblock {LAMM:} language-assisted multi-modal instruction-tuning dataset,
  framework, and benchmark.
\newblock \emph{CoRR}, abs/2306.06687, 2023.

\bibitem[Young et~al.(2014)Young, Lai, Hodosh, and Hockenmaier]{flickr30k}
Peter Young, Alice Lai, Micah Hodosh, and Julia Hockenmaier.
\newblock From image descriptions to visual denotations: New similarity metrics
  for semantic inference over event descriptions.
\newblock \emph{Transactions of the Association for Computational Linguistics},
  pp.\  67--78, 2014.

\bibitem[Yu et~al.(2023)Yu, Yang, Li, Wang, Lin, Liu, Wang, and
  Wang]{yu2023mmvet}
Weihao Yu, Zhengyuan Yang, Linjie Li, Jianfeng Wang, Kevin Lin, Zicheng Liu,
  Xinchao Wang, and Lijuan Wang.
\newblock Mm-vet: Evaluating large multimodal models for integrated
  capabilities.
\newblock \emph{CoRR}, abs/2308.02490, 2023.

\bibitem[Zhang et~al.(2022{\natexlab{a}})Zhang, Sun, Zhou, He, Yin, Wang,
  Sheng, Qiao, Shao, and Liu]{zhang2022bamboo}
Yuanhan Zhang, Qinghong Sun, Yichun Zhou, Zexin He, Zhenfei Yin, Kun Wang,
  Lu~Sheng, Yu~Qiao, Jing Shao, and Ziwei Liu.
\newblock Bamboo: Building mega-scale vision dataset continually with
  human-machine synergy.
\newblock \emph{CoRR}, abs/2203.07845, 2022{\natexlab{a}}.

\bibitem[Zhang et~al.(2022{\natexlab{b}})Zhang, Yin, Shao, and
  Liu]{Omnibenchmark}
Yuanhan Zhang, Zhenfei Yin, Jing Shao, and Ziwei Liu.
\newblock Benchmarking omni-vision representation through the lens of visual
  realms.
\newblock In \emph{ECCV}, pp.\  594--611, 2022{\natexlab{b}}.

\bibitem[Zhang et~al.(2023)Zhang, Zhang, Li, Zhao, Karypis, and Smola]{cotref}
Zhuosheng Zhang, Aston Zhang, Mu~Li, Hai Zhao, George Karypis, and Alex Smola.
\newblock Multimodal chain-of-thought reasoning in language models.
\newblock \emph{CoRR}, abs/2302.00923, 2023.

\bibitem[Zheng et~al.(2023)Zheng, Chiang, Sheng, Zhuang, Wu, Zhuang, Lin, Li,
  Li, Xing, Zhang, Gonzalez, and Stoica]{zheng2023judging}
Lianmin Zheng, Wei{-}Lin Chiang, Ying Sheng, Siyuan Zhuang, Zhanghao Wu,
  Yonghao Zhuang, Zi~Lin, Zhuohan Li, Dacheng Li, Eric~P. Xing, Hao Zhang,
  Joseph~E. Gonzalez, and Ion Stoica.
\newblock Judging llm-as-a-judge with mt-bench and chatbot arena.
\newblock \emph{CoRR}, abs/2306.05685, 2023.

\bibitem[Zhu et~al.(2023)Zhu, Chen, Shen, Li, and Elhoseiny]{minigpt4}
Deyao Zhu, Jun Chen, Xiaoqian Shen, Xiang Li, and Mohamed Elhoseiny.
\newblock Minigpt-4: Enhancing vision-language understanding with advanced
  large language models.
\newblock \emph{CoRR}, abs/2304.10592, 2023.

\end{thebibliography}

\appendix

\end{document}